%% file: main_arxiv.tex
\documentclass[11pt, a4paper, logo, acopyright]{googledeepmind}
\usepackage[authoryear, sort&compress, round]{natbib}

\usepackage{cleveref}
\usepackage{url}
\usepackage{wrapfig,booktabs}
\usepackage{xcolor}         
\usepackage{tikz}
\usetikzlibrary{patterns, shapes.geometric, positioning}
\usepackage{colortbl}
\usepackage{adjustbox}
\usepackage{multirow}
\usepackage{enumitem}
\usepackage{subfigure}
\usepackage{gensymb}
\usepackage{xspace} 
\usepackage{subcaption} 
\usepackage{graphicx}
\usepackage{makecell}  
\usepackage{minitoc}             
\usepackage{makeidx}
\usepackage{booktabs}
\usepackage{bbm}
\usepackage{algorithm}
\usepackage{algorithmic}
\usepackage{colortbl}
\usepackage{psfrag}
\usepackage{adjustbox}
\usepackage{wrapfig}
\usepackage{xspace}
\usepackage{amssymb}
\usepackage{multirow}
\usepackage{afterpage}
\usepackage{float}
\usepackage[toc,page,header]{appendix}
\usepackage{minitoc}
\usepackage{placeins} 
\usepackage{titletoc} 
\usepackage{comment}
\usepackage{svg}
\usepackage{booktabs}
\usepackage{multirow}
\usepackage{siunitx}

\usepackage{listings}

\usepackage{longtable}
\usepackage{booktabs}
\usepackage{nicematrix}
\usepackage[table]{xcolor}

\usepackage{graphicx}
\usepackage{booktabs}
\usepackage[table]{xcolor}
\usepackage{array}

\usepackage{tabularray}
\usepackage{xcolor}
\usepackage{float}

\usepackage{tcolorbox}
\tcbuselibrary{breakable}

\usepackage{minted}
\usepackage{pifont}
\usepackage{makecell, tabularx}
\usepackage{siunitx}
\usepackage[
  separate-uncertainty = true,
  multi-part-units = repeat
]{siunitx}

\newcounter{surveybox}
\newtcolorbox[use counter=surveybox]{surveybox}[2][]{
    breakable,
    colback=gray!5,        
    colframe=gray!50,      
    colbacktitle=gray!30,  
    coltitle=black,        
    fonttitle=\bfseries,   
    title={Survey \thesurveybox: #2},
    title after break={Survey \thesurveybox: #2 (Continued)},
    arc=2mm,
    boxrule=0.5mm,
    left=6mm, right=6mm, top=4mm, bottom=4mm,
    #1                     
}

\newcounter{codebox}
\newtcolorbox[use counter=codebox]{codebox}[2][]{
    breakable,
    colback=gray!5,        
    colframe=gray!50,      
    colbacktitle=gray!30,  
    coltitle=black,        
    fonttitle=\bfseries,   
    title={Code \thecodebox: #2},
    title after break={Code \thecodebox: #2 (Continued)},
    arc=2mm,
    boxrule=0.5mm,
    left=6mm, right=6mm, top=4mm, bottom=4mm,
    #1                     
}

\newlist{checklist}{itemize}{2}
\setlist[checklist]{label=$\square$}

\colorlet{darkgreen}{green!65!black}
\colorlet{darkblue}{blue!75!black}
\colorlet{darkred}{red!80!black}
\definecolor{lightblue}{HTML}{0071bc}
\definecolor{lightgreen}{HTML}{39b54a}
\definecolor{manyshot}{HTML}{6969ff}
\definecolor{medshot}{HTML}{f7c600}
\definecolor{fewshot}{HTML}{ff6969}
\definecolor{mypurple}{HTML}{412F8A}
\definecolor{myorange}{HTML}{fc8e62}
\definecolor{textgreen}{RGB}{57, 172, 57}
\definecolor{textred}{RGB}{200, 10, 10}
\definecolor{textgray}{RGB}{100, 100, 100}
\definecolor{visiongold}{RGB}{230, 184, 0}
\definecolor{speechpurple}{RGB}{204, 0, 255}
\definecolor{dataprep}{RGB}{38, 189, 128}
\definecolor{modeltraining}{RGB}{38, 189, 128}
\definecolor{backgroundcol}{RGB}{232, 230, 230}
\definecolor{gold}{rgb}{225, 215, 200} 
\definecolor{navyblue}{RGB}{40, 66, 200} 
\definecolor{orange}{RGB}{255,127,80} 
\definecolor{pink}{RGB}{219,112,147} 
\definecolor{baselinecolor}{gray}{.95}
\definecolor{deemph}{gray}{0.55}
\definecolor{bgcode}{rgb}{0.95,0.95,0.95}

\newcommand{\grayrow}{\rowcolor[gray]{.9}}

\newcommand{\bcell}[2]{\cellcolor{RoyalBlue!#1}#2}  

\definecolor{violettwilight}{HTML}{A183D8}
\definecolor{visiongold}{HTML}{F7BF3B}
\definecolor{cardiored}{HTML}{D62828}
\definecolor{pulmblue}{HTML}{0071bc}
\definecolor{sleepviolet}{HTML}{7B52AB}
\definecolor{motiongreen}{HTML}{2D8A5B}
\definecolor{surfaceyellow}{HTML}{E39600}

\newcommand{\CardiovascBadge}[1][0.92]{
    \tikz[baseline=(char.base)]\node[shape=rectangle,fill=cardiored, text=white,
          rounded corners=4pt, inner sep=2pt] (char) {CRD};\xspace
}
\newcommand{\CardiopulmBadge}[1][0.92]{
    \tikz[baseline=(char.base)]\node[shape=rectangle,fill=pulmblue, text=white,
          rounded corners=4pt, inner sep=2pt] (char) {PLM};\xspace
}
\newcommand{\SleepBadge}[1][0.92]{
    \tikz[baseline=(char.base)]\node[shape=rectangle,fill=sleepviolet, text=white,
          rounded corners=4pt, inner sep=2pt] (char) {SLP};\xspace
}
\newcommand{\MotionBadge}[1][0.92]{
    \tikz[baseline=(char.base)]\node[shape=rectangle,fill=motiongreen, text=white,
          rounded corners=4pt, inner sep=2pt] (char) {MTN};\xspace
}
\newcommand{\SkinBadge}[1][0.92]{
    \tikz[baseline=(char.base)]\node[shape=rectangle,fill=surfaceyellow, text=white,
          rounded corners=4pt, inner sep=2pt] (char) {SKN};\xspace
}
\newcommand{\AccelerometerCircle}[1][0.92]{
    \tikz[baseline=(char.base)]\node[shape=rectangle,fill=textgreen, text=white,
          rounded corners=4pt, inner sep=2pt] (char) {ACC};\xspace
}

\definecolor{regression}{HTML}{EF3E42} 
\definecolor{binary}{HTML}{1E90FF} 
\definecolor{multiclass}{HTML}{FFB482} 

\definecolor{regression_metric}{HTML}{942629}
\definecolor{binary_metric}{HTML}{0f4d8a}
\definecolor{gen_metric}{HTML}{430e8a}

\definecolor{selfsurvey}{HTML}{708090}
\definecolor{lab}{HTML}{CDB885}
\definecolor{screener}{HTML}{9F8383} 

\newcommand{\SelfSurveyLabel}[1][0.92]{
    \tikz[baseline=(char.base)]\node[shape=rectangle,fill=selfsurvey, text=white,
          rounded corners=4pt, inner sep=2pt] (char) {SLF};\xspace
}
\newcommand{\ScreenerLabel}[1][0.92]{
    \tikz[baseline=(char.base)]\node[shape=rectangle,fill=screener, text=white,
          rounded corners=4pt, inner sep=2pt] (char) {SCR};\xspace
}
\newcommand{\LabTestLabel}[1][0.92]{
    \tikz[baseline=(char.base)]\node[shape=rectangle,fill=lab, text=white,
          rounded corners=4pt, inner sep=2pt] (char) {LAB};\xspace
}

\newcommand{\RegressionLabel}[1][0.92]{
    \tikz[baseline=(char.base)]\node[shape=rectangle,fill=regression, text=white,
          rounded corners=4pt, inner sep=2pt] (char) {REG};\xspace
}

\newcommand{\Binary}[1][0.92]{
    \tikz[baseline=(char.base)]\node[shape=rectangle,fill=binary, text=white,
          rounded corners=4pt, inner sep=2pt] (char) {CLS};\xspace
}

\newcommand{\FOneLabel}[1][0.92]{
    \tikz[baseline=(char.base)]\node[shape=rectangle,fill=binary_metric, text=white,
        rounded corners=4pt, inner sep=2pt] (char) {\makebox[\widthof{MAE}][c]{F1}};\xspace
}
\newcommand{\ROCAUCLabel}[1][0.92]{
    \tikz[baseline=(char.base)]\node[shape=rectangle,fill=binary_metric, text=white,
          rounded corners=4pt, inner sep=2pt] (char) {ROC};\xspace
}
\newcommand{\PearsonLabel}[1][0.92]{
    \tikz[baseline=(char.base)]\node[shape=rectangle,fill=regression_metric, text=white,
        rounded corners=4pt, inner sep=2pt] (char) {\makebox[\widthof{MAE}][c]{$r$}};\xspace
}
\newcommand{\MAELabel}[1][0.92]{
    \tikz[baseline=(char.base)]\node[shape=rectangle,fill=regression_metric, text=white,
          rounded corners=4pt, inner sep=2pt] (char) {MAE};\xspace
}
\newcommand{\MSELabel}[1][0.92]{
    \tikz[baseline=(char.base)]\node[shape=rectangle,fill=gen_metric, text=white,
          rounded corners=4pt, inner sep=2pt] (char) {MSE};\xspace
}
\newcommand{\DeltaROCAUCLabel}[1][0.92]{
    \tikz[baseline=(char.base)]\node[shape=rectangle,fill=binary_metric, text=white,
          rounded corners=4pt, inner sep=2pt] (char) {$\Delta$ ROC};\xspace
}
\newcommand{\DeltaPearsonLabel}[1][0.92]{
    \tikz[baseline=(char.base)]\node[shape=rectangle,fill=regression_metric, text=white,
        rounded corners=4pt, inner sep=2pt] (char) {\makebox[\widthof{$\Delta$ ROC}][c]{$\Delta$ $r$}};\xspace
}

\definecolor{demographics}{HTML}{667B94}
\definecolor{lifestyle}{HTML}{2D8A5B}
\definecolor{cardiovascular}{HTML}{D62828}
\definecolor{metabolics}{HTML}{E39600}
\definecolor{mental}{HTML}{4b66db}
\definecolor{sleep}{HTML}{976dc9}

\newcommand{\Demo}{\tikz[baseline=-0.7ex]\fill[demographics] (0,0) circle (0.7ex);\xspace}
\newcommand{\Life}{\tikz[baseline=-0.7ex]\fill[lifestyle] (0,0) circle (0.7ex);\xspace}
\newcommand{\Cardio}{\tikz[baseline=-0.7ex]\fill[cardiovascular] (0,0) circle (0.7ex);\xspace}
\newcommand{\Metabolic}{\tikz[baseline=-0.7ex]\fill[metabolics] (0,0) circle (0.7ex);\xspace}
\newcommand{\Mental}{\tikz[baseline=-0.7ex]\fill[mental] (0,0) circle (0.7ex);\xspace}
\newcommand{\Sleep}{\tikz[baseline=-0.7ex]\fill[sleep] (0,0) circle (0.7ex);\xspace}

\newcommand{\modelname}{SensorFM\xspace}

\title{Towards a General Intelligence and Interface for Wearable Health Data}
\reportnumber{} 
\renewcommand{\today}{2026-05-18}

\correspondingauthor{$\{$girishvn, xumax, xliucs, dmcduff$\}$@google.com}
\author[$\circ$,$\dagger$,1,3]{Girish Narayanswamy}
\author[$\circ$,$\dagger$,1,5]{Maxwell A. Xu}
\author[$\ddagger$,1]{A. Ali Heydari}
\author[$\ddagger$,1]{Samy Abdel-Ghaffar}
\author[$\ddagger$,1]{Marius Guerard}
\author[$\ddagger$,1]{Kara Vaillancourt}
\author[$\ddagger$,1,3]{Zhihan Zhang}
\author[$\ddagger$,1]{Jake Garrison}
\author[$\ddagger$,1]{Levi Albuquerque}
\author[$\ddagger$,1]{Dimitris Spathis}
\author[$\ddagger$,1]{Hong Yu}
\author[$\ddagger$,1]{Hamid Palangi}
\author[1]{Xuhai "Orson" Xu}
\author[2]{David G.T. Barrett}
\author[1]{Joseph Breda}
\author[1,3]{Jed McGiffin}
\author[1]{Yubin Kim}
\author[1]{Yuwei Zhang}
\author[1]{Naghmeh Rezaei}
\author[1]{Samuel Solomon}
\author[1]{Karan Ahuja}
\author[1]{Tim Althoff}
\author[1,3]{Jake Sunshine}
\author[1]{Ming-Zher Poh}
\author[1]{Benjamin Yetton}
\author[4]{Ari Winbush}
\author[4]{Nicholas B. Allen}
\author[5]{James M. Rehg}
\author[2]{Isaac Galatzer-Levy}
\author[1]{Yun Liu}
\author[1]{John Hernandez}
\author[1]{Anupam Pathak}
\author[1]{Conor Heneghan}
\author[1]{Yuzhe Yang}
\author[1]{Ahmed A. Metwally}
\author[2]{Pushmeet Kohli}
\author[1]{Mark Malhotra}
\author[1,3]{Shwetak Patel} 
\author[$\triangle$,$\dagger$,1,3]{Xin Liu}
\author[$\triangle$,$\dagger$,1,3]{Daniel McDuff}

\affil[$\circ$]{Co-first}
\affil[$\triangle$]{Co-last}
\affil[$\ddagger$]{Core Contributor}
\affil[$\dagger$]{Corresponding Author}
\affil[1]{Google Research}
\affil[2]{Google DeepMind}
\affil[3]{University of Washington}
\affil[4]{University of Oregon}
\affil[5]{University of Illinois Urbana-Champaign}

\input{1-abstract}

\begin{document}
\maketitle

\newenvironment{Itemize}{
    \begin{itemize}[leftmargin=*]
    \setlength{\itemsep}{0pt}
    \setlength{\topsep}{0pt}
    \setlength{\partopsep}{0pt}
    \setlength{\parskip}{0pt}}
{\end{itemize}}
\setlength{\leftmargini}{9pt}

\input{2-intro}

\input{3-results}

\bibliography{main_arxiv}

\newpage
\renewcommand{\thesection}{M.\arabic{section}}
\renewcommand{\thefigure}{ED.\arabic{figure}}
\renewcommand{\thetable}{ED.\arabic{table}}
\renewcommand{\theequation}{M.\arabic{equation}} 

\setcounter{surveybox}{0}
\renewcommand{\thesurveybox}{ED.\arabic{surveybox}}

\setcounter{codebox}{0}
\renewcommand{\thecodebox}{ED.\arabic{codebox}}

\setcounter{section}{0}
\setcounter{figure}{0}
\setcounter{table}{0}
\setcounter{equation}{0}

\noindent \textbf{\LARGE{Methods}}
\normalfont

\input{4a-methods_data}
\input{4b-methods_model_and_training}
\input{4c-methods_latent_analysis}
\input{4d-methods_ARXIV_classroom}
\input{4e-methods_pha}
\input{4f-methods_ARXIV_additional_info}

\newpage
\input{appendix}
\end{document}

%% file: 1-abstract.tex
\begin{abstract}

While ubiquitous wearable sensors capture a wealth of behavioral and physiological information, effectively transforming these signals into personalized health insights is challenging. Specifically, converting low-level sensor data into representations capable of characterizing higher-level states is difficult due to high phenotypic diversity and variation in individual baseline health, physiology, and lifestyle factors. Moreover, collecting wearable data paired with health outcome annotations is laborious and expensive, and retrospective annotation remains practically unfeasible, contributing to a scarcity of data with high-quality labels. To overcome these limitations, we propose a foundation model for wearable health that is pretrained on more than \emph{one trillion minutes} of unlabeled sensor signals drawn from a large cohort of five million participants. We demonstrate that the joint scaling of model capacity and pretraining data volume leads to systematic improvements in performance, as evaluated on a diverse set of 35 health prediction tasks, spanning cardiovascular, metabolic, sleep, and mental health, as well as lifestyle choices and demographic factors. We find that this population scale representation unlocks label-efficient few-shot learning and generative capabilities for robust daily metric estimation. To further leverage this learned representation, we deploy a classroom of LLM agents to autonomously search the space of downstream predictive heads built on the model embeddings, showing broad performance improvements that increase with LLM model capacity. Finally, we show how integrating these downstream predictors into a Personal Health Agent can support model responses that are more relevant, contextually aware, and safe, and we validate this via 1,860 ratings from a cohort of clinicians.

\vspace{-0.1cm}
\end{abstract}

%% file: 2-intro.tex
\section{Introduction}
\vspace{-0.1cm}
\label{sec:intro}

Health is multidimensional. It comprises everything from the functioning of our many biological systems to abstract mental states to general well-being. Historically, both gathering and processing the volume of data required to provide individualized health insights to people have presented key technological bottlenecks in the pursuit of personalized healthcare.
The large-scale adoption of wearable and mobile health technologies presents an unprecedented opportunity to broaden access to personal health insights and facilitate a shift towards preventive care. The data captured by these devices, in the form of continuous and longitudinally sampled sensor streams, now enables more accessible and accurate measurements of physical activity and behavior than ever before~\citep{ringeval2020fitbit, munos2016mobile, mcduff2025evidence}. Furthermore, a growing body of evidence suggests that these high-resolution data streams may be useful in disease detection and phenotyping~\citep{yang2022artificial, metwally2026insulin}, early detection and monitoring~\citep{perez2019large} and the delivery of interventions~\citep{shah2025automated}. However, converting low-level sensor data into representations amenable to characterizing higher-level health states is challenging. Modern solutions largely take the form of bespoke supervised algorithms for individual health outcomes. Unfortunately, these algorithms are hindered by their reliance on sensor data paired with health outcome annotations, which are laborious, expensive, and infeasible to source post-hoc. Prior work has often taken a piecemeal approach, utilizing only a few sensing modalities that target a small set of health endpoints. However, this approach falls short of being able to represent the high degree of phenotypic diversity at the population level, with heterogeneity in baseline health and physiology and disparate downstream health outcomes. As such, it is yet unclear whether generalizable features can be learned from wearable sensor data that capture information useful for diverse individuals and health applications.

Foundation models, a recent advancement in machine learning, promise to address these limitations and accelerate progress in the field. These models often leverage self-supervised learning (SSL) over large-scale, heterogeneous, unlabeled datasets to learn universal representations which can generalize to a broad array of tasks. In the domain of health, these methods have successfully enabled improved performance on diverse applications, including radiology~\citep{wu2025towards}, pathology~\citep{xu2024whole}, and medical reasoning~\citep{mcduff2025towards, sellergren2025medgemma}. Adjacently, early time-series foundation models \citep{garza2023timegpt, rasul2023lag} demonstrated the utility of large-scale pretraining for signal forecasting, while subsequent families of models have emphasized the utility of learning common spectral properties across disparate domains~\citep{das2023decoder, goswami2024moment, ansari2024chronos}. The surprising capability of large-language models (LLMs) for time-series analysis further emphasizes the utility of scaled generalized pretraining across domains \citep{liu2023large, merrill2024language, thukral2025layout}.

In the wearable domain, recent efforts have demonstrated the potential to learn robust representations from large corpora of multimodal sensor data employing SSL  \citep{yuan2024self, thapa2024sleepfm, abbaspourazad2023large}. A pivotal advancement has been the recent establishment of scaling laws for wearable data and the introduction of architectures capable of learning directly from incomplete sensor signals \citep{narayanswamyscaling, xu2025lsm}. Yet, despite this progress, it remains unclear the full extent to which scaled pretraining on wearable data equates to meaningful improvements in predictive performance for diverse health outcomes and insights. Furthermore, current approaches remain constrained by two critical bottlenecks: the finite scale of pretraining data and the extensive manual engineering required to adapt a single generalist embedding to many distinct health endpoints.

\begin{figure*}[!p]
    \centering
    \includegraphics[width=\textwidth]{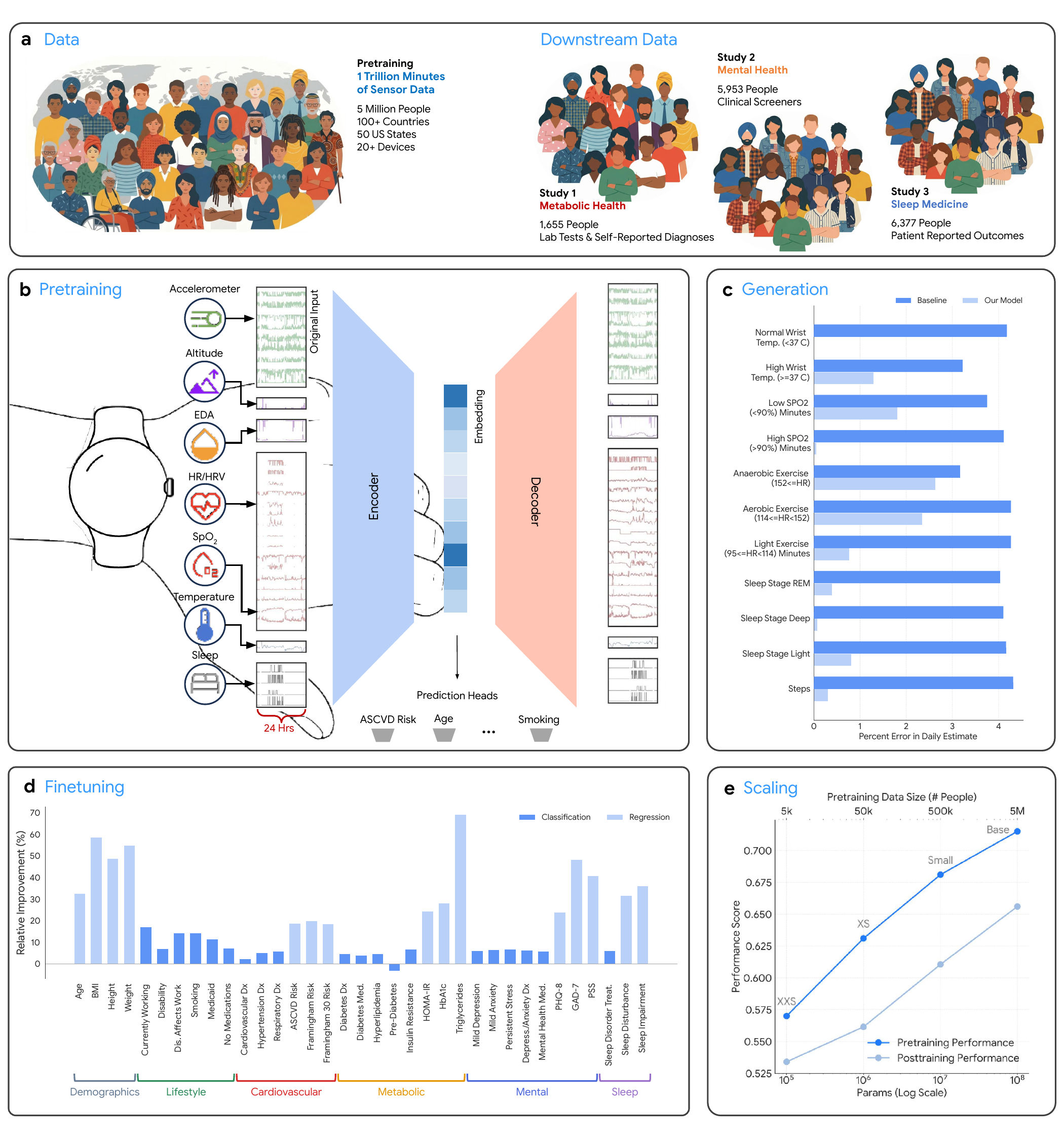}
    \caption{\textbf{Scaling and Evaluating a Sensor Foundation Model (SensorFM) for Wearable Health.} We present a versatile embedding model that scales with model and data capacity, and shows generalizability to a range of generative and discriminative tasks. \textbf{(a)} We pretrain this model on an unprecedented corpus of over one trillion minutes of sensor data drawn from five million participants and evaluate it on an independent set of data from 13,985 people featuring 35 clinical and behavioral discriminative tasks derived from three prospective studies. \textbf{(b)} The model is trained with a generative reconstruction objective with a latent ``bottleneck'' on features derived from five sensor modalities. 
    We evaluate the model and present results on a set of \textbf{(c)} generative tasks -- here the baseline represents daily estimates without generative infilling, and 
    \textbf{(d)} predictive tasks -- here we show relative performance improvement over a supervised model trained on engineered features. \textbf{(e)} Aggregated performance scores for pretraining and posttraining tasks show a linear correlation as pretrain data and model capacity are co-scaled by orders of magnitude, illustrating that a reconstruction based pretraining leads to scalable improvements in downstream tasks.}
    \label{fig:hero} 
\end{figure*}

In this work we aim to prove that leveraging large, unlabeled streams of continuous wearable sensor data during pretraining can lead to predictable improvements in both pretraining and downstream task objectives, and that resulting representations can be efficiently adapted to many outcomes. Building on this, we aim to demonstrate that providing such a model as an inference engine to an agentic health coach is more effective than having the coach process only the wearable data directly.

To that end we introduce \modelname (Figure~\ref{fig:hero}), a Large \emph{Sensor} foundation Model for wearable time-series representation learning which exhibits generalizability across health domains. By scaling pretraining to an unprecedented corpus of over one trillion minutes ($1,000,000,000,000$) of sensor data drawn from five million participants ($N=5,000,000$) and five sensor modalities, we approach a highly adaptable, universal representation of sensed human physiology. To our knowledge, this is the largest and most diverse wearable dataset utilized to date \citep{narayanswamyscaling, erturkbeyond, yuan2024self}. We evaluate \modelname across a comprehensive suite of downstream health tasks that span 
cardiovascular health, metabolic risk, sleep disorders, mental health, lifestyle choices and physiologically relevant demographics. We validate our model using rigorously phenotyped datasets derived from controlled clinical and laboratory studies ($N=13,985$). We further explore the capabilities of \modelname in reconstructing missing data and the resultant implications for daily health metric estimation. We comprehensively characterize the model's capabilities through rigorous evaluations of its scaling, label-efficiency, and interpretability.

\begin{figure*}[p!]
    \centering
    \includegraphics[width=\textwidth]{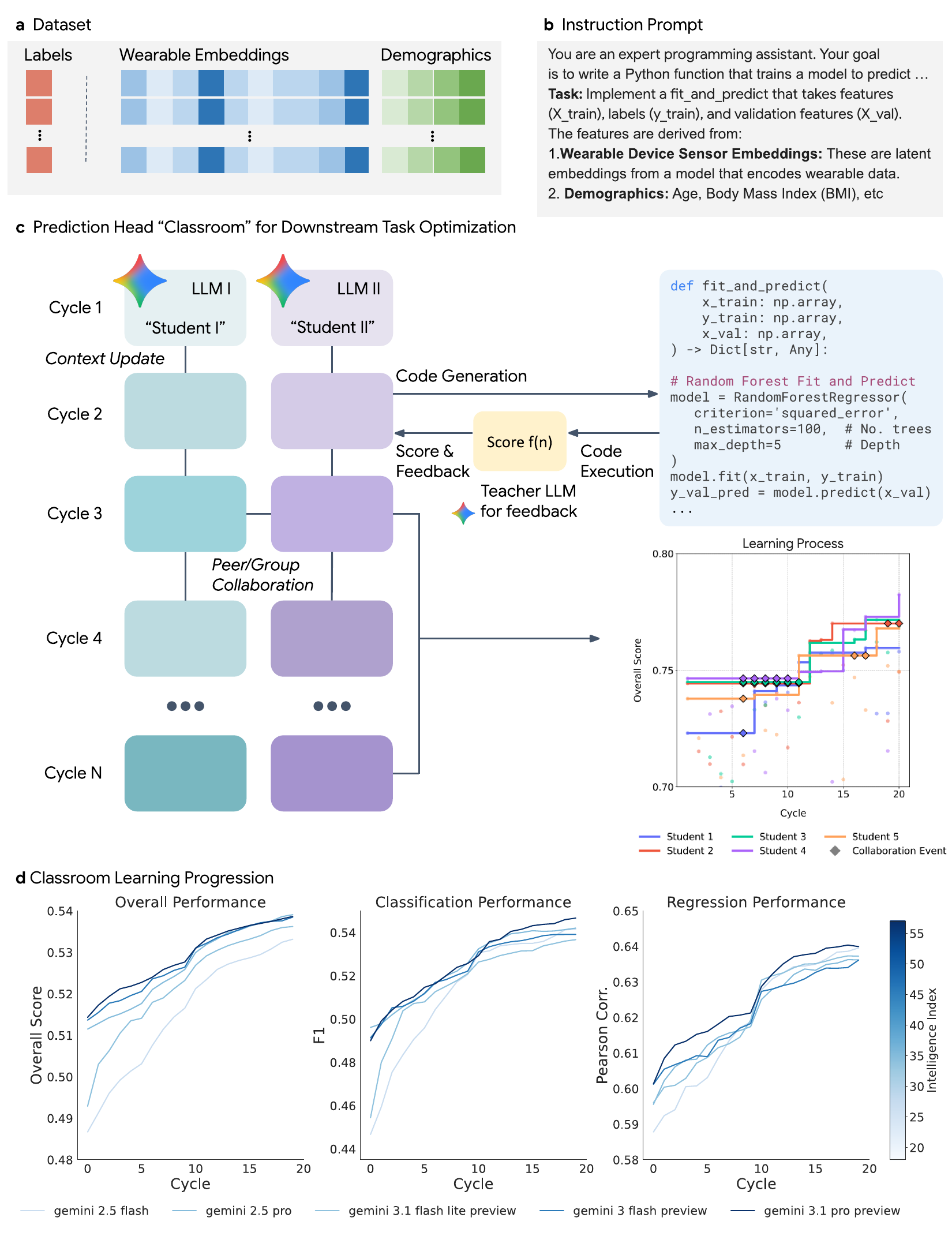}
    \caption{
    \textbf{Intelligent Search of Prediction Heads.} 
    We employ an LLM-driven architecture to efficiently search the space of solutions for each downstream task, mimicking the role of a machine learning expert. Specifically, \textbf{(a)} given a dataset of wearable embeddings, demographic features, and labels, \textbf{(b)} an instruction prompt, \textbf{(c)} a ``classroom'' of collaborative/competitive agents iteratively refines executable code solutions. \textbf{(d)} the learning progress for five ``student'' agents.}
    \label{fig:intelligent_search} 
\end{figure*}

Furthermore, to push the upper bound of predictive performance for diverse health applications, we leverage an automated agentic framework, inspired by modern self-improving code-generation systems~\citep{novikov2025alphaevolve, aygun2025ai}, to optimally adapt the \modelname embeddings to individual downstream tasks (Figure~\ref{fig:intelligent_search}). Traditionally, adapting general representations to a variety of downstream applications has required bespoke architecture engineering to train domain-specific models on the embeddings. In contrast, we provide an agentic ``classroom'' where LLM agents iteratively generate, test, and refine the code to develop models on downstream tasks using these embeddings as a starting point. We demonstrate how such a system enables more scalable and systematic exploration of the downstream solution space and leveraging this method autonomously conduct over $30,000$ individual experiments. We evaluate the benefit of allowing agents to play the role of machine learning engineers and analyze the agent-derived solutions.

\begin{figure*}[p!]
    \centering
    \includegraphics[width=0.9\textwidth]{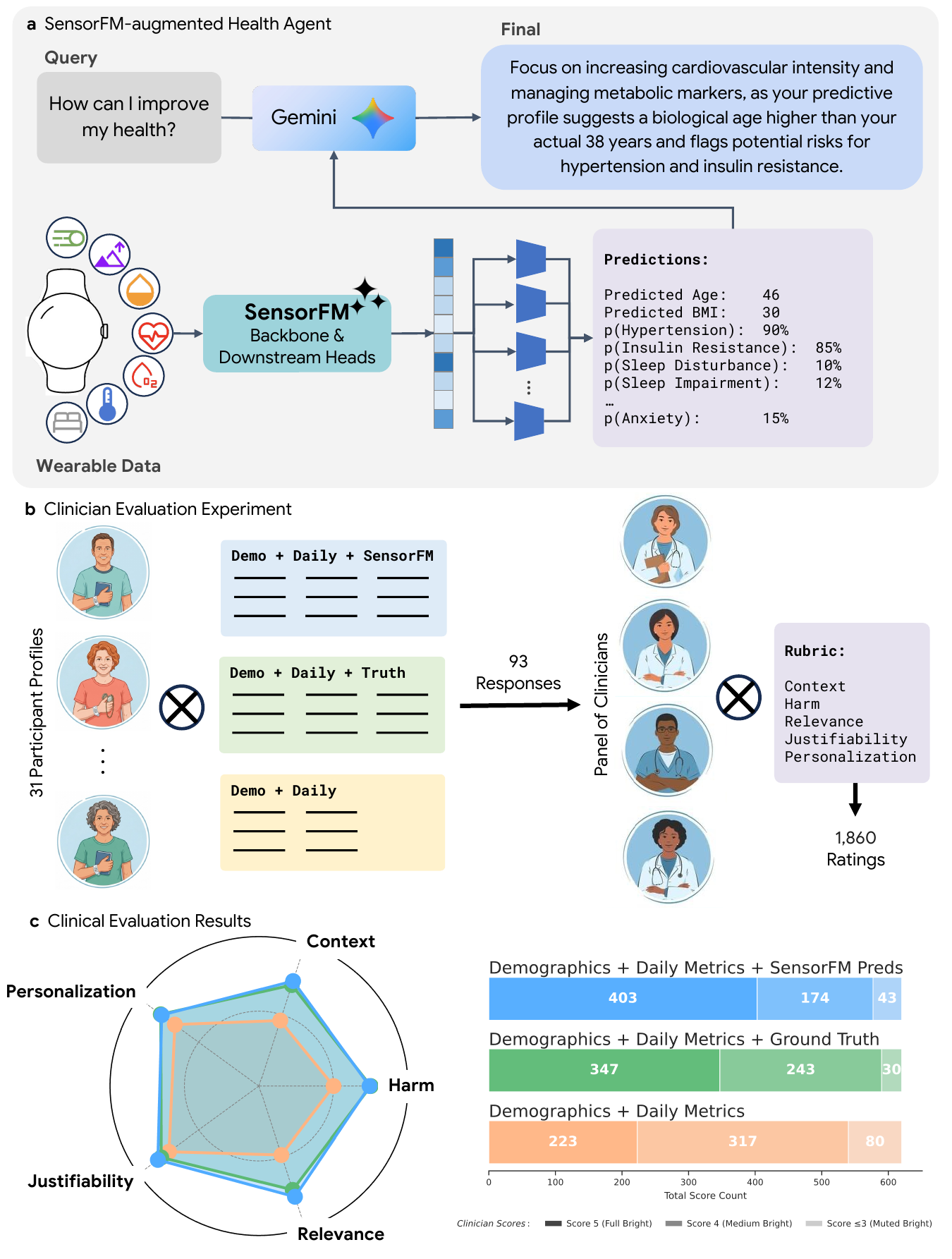}
    \caption{\textbf{Agentic Use of \modelname as a Tool.} \textbf{(a)} Architecture of the \modelname-augmented workflow, which translates raw wearable sensor data into health predictions to provide context for the LLM's response. \textbf{(b)} 
    We extracted demographic information, aggregated fitbit metrics, model predictions and ground-truth data for 31 real patient profiles and used Gemini to generate health summaries.  A panel of four clinicians evaluated the responses. \textbf{(c)} Average physician evaluation scores (Likert scale) plotted across five specific clinical rubric items (Harm, Context, Personalization, Justifiability, and Relevance) for the three evaluated conditions and stacked horizontal bar charts showing counts of Likert ratings across all rubric items. Example output is for illustrative purposes only.}
    \label{fig:agent} 
\end{figure*}

Finally, given the significant adoption of AI language models for consumer health queries~\citep{sumner2025perspectives, mcduff2025towards, tu2025towards, breda2026symptomai}, we establish the model's end-to-end utility by integrating \modelname as tool into a Personal Health Agent~\citep{heydari2025anatomy} and evaluate its utility to enhance the delivery of context-aware physiological insights to users (Figure~\ref{fig:agent}). We conduct over 40 hours of clinical evaluations of health summaries generated with wearable data and either \modelname predictions or gold-standard ground-truth measurements involving 1,860 individual ratings. When compared to a baseline in which the language model processes the wearable data directly, using \modelname as a tool improves the specificity of the responses and makes them more personalized, contextually appropriate and safer. When compared to a condition in which the agent has access to ground-truth measurements we observe no statistical inferiority.

In summary, our work represents the most comprehensive evaluation of pretrained wearable sensor foundation models to date, demonstrating how flexible embeddings, produced through scaled pretraining, can be efficiently fine-tuned for many downstream health applications, which can in turn be leveraged to provide valuable insights at the person level.

%% file: 3-results.tex
\section{Results}
\label{sec:results}

We first characterize the properties and evaluate the performance of \modelname along four dimensions: resource scaling, predictive performance across discriminative health tasks sourced from multiple prospective studies, generative capabilities for infilling and forecasting sensor data, and finally interpretation of the model embeddings in a latent space. We then present results from an agentic search method used to automatically design and refine application-specific prediction ``heads'' and evaluate this system across classification and regression-based health tasks. Finally we evaluate \modelname as a tool for a Personal Health Agent and recruit a cohort of clinicians to evaluate the utility of the model's predictions when creating health summaries for a set of 31 real (i.e., non-synthetic) health profiles.

\subsection{Scaling a Generalist Model for Wearable Health}
\label{subsec:results_lsm_scaling}

The establishment of scaling laws, and the resultant success of foundation models in domains such as language and vision, has shown that model performance is often driven not by architectural design, but rather by compute, model capacity, and ultimately the volume of training data \citep{kaplan2020scaling, zhai2022scaling}. Scaling laws provide empirical evidence that can be used to help anticipate the performance gains that could potentially be achieved if any (or all) of these resources are increased. Initial empirical evidence of scaling has been documented for time-series and wearable sensor signals \citep{shi2024scaling, narayanswamyscaling, zhang2025sensorlm}. Yet, a systematic set of scaling experiments on wearable sensor data demonstrating that progressive pretraining gains predictably translate to measurable improvements in estimating meaningful health outcomes is still absent.

In this work we substantially increase the scale of pretraining and compute beyond previous efforts. Specifically, we scale over \emph{four orders of magnitude} for both pretraining data volume (two million to two billion hours of multimodal sensor data) and model size (100K to 100M parameters). Our maximum data volume is a 50x increase in data-hours over and above the 40 million hours used by prior work for models trained on minute-resolution data \citep{narayanswamyscaling}. Compared to the 2.5 billion hours used to train models on hour-resolution data \citep{erturkbeyond}, our higher \emph{minutely} resolution data accounts for a 50x increase in total data volume.
Throughout our analysis we refer to data volumes by the number of individuals sampled or the associated multimodal data hours: \textbf{5K} ($2 \times 10^6$ hrs.), \textbf{50K} ($2 \times 10^7$ hrs.), \textbf{500K} ($2 \times 10^8$ hrs.), \textbf{5M} ($2 \times 10^9$ hrs.).
We refer to \modelname model variants by their capacities: \textbf{XXS}mall ($10^5$ params.), \textbf{XS}mall ($10^6$ params.), \textbf{S}mall ($10^7$ params.), \textbf{B}ase ($10^8$ params.). We evaluate the effect of this scaled pretraining on classification, regression, and generative tasks.

\textbf{The Importance of Scaling Pretraining Data and Model Capacity.} We observe that the pretraining validation loss inversely scales with increases in data volume and model capacity (Table~\ref{tab:scaling_results}). In so doing, we verify that scaling these resources leads to predictable improvements in model performance 
and that similar improvements are observed for both discriminative and generative downstream tasks. For example, when pretrained with the largest 5M subject data volume, \modelname-B consistently outperforms the smaller \modelname-XXS. The scaled model achieves a $31\%$ reduced validation loss (MSE) on the reconstruction pretraining task, and a $28\%$ reduced loss (avg. MSE) across generative tasks. On discriminative tasks, the scaled model achieves a mean improvement of $\Delta AUC = 0.09$ on classification tasks and $\Delta r = 0.21$ on regression tasks.

Crucially, we observe that the most significant gains are achieved through the joint scaling of \emph{both} data volume and model capacity. The proportional scaling of data and capacity by orders of magnitude, leads to near-linear improvements in both generative pretraining and discriminative post-training performance (see Figure~\ref{fig:hero}.e). \textbf{Driven by this finding, all following results, unless explicitly stated, assume that models are trained with data volumes proportionally scaled to their capacity}. The impact of this joint scaling on discriminative health tasks is further visualized in Figure~\ref{fig:downstream_bar_plots} and presented in Tables~\ref{tab:predictive_task_scaling} where across model variants, \modelname-B boasts a task win-rate of $33/35$, while XXS expectedly ranks last on $33/35$ tasks.

\subsection{Learning a Representation Useful Across Health Domains}
\label{subsec:results_lsm_discriminative}

We evaluate the \modelname-learned embeddings across a diverse range of 35 discriminative health tasks derived from multiple prospective studies (see Methods~\ref{subsec:downstream_datasets}).
These tasks span
\Cardio \emph{Cardiovascular Health} (6),
\Metabolic \emph{Metabolic Health} (8),
\Mental \emph{Mental Health} (8), 
\Sleep \emph{Sleep} (3),
\Demo \emph{Demographic} (4), and
\Life \emph{Lifestyle Factors} (6), with the full list of tasks found in Table~\ref{tab:downstream_labels}.
In order to interrogate the quality of the pretrained embeddings, we leverage a \emph{frozen} \modelname encoder and learn a computationally efficient linear head to adapt the embeddings to individual applications. To account for the limited number of annotated examples, these heads are trained with embeddings reduced to $50$ principal components. We baseline \modelname against supervised models trained with engineered features derived from the wearable sensor streams (see Methods~\ref{subsec:engineered_baseline_features}).
We further assess the lift of the learned sensor representation, by training models both with and without demographic features, comparing against baseline supervised models trained only with demographic features.

\textbf{The Learned Representation Generalizes to Diverse Health Outcomes.} We find that \modelname, through scaled pretraining, learns a representation capable of successfully generalizing to a broad range of health outcomes. As illustrated in Table~\ref{tab:predictive_tasks_againstfeatures}, linear heads trained on-top of the \modelname learned embeddings consistently outperform supervised baselines trained with engineered features. Specifically, \modelname outperforms this supervised baseline on $34$ of $35$ discriminative tasks (Figure~\ref{fig:hero}.d). \modelname outperforms a baseline trained on only demographic features on $24$ of $30$ discriminative tasks.

\textbf{The Utility of Demographic Features.} As highlighted in Table~\ref{tab:predictive_tasks_againstfeatures}, we find that the predictive power of \modelname is often, though modestly, enhanced through the addition of demographic features ($22$ of $30$ discriminative tasks). While demographic features tend to improve the performance of \modelname, interestingly we find that the dependence of \modelname on demographic features decreases with scale.
\modelname-B realizes smaller gains from added demographic features as compared to both smaller model variants and supervised baselines on $33$ of $35$ tasks, implying that these physiologically relevant traits may be implicitly learned through pretraining at scale (Table~\ref{tab:delta_predictive_tasks}). A similar trend is observed in the feature importance attributed to embeddings when adapted to discriminative downstream tasks alongside demographic features. Models pretrained at scale provide a more robust representation which reduces the reliance on demographic priors (Figure~\ref{fig:feature_importance}.b).

\begin{figure}[p!]
    \centering
    \includegraphics[width=\textwidth]{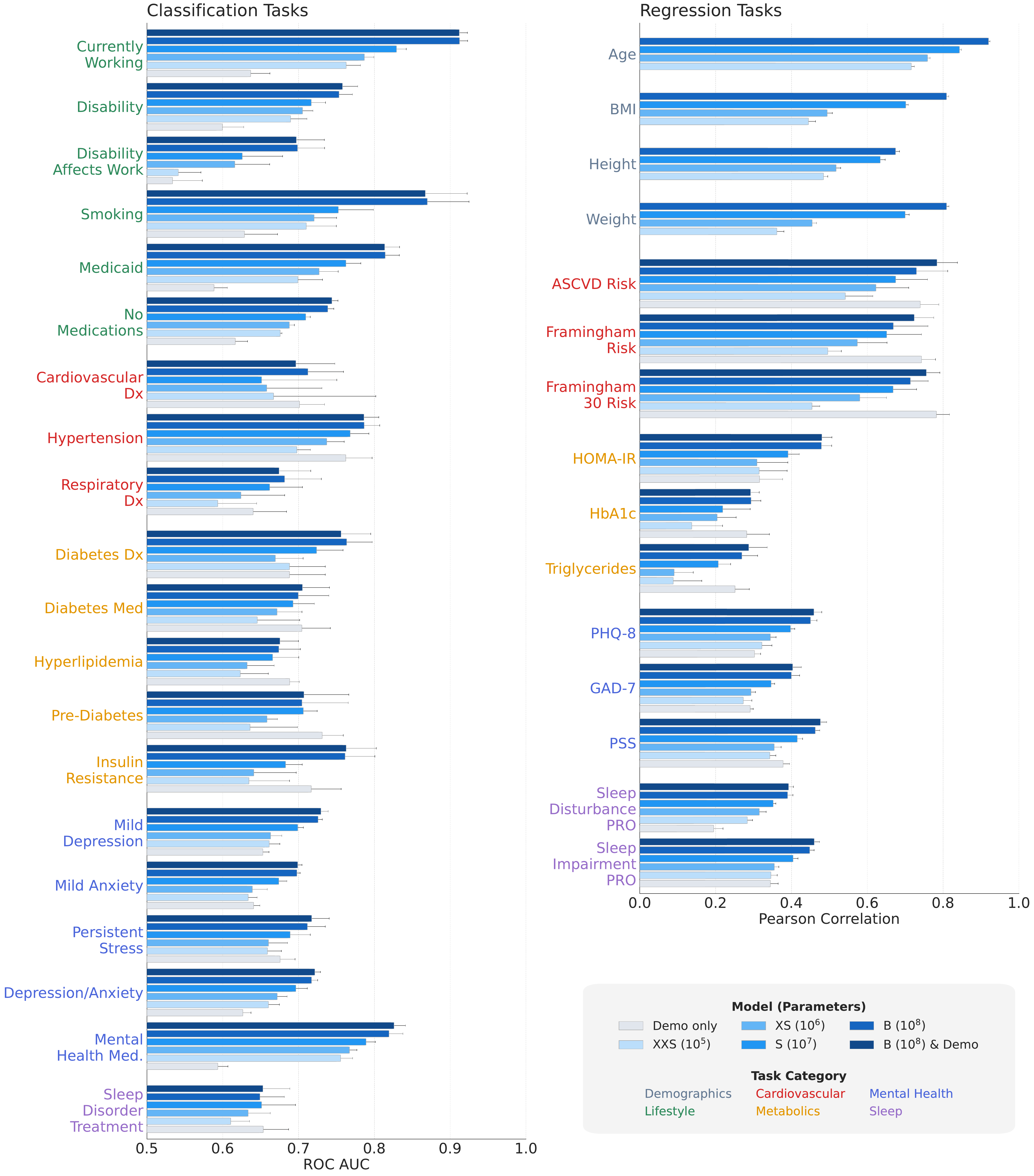}

    \caption{\textbf{Discriminative Task Linear Probe Performance.} Downstream performance across 35 discriminative tasks for \modelname variants, pretrained with proportional data scales, and a supervised baseline trained with only demographics. In general performance improves with scale with B consistently achieving the best performance.
    \modelname variants are post-trained with PCA-50 reduced embeddings.
    For each task, we report the average 5-fold cross validation performance. Average Receiver Operating Characteristic Area Under the Curve (ROC AUC) is calculated in the logit-transform space and back-transformed. Average Pearson correlation ($r$) is calculated in the z-transform space and back-transformed. Error bars are standard deviations calculated in the transformed space and back-transformed to give asymmetric error values.}

    \label{fig:downstream_bar_plots}
\end{figure}

Additionally, for some tasks (e.g., cardiovascular Dx, insulin resistance, ASCVD risk, Framingham risk, and more), we find that models trained with demographics alone provide significant predictive power, (Figure~\ref{fig:downstream_bar_plots}). For such tasks, \modelname may only outperform these demographic baselines at extreme pretraining scales (e.g., \modelname-B trained on 50M weeks of data). For a subset of these tasks (ASCVD Risk, Framingham Risk, Framingham 30 Risk), the strong performance of demographic-only models likely stems from the explicit use of demographic features to calculate these risk scores.

\textbf{Scaled Pretraining Enables Label Efficient Adaptation.}
We find that scaled pretraining and the resultant learned representation of \modelname enables improved label efficiency compared to supervised baselines. As depicted in Figure~\ref{fig:few_shot} we tested this by training models with varying percentages of downstream training volumes. With very few labeled samples, demographic priors act as a strong predictor for many tasks. However, as the number of labeled samples increases, \modelname soon outperforms demographic-only baselines and consistently outperforms the feature engineered baselines, with larger model variants (B) outperforming smaller variants (XXS). 

\subsection{Exploiting the Learned Generative Capabilities}
\label{subsec:results_lsm_generative}

\modelname leverages a reconstructive pre-text task which enables learning directly from unlabeled, passively collected sensor data. Leveraging an MAE-like architecture~\citep{he2022masked, xu2025lsm}, our method natively handles the missingness inherent in wearable sensor streams, a consequence of varying sensor configurations, operating modes, and user behaviors. During pretraining, \modelname learns a decoder capable of reconstructing ablated observations, which in-turn translates to generative capacities such as data imputation and signal forecasting. Figure~\ref{fig:reconstruction_heatmaps} presents examples of day-long windows obtained from participants in our pretraining \emph{validation} set, comparing the original model input to the reconstructed output. Figures~\ref{fig:reconstruction_line_graphs_1} and~\ref{fig:reconstruction_line_graphs_2} depict line graphs for individual sensor feature infilling and highlight the non-linear dynamics within the \modelname reconstructions.

\textbf{\modelname Learns to Fill and Forecast Sensor Data.} \modelname, through its generative pretraining, learns to successfully impute, interpolate, and extrapolate missing or unobserved data (see Table~\ref{tab:generative_results}). Specifically, we find that \modelname outperforms the best-performing baselines by 74.8\% on random imputation, 38.8\% on temporal interpolation, 39.6\% on temporal extrapolation, and 83.7\% on sensor signal imputation.

\textbf{Improved Daily Metric Estimation.} Wearable sensor feeds may be intermittently interrupted for a variety of reasons. Such interruptions can significantly skew an individual's health summary statistics. As such, it may be advantageous to provide individuals with more accurate estimates of their summary statistics allowing them to better gauge their overall health status. Towards this end, we explore the potential of leveraging \modelname's generative capabilities to impute missing data in order to more realistically estimate a person's daily metrics. Specifically, \modelname leverages temporal interpolation to infill missing segments. As highlighted in Table~\ref{tab:recovery_metrics}, we find that in the presence of missing or ablated data, \modelname is able to produce more reliable daily metrics. Specifically, we show that when ablating 60 contiguous minutes of data in a day, \modelname retains 99.7\% accuracy in daily step count prediction, 99.9\% accuracy in deep sleep prediction, and 99.2\% accuracy in light exercise tracking, mitigating the underestimation observed in baseline performance (Figure~\ref{fig:hero}.c).

\subsection{Understanding and Quantifying the Learned Latent Space}
\label{subsec:results_interpretability}

Analysis and visualization of the \modelname embeddings in latent space provide valuable insight into the learned representation, its structure, and its application to downstream tasks. Towards this end, we project and visualize the latent space for a number of health outcomes, analyze the embedding distances and intrinsic dimensionality of the model across scales, and interpret the importance of the embeddings with respect to downstream health outcomes.

\textbf{Visualizing Embeddings with Task Labels.}
We use Uniform Manifold Approximation and Projection (UMAP)~\citep{mcinnes2018umap} to reduce the high-dimensional latent embedding vectors from our largest model down to two dimensions (Figure~\ref{fig:embeddings_plot_1}). The manifold of the model provides insight into the information learned during pretraining in the absence of explicit labels. We visually annotate these points with labels from different downstream datasets. Note that although all plots use the same UMAP projection, not all participants/days have labels for all tasks. There are clear patterns of demographic shifts for BMI, Age, and Gender captured within the embedding. Ultimately, these visualizations underscore how self-supervised pretraining at this  scale naturally organizes fragmented sensor streams into a physiologically meaningful topology, validating a foundation model approach as a universal representation of sensed human health.

\textbf{Embedding Distances.} To investigate how model scale influences latent space density, we compare the dispersion of user representations across different model sizes (Figure~\ref{fig:embeddings_meta_analysis}.a), an approach to quantifying participant similarity as seen in previous works \citep{kiyasseh2021clocs}. We find that while all models yield unimodal, right-skewed distance distributions, their latent space dispersion varies significantly. The \modelname-S model learns the most tightly clustered representations, whereas the \modelname-B model produces the broadest embedding spread. The smallest model (\modelname-XXS) also exhibits a similarly broad spread, pointing to an interplay between optimal model capacity and data volume in shaping representation density. This structural evolution of the latent space confirms that the joint scaling of data and parameters is crucial for yielding an embedding space expressive enough to capture the nuanced, inter-subject variations across diverse clinical domains.

\textbf{Intrinsic Dimensionality and Compressibility}. We evaluate the compressibility and informational structure of the embeddings across model scales (Figure~\ref{fig:embeddings_meta_analysis}.b). An analysis reveals that the model embeddings are highly compressible; particularly for the larger models, representations can be reduced to 150-200 dimensions without significant loss of variance. Furthermore, the variance scaling behaviours differ significantly across model sizes. The smallest model (XXS) captures approximately 90\% of the variance within its first 20 principal components and then flatlines, indicating dimensional collapse—an over-reliance on a restricted feature manifold. Conversely, the largest model B shows strong anisotropy. It learns a large "super-feature" in its dominant direction, with the first principal component alone explaining approximately 40\% of the total variance. After this initial spike, the curve flattens out, indicating that while the large model relies heavily on a dominant primary component, it importantly preserves a significant "long tail" of nuanced physiological information distributed across higher dimensions. Consequently, large-scale sensor models provide a powerful dual advantage: they distill large, continuous streams of wearable data into highly efficient, compressible interfaces for downstream modelling, while crucially preserving the subtle, long-tail signals essential for predicting heterogeneous outcomes such as mood and sleep disorders.

\textbf{Interpreting Embedding Importances.}
In order to better understand the structure of the latent embedding vector, we leverage SHapley Additive explanations (SHAP)~\citep{lundberg2017unified} to assess the effect of individual latent dimensions.  We calculate SHAP values for each linear prediction head on the model and then compute the pair-wise $cosine$ similarity between the normalized, exact SHAP attribution (weight collapse analysis) profiles of
each pair of tasks. This indicates the degree to which distinct tasks leverage the same underlying embedding dimensions from the foundational model (Figure~\ref{fig:feature_importance}.a). Expected similarities in embeddings are observed with highly correlated labels such as ASCVD risk and Framingham risk, and weight and BMI.  However, we observe links between other tasks, including sleep impairment and PSS score, and HOMA-IR and PHQ-8 score.  

\subsection{Agent Driven Search of Downstream Model Heads}
\label{subsec:results_lsm_hybrid}

Versatile pretrained model embeddings which generalize to many predictive tasks are attractive as they enable application-specific models to be built more easily, especially when labels are sparse. As such, methods which efficiently design new prediction ``heads'' allow these embeddings to be more rapidly adapted to novel tasks.
A bottleneck to effectively adapting embeddings for downstream predictive tasks is often the expertise and iterative work required for feature engineering, model architecture selection, and hyperparameter tuning. This is especially true in health domains, where observational data is typically subject to significant constraints (e.g., sparsity, noise, limited volume, and imbalance of labels).

We implement a hybrid modeling system, adapted from \citet{aygun2025ai}, to leverage the reasoning and code-writing abilities of language models alongside an iterative-solution-search framework to autonomously adapt general embeddings to new domains. Specifically, we leverage a "classroom" configuration (described in~\ref{sec:classroom}), a set of self-evolving algorithm generation steps in which the solution synthesis is formulated as a competitive, collaborative optimisation problem solved by parallel LLM agents (Figure~\ref{fig:intelligent_search}). Leveraging this framework we rapidly iterate over $30,000$ agent-proposed solutions.

\begin{figure}[p!]
    \centering
    \includegraphics[width=\textwidth]{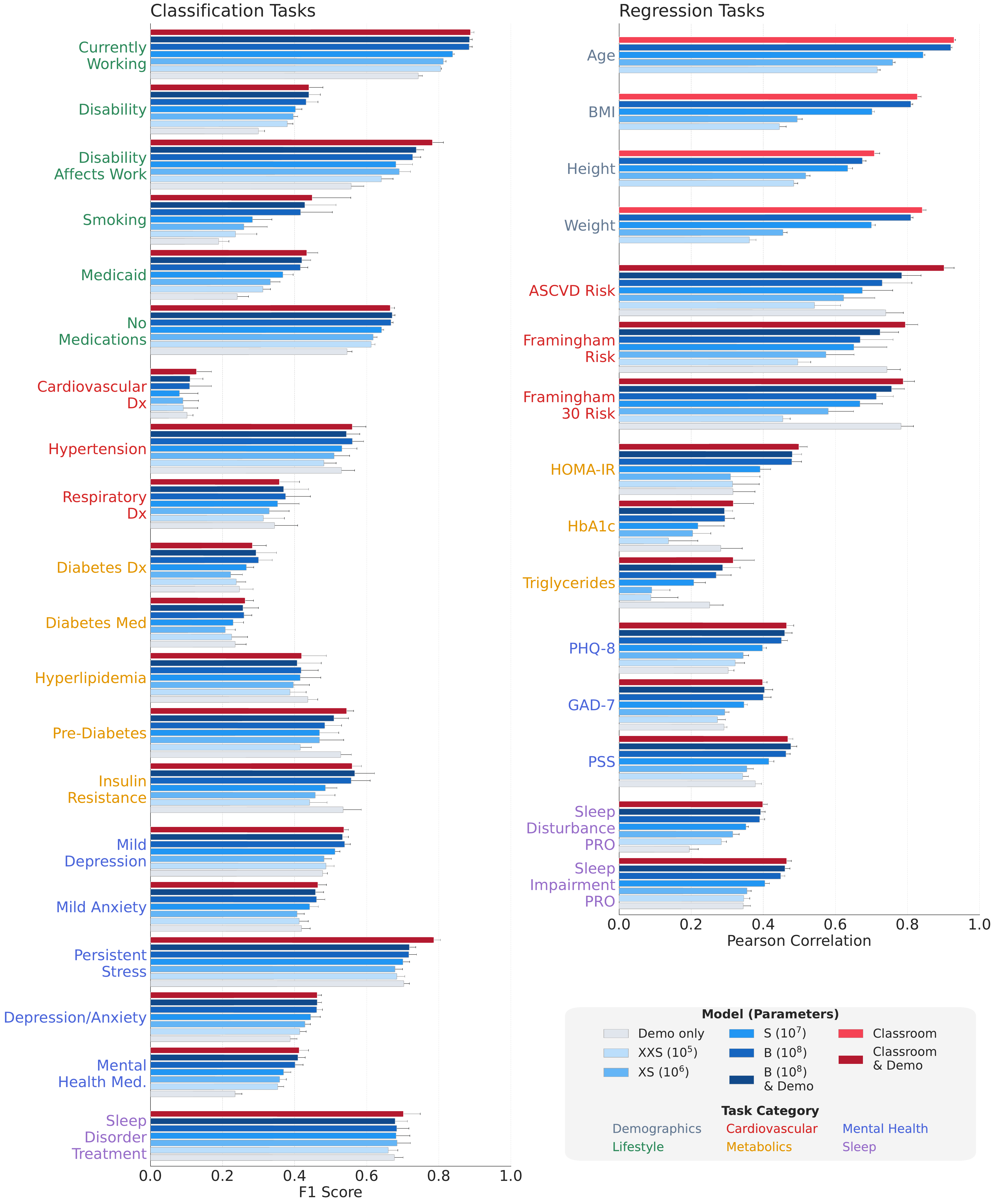}

    \caption{\textbf{Discriminative Task Agentic Classroom Solution Performance.} Downstream performance across 35 discriminative tasks for \modelname-B embeddings adapted with agentic classroom found solutions, linear probes of \modelname variants, and a demographic baseline. In general the classroom found solutions improve upon simple linear probes.
    Linear probes use PCA-50 reduced embeddings, the classroom uses unreduced embeddings.
    For each task, we report the average 5-fold cross validation performance. Average $F1$ is calculated with an arithmetic mean. Average Pearson correlation ($r$) is calculated in the z-transform space and back-transformed. Error bars are standard deviations calculated in the transformed space and back-transformed to give asymmetric error values.}

    \label{fig:downstream_bar_plots_with_classroom}
\end{figure}

\textbf{Improvements Across Health Tasks.}
We find that this agentic approach leads to improved performance as compared to a simple linear head applied to the \modelname embeddings across a breadth of discriminative health tasks. Specifically, classroom-derived agent solutions boast improved performance on $16$ of $20$ classification tasks and greater Pearson correlations on $12$ of $15$ regression tasks (Figure ~\ref{fig:downstream_bar_plots_with_classroom} and Table~\ref{tab:predictive_tasks_scaling_tree_search}). 
Note that we report F1 for these classification results as many solutions were ensemble methods from which it is not possible to obtain a continuous output with which to compute an ROC curve. In so doing we demonstrate the potential of AI agents to act as machine learning scientists, reducing the engineering burden traditionally associated with adapting general embeddings to multiple endpoints.

\textbf{Solution Performance Scales with LLM Capabilities.}
Analyses of the agent solutions organized by LLM model variant reveals an interesting pattern, with more recent models exhibiting stronger performance (see Figure~\ref{fig:codegen_meta_analysis}.a). When these models are characterized by the commonly used Artificial Analysis Intelligence Index\footnote{https://artificialanalysis.ai/evaluations/artificial-analysis-intelligence-index}, models with higher intelligence indexes provide better solutions on average.
Furthermore, we find that collaboration events between ``students'' (agents), triggered when a given agent demonstrates plateauing performance and is allowed to reflect on its own solutions or the solutions of other agents, enables less intelligent models to close this performance gap. However, the best performance is typically observed from more recent versions of Gemini~\citep{comanici2025gemini}.

\textbf{Analysis of the Found Solutions.}
Meta analysis of the best solutions found through the classroom algorithm search (see Figure~\ref{fig:codegen_meta_analysis}.b) reveals that almost all top-scoring solutions reduced the dimensionality of the embedding feature space to between $50-100$ dimensions, most likely to reduce the variance of the input space to match the scarcity of labeled examples. The results additionally reveal that linear models were more common than non-linear models. Ensembles were employed in just under a quarter of the best solutions. While manually searching this space of solutions is tractable, it is time consuming and often inefficient, becoming increasingly less feasible as the number of downstream prediction tasks increases. By contrast, this agent driven approach allows for efficient iteration, with the average quality of final solutions improving monotonically over time (Figure~\ref{fig:intelligent_search}.d).

\subsection{Agentic Use of the \modelname as a Tool}

A critical open question is whether providing \modelname, adapted to multiple endpoints, as an inference engine to a Personal Health Agent, yields improvements in the responses to user health queries as compared to the agent reasoning limited to handcrafted features. Towards this end, we evaluate the performance of a health agent under three distinct experimental conditions:

\begin{itemize}
\item \textbf{Condition (A):} \textit{Demographics + Daily Wearable Metrics + \modelname Predictions} 

Agent receives demographics, feature-engineered daily metrics, and \modelname predictions (e.g., predicted hyperlipidemia state, predicted hypertension state, etc.)

\item \textbf{Condition (B):} \textit{Demographics + Daily Wearable Metrics + Ground Truth}.

Similar to Condition (A) but with \modelname predictions replaced by the patient's actual ground-truth targets (e.g., actual hyperlipidemia state, actual hypertension state, etc.).

\item \textbf{Condition (C):} \textit{Demographics + Daily Wearable Metrics}

Baseline comparator: agent receives only demographics and feature-engineered daily metrics.
\end{itemize}

An overview of the \modelname-augmented agent setup, alongside the evaluation results, is presented in Figure \ref{fig:agent}. A set of experienced clinicians evaluated health summary responses generated by Gemini 3 Flash from experimental conditions A, B, and C, while blinded to the conditions. Over 40 hours of expert clinical annotations yielded a total of $n=1,860$ ratings across five rubric items (See Survey~\ref{box:pha_eval}).

\textbf{Extra Context Improves Health Agent Performance.} Pairwise comparisons using Wilcoxon signed-rank tests with a Bonferroni correction reveal that the extra context from either \modelname or ground-truth labels leads to significant improvements over the baseline condition (C) (\modelname A vs. C: $W = 10110$, $p < 0.001$; Ground truth B vs. C: $W = 9596$, $p < 0.001$).

The pattern is consistent across all five rubric dimensions. With \modelname predictions (A), the agent generates significantly stronger responses compared to the baseline (C) across every individual axis: Context ($W = 451$, $p < 0.001$), Personalization ($W = 378$, $p < 0.001$), Justifiability ($W = 300$, $p < 0.01$), Relevance ($W = 412$, $p < 0.001$), and Harm ($W = 510$, $p < 0.001$).
Providing the agent with ground-truth labels (Condition B) yields similar benefits across all dimensions (all $p$s $<0.0167$ after correction).

\textbf{Extra Context via \modelname Predictions Matches Extra Context via Ground Truth.} We find no statistically significant differences in performance between \modelname prediction (Condition A) and ground truth (Condition B) ($ p = 0.396$). This demonstrates that the diagnostic inferences generated by \modelname are comparable to the diagnostic ground truth when supplied as contextual inputs to a Personal Health Agent.

\section{Discussion}
\label{sec:discussion}

\subsection{Learning a Scaled Representation of Sensed Human Physiology}
Our scaling experiments reveal a predictable relationship between model performance, model capacity, and pretraining data volume. The most consistent performance gains were observed when model size and data volume were increased concurrently, with Figure~\ref{fig:hero}.e indicating that our pretraining method has yet to saturate.
This reinforces the hypothesis that both dimensions are essential to maximize gains through self-supervised pretraining and to learn robust representations of sensed physiology. 
Overall, our findings demonstrate how advancements in wearable AI may be driven by leveraging expansive, diverse, high-fidelity datasets. Unlike computer vision or natural language processing, which benefit from web-scale open corpora, wearable data is inherently difficult to aggregate, standardize, and share due to privacy and technical considerations. Consequently, data fidelity, additions of new modalities and longitudinal scale will likely be a primary scaling factor for the continued evolution of sensor foundation models.

\subsection{The Utility of Generalizable Embeddings for Wearable Health}

We demonstrate that large-scale self-supervised pretraining on a large corpora of wearable sensor data produces a robust representation of sensed human physiology and behavior that transfers effectively to diverse health domains. Across 35 distinct health and behavioral tasks, encompassing cardiovascular health, metabolic function, sleep architecture, mental health, lifestyle and demographic factors, linear probes of the \modelname embeddings consistently outperformed supervised baselines trained with engineered features. Crucially, \modelname achieves robust predictive accuracy even without task-specific architectures, suggesting that when trained at sufficient scale, wearable foundation models can learn representations that are broadly useful and label-efficient. This is particularly critical across healthcare domains, where high-quality, ground-truth labels are often expensive and labor-intensive to obtain.

To this end, our results suggest that scaled pretraining may be particularly valuable for applications involving heterogeneous and weakly expressed phenotypes, such as mental health (e.g. Depression/Anxiety, PHQ-8 score, etc.). Mental health conditions involve a blend of subjective and objective indicators \citep{newson2020heterogeneity}, present with diverse clinical manifestations ~\citep{mclean2011gender, hwang2008conceptual, kivimaki2020association}, and are governed by complex temporal dynamics \citep{nelson2017moving}.
By pretraining on a large corpus with broad variation in daily routines, \modelname may better marginalize ``nuisance'' variation and retain latent physiological information that generalizes across diverse populations. However, we caution that population-level models are distinct from individual-level forecasting; future research should employ longitudinal personalized modeling to better evaluate within-person changes in conditions and health over time.

\subsection{Accelerating Model Development}
Flexible and expressive pretrained embeddings can dramatically increase how efficiently we can develop new downstream models. The few-shot performance of \modelname speaks to the generalizability and efficient adaptability of the representation learned through scaled pretraining.
However, the limited volume of data with high-quality labels continues to pose a bottleneck on downstream task performance.
With such sparse annotations of downstream health outcomes, significant engineering may be required to \emph{optimally} adapt the general \modelname embeddings to each endpoint.
Towards this, the capabilities of AI language models to reason over problems and write code presents a promising approach to expedite model development, and partially bridge the limitations of data sparsity.
In this work we establish that an agent-driven framework for iterative solution discovery is able to  efficiently adapt general embeddings to diverse health domains showing sweeping benefits across discriminative health tasks as compared to a linear probe. Agent driven research and scientific discovery is a rapidly evolving field~\citep{Gottweis2026, aygun2025ai} and future work should continue to explore the extent to which these methods may be exploited to provide benefit to the sparsely labeled data found in healthcare.

\subsection{Implications for Digital Health and Clinical Monitoring}
The efficacy of \modelname carries significant implications for the future of digital health. Our results support a transition from task-specific wearable applications toward a general-purpose interface for continuous health monitoring. 
\modelname achieves robust predictive accuracy without requiring complex, task-specific architectures; applying simple linear probes to the pretrained embeddings is sufficient for a wide variety of downstream applications.
By providing a common substrate for various health models, pretrained representations eliminate the need for bespoke, complex, and end-to-end machine learning pipelines for every individual outcome, streamlining the deployment of predictive analytics in digital health.

Furthermore, while traditional healthcare relies on episodic, "snapshot" measurements captured during clinical visits or in laboratory settings, wearable sensors provide dense, longitudinal observations of physiology and behavior in free-living conditions. In this context, a generalist model for wearable health may be useful in identifying individuals who would benefit from confirmatory testing or early intervention~\citep{perez2019large, lubitz2022detection}. This is especially valuable for conditions that remain asymptomatic until advanced stages \citep{yang2022artificial, liu2022monitoring}. We emphasize, however, that these predictions are intended for screening, risk stratification, and longitudinal tracking rather than as a definitive replacement for clinical diagnosis.

\subsection{Towards More Grounded Question Answering for Personal Health}
Recent works have emphasized the scale on which people leverage AI systems for medical question answering~\citep{costa2026public}, and the efficacy with which these systems are able to provide feedback to queries~\citep{mcduff2025towards, breda2026symptomai}. Wearable health data offers the opportunity to improve the quality of responses to user queries by grounding answers in measures of their own sensed physiology and behavior. Our experiments on the use of \modelname as tool by a Personal Health Agent demonstrate that incorporating \modelname predictions into the agent's context results 
in statistically significant improvements over baseline systems that lack this extra context. Notably, despite the inherent imperfections of model-derived predictions compared to the clinical ground truth, we observed no statistically significant difference in the quality of the resulting agentic interactions. Looking forward, systems like ours, which pair AI coaches with additional grounding, may enable the delivery of more personalized, proactive, and accessible guidance, bridging the gap between continuous health sensing and sporadic clinical consultations.

\subsection{Limitations and Future Work}
This study has several limitations that we acknowledge.
\textit{First}, consumer wearable devices are heterogeneous in both hardware and signal processing, and there is limited standardization in how measurements are derived across platforms. Although \modelname was trained and evaluated on data from multiple Fitbit and Pixel Watch devices, transfer to other device ecosystems is not guaranteed and would likely require additional adaptation.

\textit{Second}, to support large-scale modeling on consumer devices, our input representation uses one-minute aggregated features rather than raw sensor waveforms. This allows the model to capture long-range daily and circadian context, but it necessarily discards finer temporal structure, such as beat-to-beat cardiac variability and sub-second motion patterns, that may be informative for some clinical tasks. As such, more work is needed to understand the optimal settings (sampling frequency, temporal context) for different health outcomes. We additionally note that more granular signals are also not necessarily stored due to considerations such as storage footprints, and power constraints.

\textit{Third}, collecting large and reliable downstream labels paired with wearable data remains difficult. While some of the outcomes studied here are lab tests, a sizeable portion rely on self-reported diagnoses, medication use or screening questionnaires. Converting continuous screening measures into binary outcomes may introduce threshold-dependent noise. Future work should therefore evaluate these models against clinically verified outcomes, including electronic health record (EHR) labels and gold-standard physiological measurements.

\textit{Fourth}, our evaluation of the \modelname-augmented health agent was constrained to a static, single-turn interaction paradigm. Real-world clinical consultations and AI agent interactions are inherently multi-turn, allowing users or physicians to ask follow-up questions to clarify ambiguities or gather missing context. In our experimental design, physician evaluators were restricted solely to the static data presented to them. Furthermore, the downstream datasets utilized inherently lack comprehensive ground-truth labels across all health dimensions (e.g., missing mental health or sleep targets). While this missingness highlights a primary strength of our method, predicting unknown labels, it also means evaluators had limited holistic patient context. Although restricting physicians from asking follow-up questions was an intentional methodological choice to tightly control variables and prevent evaluation bias, it does limit the direct generalizability of these findings to dynamic, conversational real-world deployments.

\textit{Finally}, both the pretraining and downstream evaluation populations reflect the characteristics of Fitbit and Pixel Watch users and therefore may not be fully representative of the broader United States or global population. Although the pretraining cohort spans a broad range of age, sex and BMI, the population of wearable users do not necessarily represent the United States population distribution, such as having a greater proportion of female users than the population ratio. Reported performance may therefore not generalize directly to the overall population and future evaluation will require more targeted recruitment and subgroup-specific validation.

\section{Conclusion}

The ubiquity of wearable health monitors offers the opportunity to improve access to personal health insights and in so doing drive preventative care. To this end, we present \modelname, a foundation model for wearable health which generalizes across diverse health domains. While data labeled with health outcome annotations are rare, our method learns directly from over one trillion minutes of unlabeled multimodal wearable sensor data, producing a robust representation of sensed physiology with predictive benefits on tasks spanning cardiovascular, metabolic, and mental health, sleep, lifestyle factors, and physiologically-relevant demographics. Through our analysis we establish the utility of this scaled pretraining, and further investigate the label efficiency, generative capabilities, and latent structure of \modelname. To enable more efficient adaptation of the \modelname embeddings to diverse downstream tasks, we present an agentic code generation framework that iteratively explores each solution space. Finally, to understand the end-to-end utility of \modelname we assess the benefit of providing \modelname predictions to a Personal Health Agent. In summary, our work represents a shift towards more generalist models for wearable health, which enable more flexible prediction across health domains, more grounded downstream reasoning, and opportunities to provide users with tangible insights about their own health.

%% file: 4a-methods_data.tex
\section{Sensor Signals for Wearable Foundation Models}
\label{sec:data_construction}

The Fitbit Sense 2 and Pixel Watch 2 utilize five sensors of primary relevance to this study: Photoplethysmography (PPG), skin temperature, accelerometer, electrodermal activity (EDA), and an altimeter. From these raw inputs, we derive a set of 34 aggregate signals (features), detailed in Table~\ref{tab:features}. To optimize device battery life and storage, raw sensor data is not retained; instead, we rely on one-minute aggregate signals. We tested whether these signals were strongly co-linear before proceeding with model training. Figure~\ref{fig:correlation_heatmap} of Appendix~\ref{sec:dataset_additional_details} shows no pair of signals has a correlation greater than 0.8.

\CardiovascBadge \textbf{Cardiovascular.} Heart rate (HR) is extracted from the PPG signal at 1 Hz using a validated algorithm~\citep{nissen2022heart}. Per-minute HR is computed as the mean of the instantaneous heart rate across non-overlapping one-minute windows. An on-device peak detection algorithm identifies R-wave peaks to calculate RR intervals. To mitigate noise, standard HRV metrics are computed using robust statistical methods. We derive the median RR interval, the Shannon entropy of the RR intervals (ShEnRR), and the Coherence of breathing frequency to heart rate. Time-domain variability metrics—standard deviation of RR intervals (SDNN) and root mean squared of successive differences (RMSSD)—are calculated using RR intervals between the 5\textsuperscript{th} and 95\textsuperscript{th} percentiles to exclude outliers. We also compute pNN20, the percentage of successive RR intervals differing by more than 20ms.In the frequency domain, we extract the power in the Very Low (VLF), Low (LF), and High (HF) frequency bands, the LF/HF ratio, and the Shannon entropy of the power spectrum (SpectralEn). Finally, the "Valid RR" metric quantifies the percentage of the 5-minute window containing valid RR intervals.

\CardiopulmBadge \textbf{Cardiopulmonary.} We derive two features related to blood oxygen saturation. SpO2 represents the blood oxygen saturation level, while SpO2 Confidence provides the confidence level of the reading. SpO2 is computed exclusively during stationary periods; raw sensor data is filtered to isolate segments where accelerometer variance is low. The PPG waveform is processed by a convolutional neural network to extract features, which are subsequently classified via a fully connected layer. We also track SpO2 Coverage, defined as the percentage of the minute with valid SpO2 data.

\SleepBadge \textbf{Sleep.} Sleep metrics are inferred using a validated multi-modal algorithm that fuses accelerometer-derived actigraphy with PPG-derived heart rate and HRV data. The model classifies sleep epochs into four primary stages: Awake, Light, Deep, and Rapid Eye Movement (REM). For each one-minute window, we record the time spent in each of these stages in seconds.

\MotionBadge \textbf{Motion.} We extract ten features from the 3-axis accelerometer to characterize motion and physical activity. These include the step count, Axis Mean (mean of the 3-axis data), and Kurtosis (of the 3-axis root mean squared magnitude). Complex signal features include Jerk Autocorrelation (ratio of lag-1 autocorrelation to energy), Log Energy, and Log Energy Ratio. We calculate the Covariance as an estimate of the condition number for the 3-axis covariance. We also examine the zero-crossings of the 1st 3-axis principal component, extracting both the Zero Crossing Average (mean time between crossings) and Zero Crossing St.Dev. (standard deviation of time between crossings). Additionally, we derive a "Sleep Coefficient" metadata feature, calculated as the sum of the 3-axis max-min range within 16 log-scaled bins.From the barometer, we compute the Altitude St.Dev., which represents the standard deviation of altimeter readings (in hPa) to isolate vertical displacement from atmospheric drift.

\SkinBadge \textbf{Skin Surface.} The device measures skin conductance and temperature to infer physiological states. The Electro-Dermal Activity (EDA) sensor measures Skin Conductance Level (SCL), which correlates with sympathetic nervous system arousal. We derive the "Conductance" feature, defined as the center of the linear tonic SCL value fit (in $\mu$Siemens). Lead Contact Counts are recorded to track the number of times sensor leads contact the wrist. Concurrently, a temperature sensor on the wrist-facing surface samples skin temperature. We report the "Temperature" feature as the mean value of skin temperature (in \degree C) for the minute.

\input{Tables_Final/tbED1_sensor_features}

\begin{figure*}[h]
    \centering
    \includegraphics[width=\textwidth]{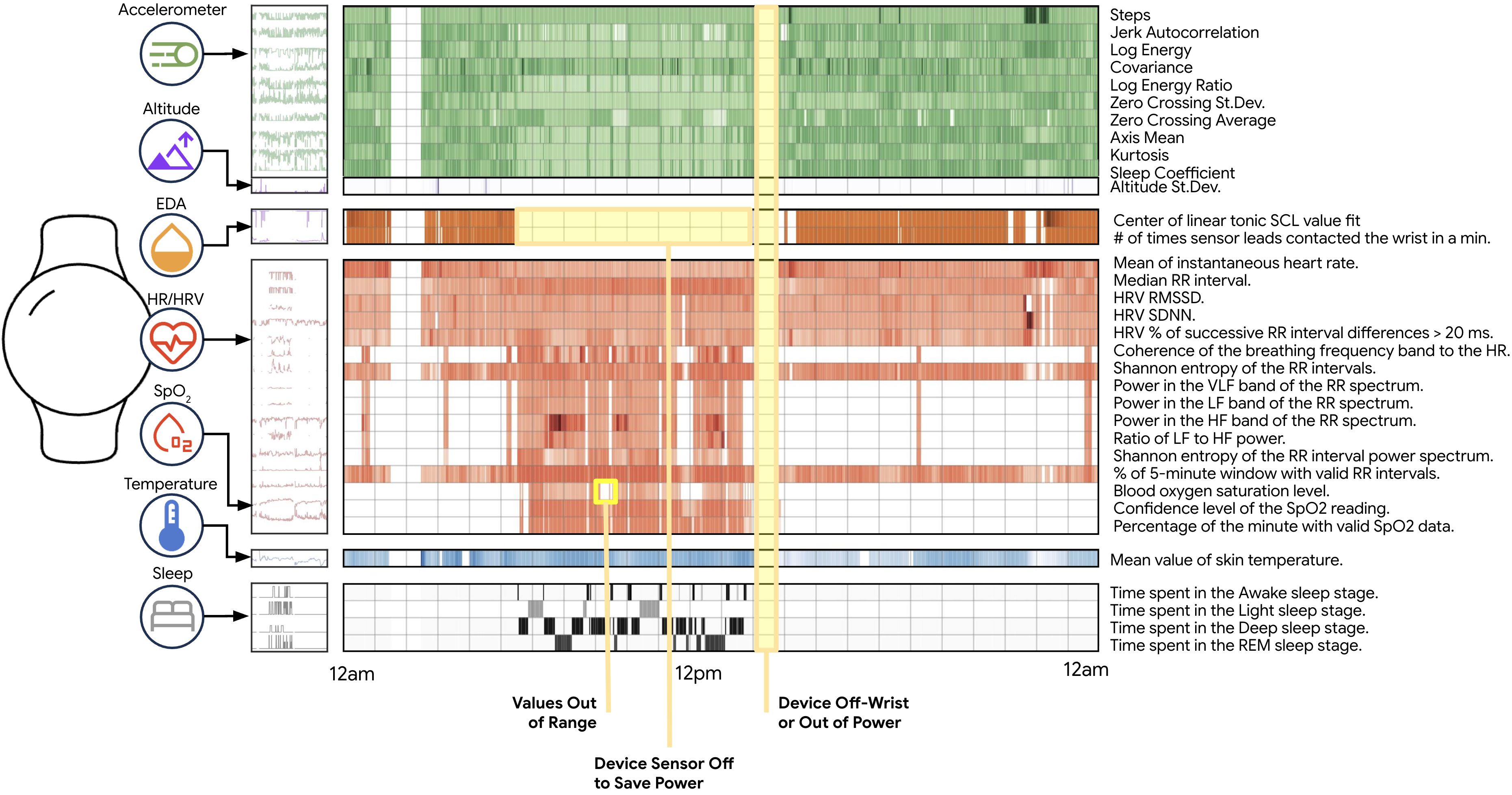}
    \caption{\textbf{\modelname Input Data.} The model ingests 34 one-minute aggregate sensor features derived from five sensor modalities (PPG, Accelerometer, EDA, Skin Temperature, and Altimeter) organized into seven categories (Accelerometer, Altitude, EDA, HR/HRV, SpO\textsubscript{2}, Temperature, Sleep) over a 24-hour context window, processed through outlier removal and normalization before encoding. Note that these features are derived from a set of five sensors as described in Appendix~\ref{sec:dataset_additional_details}. This figure also highlights various modes of data missingness including out of range values, sensor power cycling, and human behaviors.}
    \label{fig:missing_modes} 
\end{figure*}

\begin{figure*}[h]
    \centering
    \includegraphics[width=\textwidth]{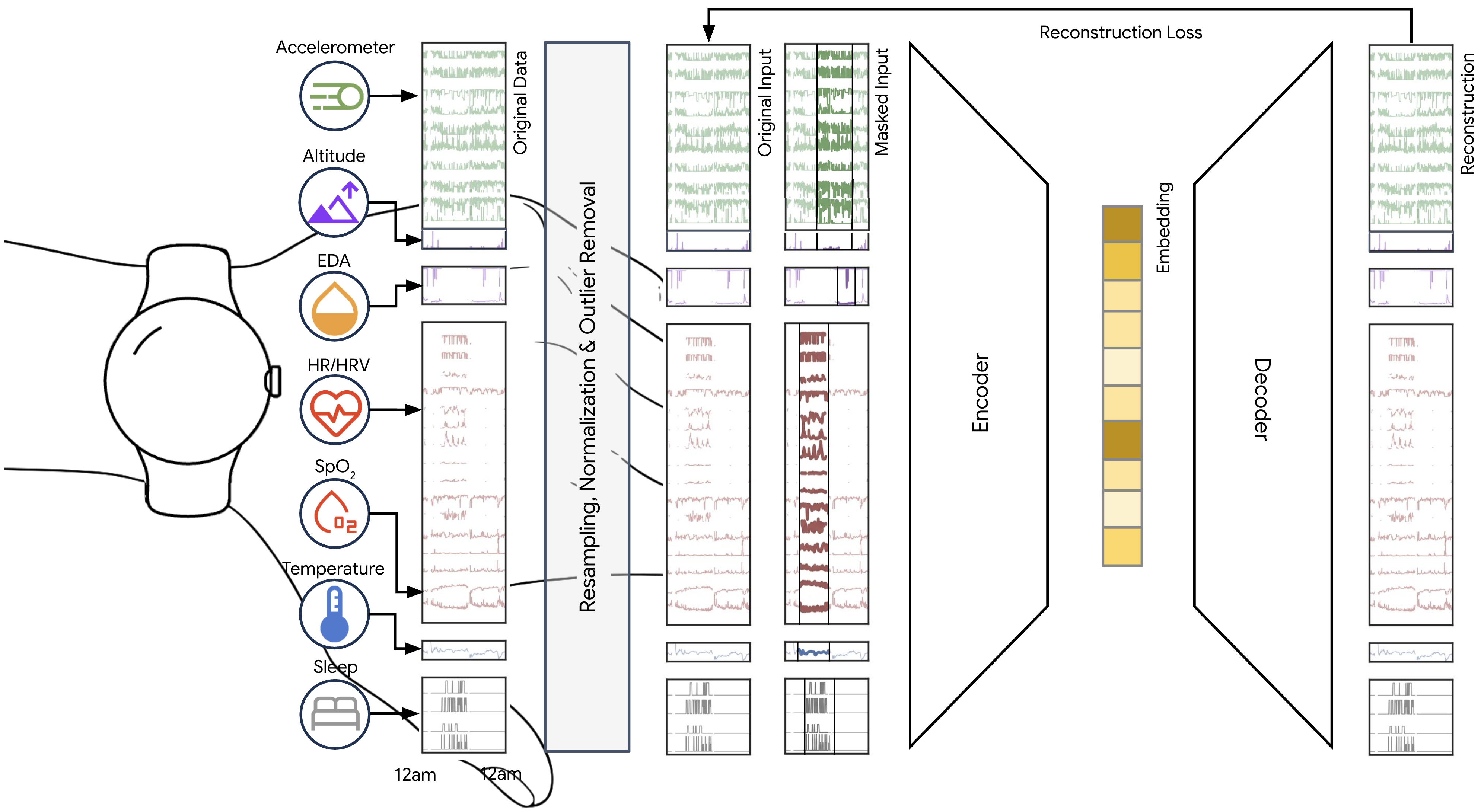}
    \caption{\textbf{Pretraining Pipeline.} Original multimodal sensor inputs are resampled and normalized, then artificially masked before being passed to the encoder. The decoder reconstructs the masked patches, with MSE loss computed on the artificially masked regions.}
    \label{fig:pretraining} 
\end{figure*}

\section{Datasets}
\label{sec:datasets}

\subsection{Pretraining Dataset}
\label{subsec:pretraining_datasets}

\begin{figure*}[!htbp]
    \centering
    \includegraphics[width=\textwidth]{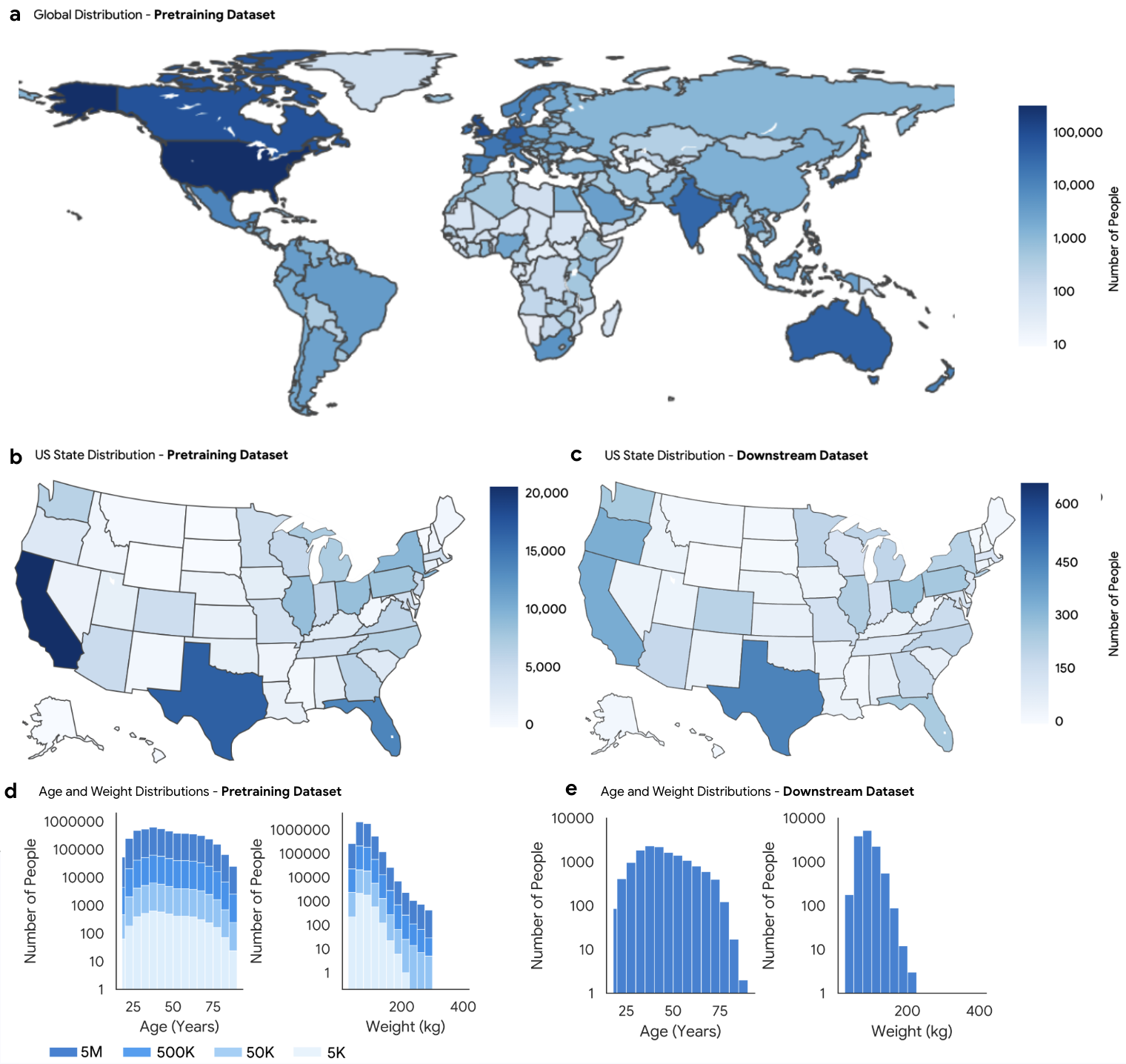}      
    \caption{\textbf{Global Demographic Distributions} Geographic and demographic distribution of pretraining and downstream datasets. \textbf{(a)} Global distribution of pretraining participants across countries. \textbf{(b)} US state-level distribution of pretraining data. \textbf{(c)} US state-level distribution of downstream study participants. \textbf{(d)} Age and weight distributions in the pretraining cohort, \textbf{(e)} age and weight distributions in the downstream cohort. Note the log scale of the Y axis for (d) and (e).}
    \label{fig:global_demographic_distributions} 
\end{figure*}

To build the large dataset for our experiments we sampled wearable data from 5 million participants during the period September 1$^{st}$ 2024 to September 1$^{st}$ 2025. As participants could opt-in using any Fitbit or Pixel Watch device, the dataset contains data from a wide range of models released between 2012 to 2024 (see Table~\ref{tab:device_stats_year}).
The most common devices were Fitbit Inspire 3, Fitbit Charge 6, Fitbit Versa 2, 3 and 4, Fitbit Sense and Pixel Watch 1, 2, 3. Participants provided voluntary consent for their de-identified data to be used for research and development of new health and wellness products and services and we obtained a secondary research exemption determination from a centralized IRB for this research (WCG 20253840). We sub-selected from people wearing one of these devices as older device generations included fewer sensors. The subjects were asked for self-reported sex, age and weight. Table~\ref{tab:overall_demographics} summarizes the characteristics of the pretraining data and Figure~\ref{fig:global_demographic_distributions} shows the distribution by age and body-mass index (BMI). All data were de-identified and not linked with any other information. 

To create a dataset that maximized the number of subjects we randomly sampled 10-weeks of data from each of the 5 million subjects. From this set of up to 350 million human-days, we construct a pretraining dataset encompassing over 2 billion hours of multimodal sensor data. In total our pretrain set contains over \emph{one trillion minutes} of minute-resolution data observed from a suite of five wearable sensors. In preparing our data for modeling, global normalization parameters (mean and standard deviation) for each of the sensor signals were computed on this pretraining dataset. These parameters were used to normalize (z-score) the pretraining and downstream data (Figure~\ref{fig:pretraining}). In addition to this pretraining set, one week of data from an independent 10,000 subjects were used for validation, and one week of data from another independent set of 10,000 subjects were used for test.

\subsection{Downstream Datasets}
\label{subsec:downstream_datasets}

We compiled an extensive list of downstream tasks by combining de-identified data from multiple prospective IRB-approved observational studies as described below. 
In total our downstream set contains data from $13,985$ subjects sourced from three prospective studies spanning Metabolic, Cardiac and Respiratory Health ($N = 1,655$), Sleep ($N = 6,377$), and Mental Health ($N = 5,953$). These studies and the collected data are described below and definitions of the downstream task labels are provided in Table~\ref{tab:downstream_labels}.

\subsubsection{Metabolic, Cardiac and Respiratory Health}
\label{metabolic_data}
We designed a prospective observational study and recruited adult participants from the United States ~\cite{metwally2026insulin}. The study was approved by Advarra (Pro00074093). We enrolled 4,416 participants, of which $N = 1,655$ had complete data (at least 14 days of data wearable data, lab results, and completed demographics) and were included in our analysis. 
The study, which spanned a maximum of 70 days, was designed to gauge the feasibility of leveraging wrist-worn wearable devices to develop algorithms for assessing metabolic health, cardiovascular health, respiratory health, deriving to biological age, and regressing to blood biomarkers. Specifically, participants completed surveys, obtained a one-time blood test, and were asked to continuously wear their wearable device for the duration of the study.
Table~\ref{tab:downstream_labels} shows the numbers of people with confirmed responses for each of the data types (including lab reports and self-reported health history.

\textbf{Study Population.} The study was limited to Fitbit users with a heart-rate sensing capable device, living in the US, and aged 21 - 80. These users were required to have at least three months prior data, with use on at least 75\% of these days.

\textbf{Self-Report Demographics, Biometrics, and Medical History Questionnaires.} Participants completed surveys reporting their demographic data (age, sex, ethnicity, weight, height), blood pressure, waist circumference, medications, diagnosed conditions, health habits and management strategies. All collected data are listed in the Methods section.

\textbf{Blood Panels.} Participants obtained blood testing early in the morning from Quest Diagnostics after fasting for at least 8 hours in order to minimize the effect of solar diurnal cycle. Tested blood biomarkers included insulin, HbA1c, Comprehensive Metabolic Panel (including glucose), lipids (total cholesterol, triglycerides, HDL, LDL), Complete Blood Count, hs-CRP, GGT, and total testosterone. Test results were also used to calculate a Framingham Risk Score for 10-year prediction of cardiovascular event risk ~\citep{lloyd2004framingham}.

\textbf{Wearable Data.} Data was longitudinally collected from participants' wearable devices for the duration of the study, and up to three months of wearable data prior to study enrollment was also linked.

\textbf{Compensation.} Participants received the results of their blood draw free of charge.

\subsubsection{Sleep}
\label{sleep_data}
We designed a prospective observational study and recruited over 10,000 participants from all 50 states of the United States between August 2023 and January 2024. From this cohort we subselect participants with labels of interest ($N=6,377$). The study was approved by a centralized IRB (Advarra) (Pro00069849). The study was designed to support improvements to the Fitbit Sleep Score and to better understand the relationship between sleep, next day sleep, and health outcomes (e.g., alertness, mood, etc.). An additional aim was to leverage the data to develop models of sleep and health (e.g. circadian rhythm), enabling users to better understand their sleep and allowing for more personalized recommendations. The study duration was 15 days (14 nights) per participant during which time participants were asked to continuously wear their Fitbit except during charging. Additionally, up to one month of prior Fitbit data was included in the study data.

\textbf{Study Population.} The study was limited to current Fitbit users whose devices included a heart rate sensor and who owned an Android smartphone capable of installing the Google Health Studies application. Participants were further required to be age 18 - 88, and located in the United States.

\textbf{Baseline and Post-Study Questionnaires.} Participants completed a battery of self-report questionnaires at baseline including demographics, Sleep Habits, Health Habits, Sleep Environment, the National Sleep Foundation's Sleep Satisfaction Tool~\citep{ohayon2019national}, and the Morningness-Eveningness Survey~\citep{horne1976self}.

\textbf{Morning and Evening Survey.} Each morning, participants were asked to complete a 5-item survey reporting on their sleep the previous night. Each evening, participants were asked to report on that day's activities including food intake, exercise, and other activities.

\textbf{Alertness and Mood Surveys.} Up to three times daily, participants were asked to complete Ecological Momentary Assessment (EMA) questions on their alertness and mood. Surveys were timed for mid-morning, afternoon, and evening, and were personalized for each participants' approximate sleep schedule.

\textbf{Alertness Tasks.} Up to three times daily, participants were asked to complete two standardized tasks meant to gauge alertness. These included a three-minute reaction time task where participants would tap their smartphone screen in response to some stimuli, and a five-minute gaze task where participants' gaze was tracked as stimuli were presented on screen.

\textbf{Compensation.} Subjects were compensated with a \$25 Google Merchandise Purchase Code if they attempted at least 12 of 15 (~80\%) of the days/nights of the protocol.

\subsubsection{Mental Health}
\label{mental_health_data}
We designed a prospective, observational study and recruited 7,500 Fitbit users in the United States~\citep{mcduff2024google, winbush2025smartphone}. From this cohort we subselect participants with labels of interest ($N=5,953$).
The study was approved by the IRB of the University of Oregon (MOD00000379). The study was designed to investigate patterns and relationships between digital device use, sensor based measures (including both behavioral and physiological signals), and self-reported measures of mental health and well-being. The study duration was four-weeks long per participant and included a wearable for the complete four-week period. The study and recruitment were designed to increase participation of under-represented groups, as defined by race and ethnicity (e.g., Caucasian, African American, Asian, Latina/Latino, Native Americans/ Indigenous Populations), biological sex at birth (Female; Male), age (18 - 40; over 40), sexual orientation/gender identity (Heterosexual; LGBTQIA+). 

\textbf{Study Population.} The study was limited to participants aged 18-80 with an Android smartphone capable of installing the Google Health Studies application. Participants were further required to be free of major health conditions which severely restricted mobility and physical activity. A subset of this group, who owned a Fitbit device, were invited to share their Fitbit data.

\textbf{Baseline and Post-Study Questionnaires.} Participants completed a battery of self-report questionnaires at baseline and study conclusion, as follows: (Baseline only) Demographics questionnaire, Sleep and Health Habits questionnaire, Patient Health Questionnaire (PHQ-8)~\citep{kroenke2009phq}, Generalized Anxiety Disorder Scale (GAD-7)~\citep{spitzer2006brief, lowe2008validation}, Patient Reported Outcomes Measurement Information System (PROMIS) Sleep Disturbance \& Sleep Related Impairment short form~\citep{cella2010patient,yu2012development}, Shortened Smartphone Addiction Scale (SAS)~\citep{kwon2013smartphone}, Perceived Stress Scale (PSS)~\citep{cohen1983global}. 

\textbf{Ecological Momentary Assessment Surveys.}
During the first and last week of the study, participants were asked to fill out three EMA surveys spread throughout the day. Each EMA assessed the participant's mood across five affects (\textit{happy}, \textit{calm}, \textit{anxious}, \textit{sad}, and \textit{stressed}), each assessed on a single-select 5-item Likert scale, spanning \textit{not at all} to \textit{very}. Participants were also asked to report who they had spent the most time with (since the previous EMA). Options included \textit{Alone}, \textit{Friends}, \textit{Family}, \textit{Spouse/Partner}, \textit{Co-workers}, \textit{Co-students}.

\textbf{Daily Status Reports.}
Each morning, throughout the four-week period, participants were prompted to report how they had been feeling over the past day. Participants responded via a single-select 5-item Likert scale spanning \textit{very bad} to \textit{very good}.

\textbf{Mobile and Wearable Data.} Data was longitudinally collected from participants' devices over the course of the four-week study. Mobile phone metrics included screen on time, usage time by application category, battery status, and number of phone unlocks. Aggregated measures of geolocation were also collected, binning the participants locations at either being at \textit{home}, \textit{work}, or \textit{other}. For a subset of participants, wearable sensor data was also logged continuously. Note that sensor data, described in this work, refers to the data from these wearable devices rather than from mobile phones.

\textbf{Compensation.} Subjects were entered into a raffle to win a \$50 gift card with an 11.5\% win probability. The conditions for eligibility were to: 1) consent and enable sensor collection at study start, 2) complete the pre-study assessments, 3) complete daily status assessments for a minimum of seven study days (one week, cumulatively), and 4) to complete the post-study assessments.

\subsubsection{Downstream Data Statistics}
\begin{figure*}[!htbp]
    \centering
    \includegraphics[width=\textwidth]{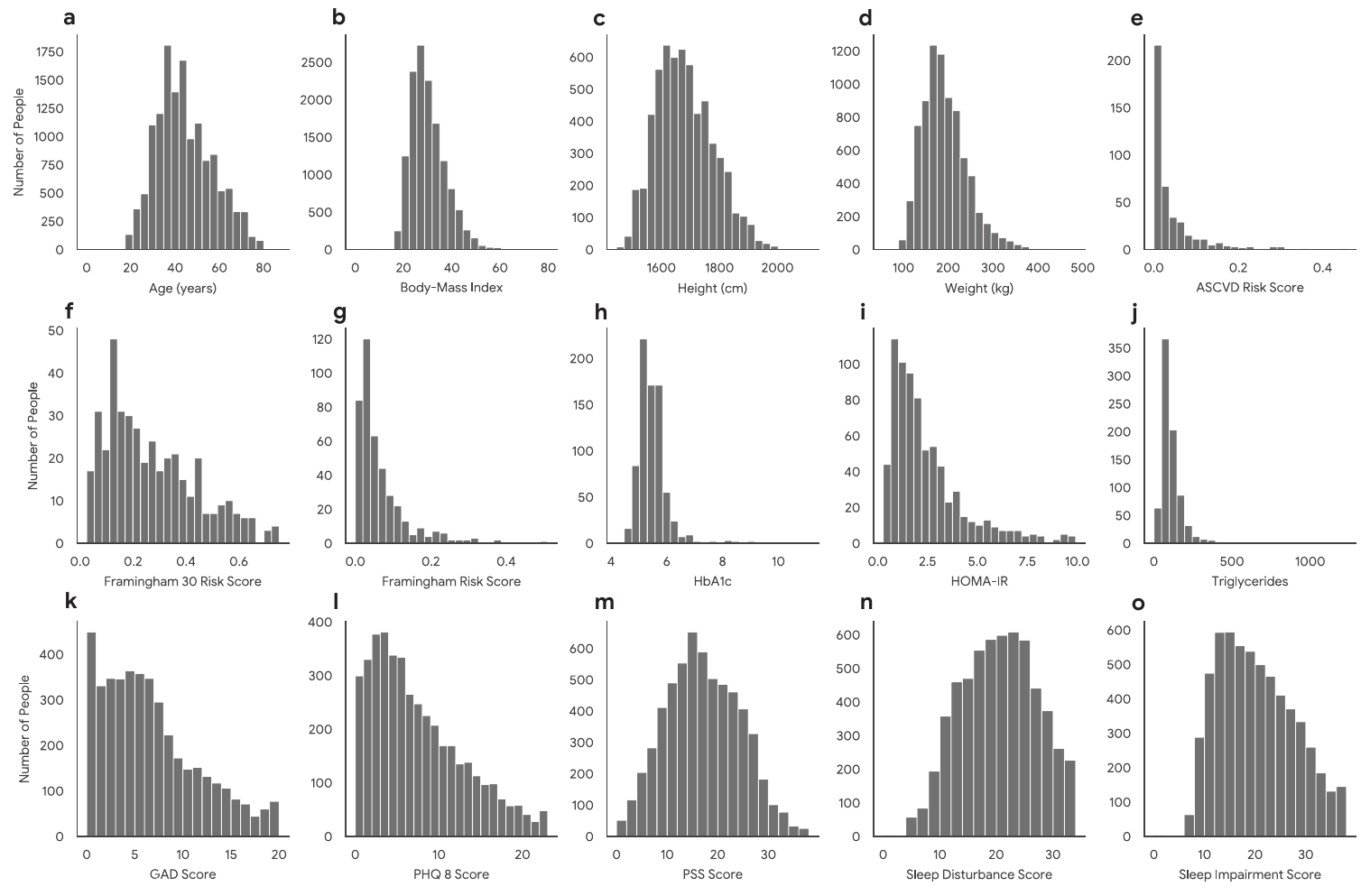}
    \includegraphics[width=\textwidth]{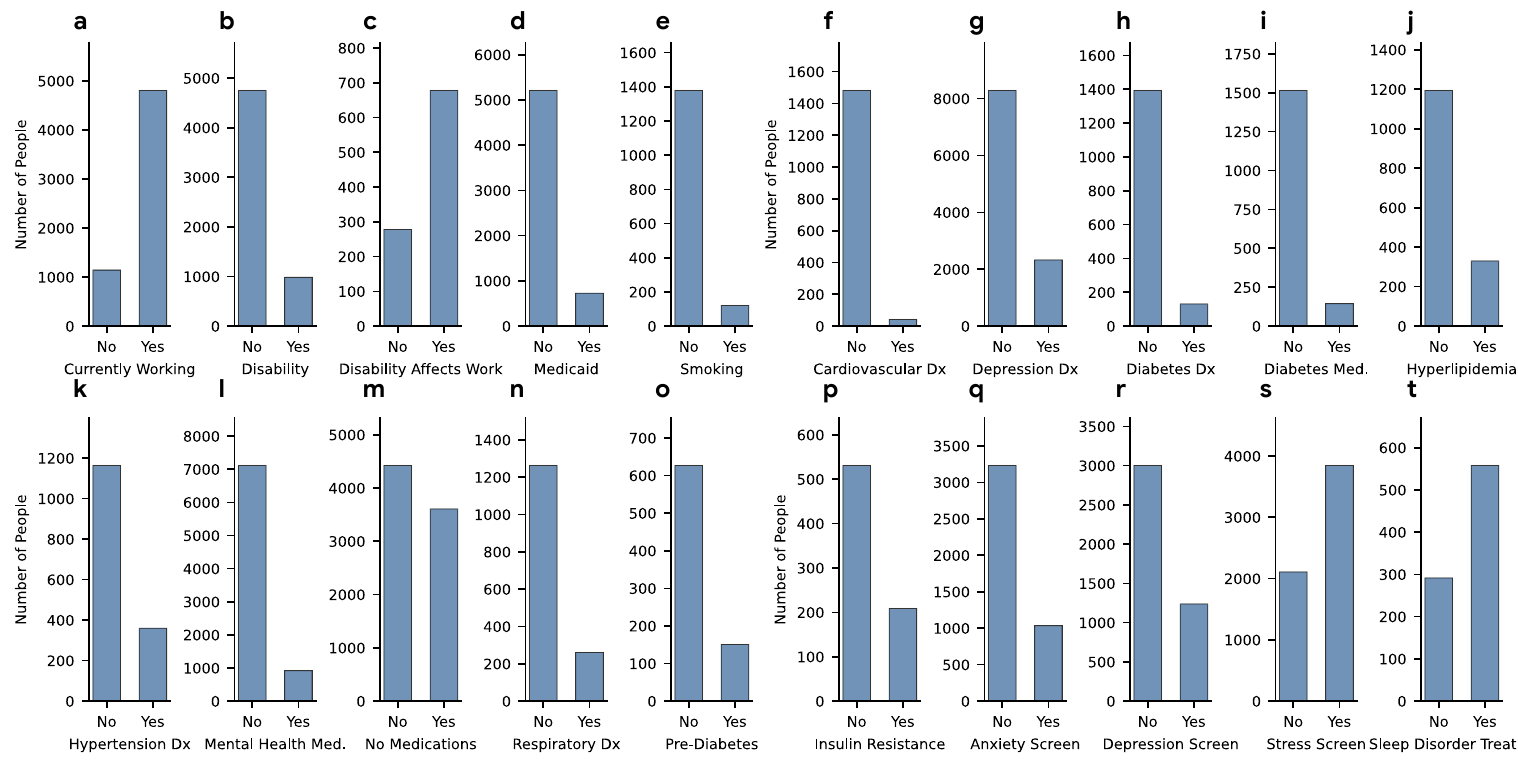}
    \caption{\textbf{Task Label Statistics.} Distributions of regression (top) and classification (bottom) task labels across the 35 discriminative downstream tasks derived from multiple prospective studies.}
    \label{fig:dataset_statistics} 
\end{figure*}

Table~\ref{tab:overall_demographics} shows the counts and distributions of participants who feature in the pretraining and downstream datasets. The pretraining set includes 5,020,000 unique participants and the downstream set 13,985 unique participants. 
We compare these to US Census and CDC statistics and also to the distribution of data in the widely used All of Us dataset. While our datasets represent a broad group from the global (pretraining) and US (downstream) populace there are some areas in which representation is still lacking, pretraining data is skewed towards women who are more frequent adopters of Fitbit devices and White/Caucasians.  Figure~\ref{fig:dataset_statistics} shows the geographic distributions of the country or US state which these participants registered as their home state.  The pretraining data was sampled globally with representation from over 100 countries.  The downstream data was limited only to US participants with representation from all 50 states. Note that the age and weight distributions appear similar between pretraining and downstream datasets. 

\input{Tables_Final/tbED2_data_demographics}

\input{Tables_Final/tbED3_data_downstream_tasks}

%% file: Tables_Final/tbED1_sensor_features.tex
\begin{table}[!htbp]
\small
    \centering
    \caption{\textbf{Sensor Feature Definitions and the Sensor they are Derived From.}}
    \resizebox{1\linewidth}{!}{%
    \begin{tabular}{rllp{12cm}}
    \toprule
        \textbf{Feature} & \textbf{Sensor} & \textbf{Unit} & \textbf{Definition} \\
        \midrule
        \grayrow
        \multicolumn{1}{l}{\CardiovascBadge \textbf{Cardiovascular}} & & & \\
        Heart Rate & PPG & Beats/Min & Mean of instantaneous heart rate. \\ 
        RR Median & PPG & Msec & Median RR interval. \\ 
        RMSSD 05-95 & PPG & Msec & RMSSD calculated using RR intervals between the $5^{th}$ and $95^{th}$ percentile. \\
        SDNN 05-95 & PPG & Msec & SDNN calculated using RR intervals between the $5^{th}$ and $95^{th}$ percentile. \\
        pNN20 & PPG & \% & Percentage of successive RR interval differences greater than 20 ms. \\
        Coherence & PPG & a.u. & Coherence of the breathing frequency band to the heart rate. \\
        ShEnRR & PPG & Nats & Shannon entropy of the RR intervals. \\
        VLF & PPG & $\text{Msec}^2$ & Power in the Very Low Frequency band (0.003-0.04 Hz) of the RR spectrum. \\
        LF & PPG & $\text{Msec}^2$ & Power in the Low Frequency band (0.04-0.15 Hz) of the RR spectrum. \\
        HF & PPG & $\text{Msec}^2$ & Power in the High Frequency band (0.15-0.4 Hz) of the RR spectrum. \\
        LF/HF & PPG & a.u. & Ratio of LF to HF power. \\
        SpectralEn & PPG & Nats & Shannon entropy of the RR interval power spectrum. \\
        Valid RR & PPG & \% & \% of 5-minute window with valid RR intervals. \\

        \grayrow
        \multicolumn{1}{l}{\CardiopulmBadge \textbf{Cardiopulmonary}} & & & \\
        SpO2 & PPG & \% & Blood oxygen saturation level. \\
        SpO2 Confidence & PPG & a.u. & Confidence level of the SpO2 reading. \\
        SpO2 Coverage & PPG & \% & Percentage of the minute with valid SpO2 data. \\
        \grayrow
        \multicolumn{1}{l}{\SleepBadge \textbf{Sleep}} & & & \\
        Stage Awake & PPG + ACCEL & Seconds & Time spent in the Awake sleep stage. \\
        Stage Light & PPG + ACCEL & Seconds & Time spent in the Light sleep stage. \\
        Stage Deep & PPG + ACCEL & Seconds & Time spent in the Deep sleep stage. \\
        Stage REM & PPG + ACCEL & Seconds & Time spent in the REM sleep stage. \\
        Sleep Coefficient & ACCEL & a.u. & Sum of 3-axis max-min range with 16 log-scaled bins.\\
        \grayrow
        \multicolumn{1}{l}{\MotionBadge \textbf{Motion}} & & & \\
        Steps & ACCEL & Steps & Number of steps. \\
        Jerk Autocorrelation & ACCEL & a.u. & Ratio of lag=1 autocorrelation to energy in 1st 3-axis principal component. \\
        Log Energy & ACCEL & a.u. & Log of sum of 3-axis root mean squared magnitude. \\ 
        Covariance & ACCEL & a.u. & Estimate of condition number for the 3-axis covariance. \\
        Log Energy Ratio & ACCEL & a.u. & Log of ratio of sum of energy in 1st 3-axis principal component over energy of 3-axis root mean squared magnitude.  \\ 
        Zero Crossing St.Dev. & ACCEL & Seconds & Standard deviation of time between zero crossing of 1st 3-axis principal component. \\ 
        Zero Crossing Average & ACCEL & Seconds & Mean of time between zero crossing of 1st 3-axis principal component. \\ 
        Axis Mean & ACCEL & a.u. & Mean of 3-axis\\ 
        Kurtosis & ACCEL & a.u. & Kurtosis of 3-axis root mean squared magnitude.\\
        Altitude St.Dev. & Barometer & hPa & Standard deviation of altimeter readings. \\
        \grayrow
        \multicolumn{1}{l}{\SkinBadge \textbf{Skin Surface}} & & & \\
        Temperature & TEMP & $\degree$ C & Mean value of skin temperature. \\
        Conductance & EDA & $\mu$Siemens & Center of linear tonic SCL value fit. \\
        Lead Contact Counts & EDA & Counts & Number of times sensor leads contacted the wrist in a minute. \\
        \grayrow
        \bottomrule
        \end{tabular}}
    \label{tab:features}
\end{table}

%% file: Tables_Final/tbED2_data_demographics.tex
\begin{table}[!htbp]
  \centering
  \small
  \caption{\textbf{Demographics.} Counts and distributions of pretraining and downstream study populations. n.c. = Data not collected.}
  \label{tab:overall_demographics}
  \begin{tabular}{lrrrr|rr}
    \toprule[1.5pt]
    \rowcolor{white!20}
    \textbf{Category} & \multicolumn{2}{c}{\textbf{Pretraining}} & \multicolumn{2}{c}{\textbf{Downstream}} & \textbf{US Census} & \textbf{All of Us (\%)} \\
    \cmidrule(lr){2-3} \cmidrule(lr){4-5}
    & \multicolumn{1}{c}{\textbf{N}} & \multicolumn{1}{c}{\textbf{\%}} & \multicolumn{1}{c}{\textbf{N}} & \multicolumn{1}{c}{\textbf{\%}} &  \textbf{2023 (\%)} \\
    \midrule
    \rowcolor{gray!20}
    \textbf{Age} & &  &  & & & \\
    18--39 & 1,944,258 & 38.7\% & 5,587 & 40.0\% & 29.7\% & 25.0\%\\
    40--59 & 1,772,338 & 35.3\% &  6,307 & 45.1\% & 25.4\% & 38.7\%\\
    60--79 & 1,083,315 & 21.6\% & 1,914 & 13.7\% & 17.9\% & 32.4\%\\
    80+ & 104,315 & 2.1\% & 19 & 0.1\% & 4.0\% & 3.8\%\\
    Unknown & 115,774 & 2.31\% & 158 & 1.1\% \\    
    
    \rowcolor{gray!20}
    \textbf{Gender} & & & & & & \\
    Male & 2,009,581 & 40.0\% & 4,835 & 34.6\%  & 50.8\%  & 34.0\%\\
    Female & 2,943,605 & 58.6\% & 8,601 & 61.5\% & 49.2\% & 63.8\%\\
    Non-Binary & n.c. & n.c. & 388 & 2.8\% \\
    Unknown & 66,814 & 1.33\% & 387 & 1.2\% & \\
    
    \rowcolor{gray!20}
    \textbf{BMI}  & & & & & & \\
    Underweight ($<$18.5)  & 231,866 & 4.6\% &  162 & 1.2\% & 1.6\%$^\dagger$ & 3.7\%\\
    Healthy (18.5-25)  & 1,776,477 & 35.4\% & 3,303 & 23.6\% & 27.0\%$^\dagger$ & 39.1\%\\
    Overweight (25--30)  & 1,532,795 & 30.5\% & 4,114 & 29.4\% & 31.1\%$^\dagger$  & 22.1\%\\
    Obese ($\ge$30)  & 1,400,870 & 27.9\% & 6,062 & 43.4\% & 40.3\%$^\dagger$ & 35.1\%\\
    Unknown  & 77,992 & 1.6\%  & 342 & 2.4\% \\

    \rowcolor{gray!20}
    \textbf{Height}  & & & & & & \\
    $<$170cm  & 2,647,970 & 52.8\% & 7,112 & 50.9\% & 56\%$^{\dagger\dagger}$ & 62.8\%   \\
    170--183cm  & 1,966,545 & 39.2\% & 5,561 & 39.8\% & 37\%$^{\dagger\dagger}$  & 31.5\%  \\
    $>$183cm  & 334,550 & 6.7\% & 1,091 & 7.8\% & 7\%$^{\dagger\dagger}$ & 5.7\%  \\
    Unknown  & 70,935 & 1.4\% & 219  & 1.6\%  &\\
   
    \rowcolor{gray!20}
    \textbf{Weight}  & & & & & & \\
    $<$63kg  & 1,186,701 & 23.6\% & 1,660 & 11.9\% & 11.9\% & 25.6\%\\
    63--86kg  & 2,262,533 & 45.1\% & 5,681 & 40.7\% & 41\% & 24.8\%\\
    $>$86kg  & 1,498,675 & 29.9\% & 6,302 & 45.1\% & 42\% & 49.6\%\\
    Unknown  & 72,091 & 1.4\% &  315 & 2.3\% \\
    \rowcolor{gray!20}
    \textbf{Ethnicity}  & & & & & & \\
    White/Caucasian & n.c. & & 11,030 & 78.9\% &  57.8\% & 54.7\%\\
    Black  & n.c. & & 563 & 4.0\% & 18.7\% & 15.4\%\\
    Hispanic  & n.c. & & 773 & 5.5\% & 12.1\% & 14.4\%\\
    Asian  & n.c. & & 582 & 4.2\% & 5.9\% & 3.45\%\\
    Native  & n.c. & & 88 & 0.6\% & 0.7\% \\
    Mixed Race  & n.c. & & 60 & 0.4\%  & 4.1\% & 7.4\%\\
    Unknown  & n.c. & & 889 & 6.4\% & 0.7\% & 0.5\%\\

    \midrule
    \textbf{Total} & 5,020,000 & & 13,985  \\
    \bottomrule[1.5pt]
  \end{tabular}
  \footnotesize \\
  $^\dagger$ Source: CDC National Center for Health Statistics (NHANES 2021–2023 \& 2017–2018). 
  \newline
  $^{\dagger\dagger}$ Source: CDC Anthropometric Reference Data for Children and Adults (NHANES 2015–2018).
\end{table}

%% file: Tables_Final/tbED3_data_downstream_tasks.tex
\begingroup
\renewcommand{\arraystretch}{1.1}
\begin{table}[!htbp]
\footnotesize
    \centering
    \caption{\textbf{Downstream Tasks.} Summary and statistics of downstream data labels. For task type \RegressionLabel is regression, \Binary is binary classification. For data source \SelfSurveyLabel is self-reported, \LabTestLabel is lab tested, \ScreenerLabel is standardized screener survey.}
    \begin{tabular}{rp{5.3cm}llcc}
    \toprule
        \textbf{Category} & \textbf{Definition} & \textbf{Type} & \textbf{Source} & \textbf{N (Test)} & \textbf{\% Pos. or Mean} \\

        \midrule

        \grayrow
        \multicolumn{1}{l}{\textbf{Demographics}} & & & & &\\
        Age & Chronological age in years. & \RegressionLabel & \SelfSurveyLabel & 13,831 & 44.3 years \\
        BMI & Body-mass index in kg/m$^{2}$. & \RegressionLabel & \SelfSurveyLabel & 13,660 & 30.2 kg/m$^2$ \\
        Height & Height in centimeters. & \RegressionLabel & \SelfSurveyLabel & 13,768 & 1.68 meters  \\
        Weight & Weight in kilograms. & \RegressionLabel & \SelfSurveyLabel & 13,777 & 85.7 kg
 \\

        \grayrow
        \multicolumn{1}{l}{\textbf{Lifestyle}} & & & & &\\
        Currently Working & Working status. & \Binary & \SelfSurveyLabel & 5,950 & 80.8\%  \\ 
        Disability & Disability status. &  \Binary & \SelfSurveyLabel & 5,737 & 17.2\%  \\
        Disability Affects Work & Disability impacts working & \Binary  & \SelfSurveyLabel & 956 & 70.9\% \\
        Smoking & Smoking behavior. &  \Binary & \SelfSurveyLabel & 1,500 & 8.1\% \\ 
        Medicaid & On Medicaid Insurance. & \Binary & \SelfSurveyLabel & 5,940 & 12.3\%  \\ 
        No Medications & No regular medication use in daily life. & \Binary & \SelfSurveyLabel & 8,034 & 44.9\% \\ 

        \grayrow
        \multicolumn{1}{l}{\textbf{Cardiovascular}} & & & & & \\
        Cardiovascular Dx & Diagnosis of cardiovascular condition & \Binary & \SelfSurveyLabel & 1,524 & 2.3\% \\
        Hypertension Dx & Diagnosis of hypertension & \Binary & \SelfSurveyLabel & 360 & 23.6\% \\ 
        Respiratory Dx & Diagnosis of cardiovascular condition & \Binary & \SelfSurveyLabel & 261 & 17.1\% \\
        ASCVD Risk & 10 year risk atherosclerotic cardiovascular disease & \RegressionLabel & \SelfSurveyLabel & 417 & 0.04  \\
        Framingham Risk & 10 year risk of cardiovascular disease & \RegressionLabel & \SelfSurveyLabel & 417 & 0.06  \\
        Framingham 30 Risk & 30 year risk of cardiovascular disease & \RegressionLabel & \SelfSurveyLabel & 412 & 0.27 \\

        \grayrow
        \multicolumn{1}{l}{\textbf{Metabolics}} & & & & & \\
        Diabetes Dx & Diagnosis of diabetes condition & \Binary & \SelfSurveyLabel & 1,524 & 8.6\% \\
        Diabetes Med & Use of a Diabetes Medication & \Binary & \SelfSurveyLabel & 1,659 & 8.6\% \\
        Hyperlipidemia  & Hyperlipidemia Dx & \Binary & \LabTestLabel & 1,524 & 21.7\% \\
        Pre-Diabetes & Classification of HbA1c $\geq$ 5.7 & \Binary & \LabTestLabel & 778 & 11.1\% \\
        Insulin Resistance & Classification of HOMA-IR $\geq$ 2.9 & \Binary & \LabTestLabel & 779 & 32.2\% \\
        HOMA-IR & Homeostatic Model Assessment for Insulin Resistance score & \RegressionLabel & \LabTestLabel & 779 & 3.3 \\ 
        HbA1c & Hemoglobin A1c Score & \RegressionLabel & \LabTestLabel &  778 & 5.5 mmol/mol \\
        Triglycerides & Triglyceride level  & \RegressionLabel & \LabTestLabel & 781 & 112.2 mg/dL \\ 

        \grayrow
        \multicolumn{1}{l}{\textbf{Mental Health}} & & & & & \\
        Mild Depression & Depression score (PHQ-8 $\geq$ 10) & \Binary & \ScreenerLabel  & 4,241 & 29.2\% \\
        Mild Anxiety & Anxiety disorder score (GAD-7 $\geq$ 10) & \Binary & \ScreenerLabel & 4,267 &  24.3\% \\
        Persistent Stress & Stress score (PSS $\geq$ 14) & \Binary & \ScreenerLabel & 5,955 & 64.6\% \\

        Depression/Anxiety Dx & Diagnosis of depression/anxiety & \Binary & \SelfSurveyLabel & 10,615 & 21.9\% \\
        Mental Health Med. & Depression/anxiety med. & \Binary & \SelfSurveyLabel & 8,034 & 11.4\% \\
        PHQ-8 & Depression score & \RegressionLabel & \ScreenerLabel & 4,241 & 7.1 \\
        GAD-7 & Generalized anxiety disorder score & \RegressionLabel & \ScreenerLabel & 4,267 & 6.4 \\
        PSS & Perceived stress scale score and subscores & \RegressionLabel & \ScreenerLabel & 5,955 &  16.6 \\

        \grayrow
        \multicolumn{1}{l}{\textbf{Sleep}} & & & & & \\
        Sleep Disorder Treatment & For individuals with a sleep disorder whether they are on a treatment & \Binary & \SelfSurveyLabel & 850 & 65.6\% \\ 
        Sleep Disturbance PRO & Sleep disturbance screener score & \RegressionLabel & \ScreenerLabel & 5,955 & 20.2 \\ 
        Sleep Impairment PRO & Sleep impairment screener score & \RegressionLabel & \ScreenerLabel & 5,955 & 19.9 \\ 
        
        \bottomrule
        \end{tabular}
    \label{tab:downstream_labels}
\end{table}
\endgroup

%% file: 4b-methods_model_and_training.tex
\section{Modeling}
\label{sec:model}

\subsection{Model Architecture}
\label{subsec:model_arch}

Prior work has shown the utility of using a masked reconstruction~\citep{he2022masked} pretraining objective for modeling long-context (hours or days) multimodal sensor data~\citep{narayanswamyscaling, xu2025lsm, erturkbeyond}. An important consideration in modeling these long-context sensor streams is the fragmentation inherent in the data, with missingness occurring for reasons such as loss of charge events, intermittent removal of the device, sensor or environmental noise, or various device operation modes (see Figure~\ref{fig:missing_modes}).
To this end, we designed a pretraining technique (see Figure~\ref{fig:pretraining}) that patches multimodal data inputs and replaces patches with missing observations with mask tokens. Following the Adaptive and Inherited Masking (AIM) strategy introduced by ~\cite{xu2025lsm}, our method treats the full applied mask as the union of the missing data mask and the artificial mask generated for the reconstruction pretraining objective. 
Two stage token masking, leveraging token dropout and attention masking, is used to ensure flexibility under the inherent variability of real world missingness while retaining computational efficiency. Additional details regarding AIM may be found in the original paper~\citep{xu2025lsm}.

Specifically, our model leverages a ViT-1D encoder backbone~\citep{dosovitskiyimage}, with the hyperparameters described in Table \ref{tab:vit_model_configs}. During AIM-based pretraining the latent embeddings are passed through a ViT-1D decoder which learns to reconstruct the ablated input. 
Masked tokens are represented through a learnable mask token. To represent temporal and sensor feature dimensions of the input, 2D additive positional encodings are applied to the tokens. Half of the encoding dimensions, corresponding to the feature dimension, are learned. Half of the positional encoding dimensions, corresponding to the temporal dimension, are 1D sinusoidal encodings~\citep{vaswani2017attention}, except for eight embedding features which correspond to cyclic datetime features~\citep{spathis2021self}: minute of the hour [0, 59], hour of the day [0, 23], day of the week [0, 6], and day of the year [0, 364]. All \modelname models were pretrained with a mean-squared error (MSE) reconstruction loss. The loss was calculated exclusively on the masked tokens which were originally observed (i.e., not missing).

\input{Tables_Final/tbED4_vit_hyperparams}

\subsection{Data Curation and Preprocessing}
\label{subsec:data_curation}

While large pretraining datasets are essential for building foundation models, evidence suggests that rigorous curation pipelines significantly improve model quality~\citep{grattafiori2024llama}. To ensure the integrity of our trillion-minute data corpus, we carefully implemented a pipeline for both pretraining and downstream tasks. First, to harmonize diverse sensor streams, all inputs were resampled to a uniform one-minute resolution and corrected for timezone offsets. We then applied physiological masking to eliminate non-biological artifacts. Skin conductance level (SCL) values were restricted to the physiological range of 0--60 $\mu$Siemens, and skin temperature readings were constrained to $0-41^{\circ}$C. Blood oxygen saturation ($SpO_2$) values below 70\% or flagged as invalid were treated as missing to remove spurious drops, while values exceeding 100\% were capped. For heart rate variability, variance metrics (RMSSD, SDNN) were capped at 125 ms to limit the impact of extreme outliers, and all HRV metrics were nullified if derived from windows with $<20\%$ valid inter-beat intervals. Additionally, data collected during periods where the device was detected as off-wrist were discarded to prevent noise injection. Finally, to stabilize training dynamics against remaining outliers, all features were z-score normalized using global per-feature statistics derived from the pretraining corpus and clipped to the interval $[-5, 5]$. We applied a sliding window with consecutive windows shifted by a variable interval. This shift was randomized between 8 and 24 minutes to enhance the diversity of the generated data windows. The windows with more than 80\% of inherent data missingness were removed. In total, our largest pretrain data volume contains $175,062,146$ day-long samples.

\subsection{Pretraining}
\label{subsec:pretraining_procedures}

We pretrain our model using a self-supervised masked autoencoder framework designed to reconstruct data from incomplete inputs. Our masking strategy AIM \citep{xu2025lsm} handles both inherent ("inherited") missingness and synthetic ("artificial") masking. Crucially, the reconstruction loss is computed exclusively on the artificially masked patches where ground truth values were originally present, ignoring inherited missingness. The artificial masking strategy employs a mixed probabilistic approach to simulate diverse real-world modes of multimodal sensor data fragmentation. Specifically, for a given sample, our model randomly applies one of the following masks: 80\% random patch masking, 50\% temporal block masking (simulating device removal), and 50\% modality block masking (simulating sensor dropout). The model utilizes a patch size of $[20, 1]$, processing 20-minute windows per sensor feature. Training was conducted with a global batch size of 4096 and an AdamW optimizer with a weight decay of $1 \times 10^{-4}$.
We leverage a cosine annealing learning rate scheduler with a base learning rate of $5 \times 10^{-4}$ and a linear warm up equal to $5\%$ of the number of steps. We train for a maximum of $N=1,000,000$ steps. The only exceptions occur when models exhibit overfitting. In these instances, pre-training was terminated early (specifically, at 100k steps for base model with 50k subjects, 80k steps for base Model with 5k subjects, and 60k steps for small model with 5k subjects).

\subsection{Discriminative Post-Training and Evaluation}
\label{subsec:post-training}

We independently fine-tuned a light-weight head on top of the frozen pretrained \modelname encoder for each downstream task to prevent conflating learning objectives. Our evaluation suite comprises a total of 35 discriminative tasks (binary and regression) which are described in more detail in Table~\ref{tab:downstream_labels}.
All the tasks correspond to person-level predictions where a single label applies to the entire participant (e.g., hypertension diagnosis, age). As such we aggregated the embeddings per person across all non-masked (no inherited missingness) tokens, computing the mean and standard deviations of each person's embeddings across all days of their data. To better match the variance of the embeddings with the sparsity of downstream labels, we reduce the \modelname aggregated embeddings to 50 principal components (PCA-50). For tasks where demographic features are included, age, sex, BMI, and race are concatenated with the reduced embeddings prior to the linear probe.

For classification tasks, we attached a linear probe to the reduced embeddings from the frozen encoder, and optionally demographic features. We trained using a logistic regression head with an AdamW optimizer (learning rate $5 \times 10^{-3}$, weight decay $1 \times 10^{-4}$) for 400 steps. We evaluate classification performance with the Receiver Operating Characteristic Area Under the Curve ($ROC AUC$) and the $F1$ score.
For regression tasks, we employed a similar setup with a linear regression head, evaluating performance via Pearson correlation ($r$) and Mean Absolute Error ($MAE$). To ensure robust evaluation, all presented results are the aggregated out-of-fold (OOF) performance with a five-fold cross validation setup. Note as each subject accounts for a single data sample these folds are naturally person independent. The subjects per split remain fixed across all downstreams. Additionally, while F1 and MAE were aggregated with arithmetic mean and standard deviation, Pearson ($r$) and $ROCAUC$ were aggregated in transform spaces to account for their skewness; Pearson correlation in the z-transform space, and ROC AUC the logit-transform space, before being back-transformed. Standard deviation was also calculated in the transformed space and back-transformed to give asymmetric error values.

\subsection{Generative Evaluation}
\label{subsec:gen_task_eval}

Since \modelname was trained with an MAE-like \citep{he2022masked, xu2025lsm} objective, it naturally exhibits out-of-box generative capabilities which allow it to infill and extend unobserved multimodal sensor data. To evaluate these generative capabilities we formulate the following generative tasks.
\textit{Random Imputation} masks out 80\% of total tokens across signals and time, emulating generic random noise (e.g., random signals missing at random times).  \textit{Temporal Imputation} and \textit{Temporal Extrapolation} ablation of a contiguous temporal window of length [10, 30, 60 minutes], either in the middle or at the end of the sequence, emulating intermittent removal, loss of charge events, etc. (e.g., all sensor features missing for a contiguous block of time).
\textit{Signal Imputation} masks all time points for a random set of [2/26, 6/26,
12/26] signal channels, emulating missing sensor channels (e.g., various sensor loadouts, sensor dropout, non-random missingness). Reconstruction performance was calculated with mean squared
error (MSE) calculated originally on observed masked tokens, averaging only over the data points that have a ground truth. 95\% confidence intervals were generated by bootstrapping the generative errors across 100 iterations. The results of this evaluation are presented in Table~\ref{tab:generative_results}.

To better contextualize the utility of these generative capabilities, we evaluate \modelname's ability to reconstruct partially observed data to provide more meaningful daily summary statistics. Specifically, we simulate missingness by masking a contiguous 1-hour duration across all sensor features. We then use \modelname to reconstruct this missing data. We predict daily metric predictions across \textit{Steps}, \textit{Sleep Stage Minutes}, \textit{Exercise Minutes}, \textit{SPO2 Level Minutes}, and \textit{Wrist Temperature Level Minutes}. We compare \modelname to the ground truth, un-ablated data, and baseline against a method which does not infill missingness. Total recovered minutes from the hour of data loss were compared to the ground truth, with a 95\% confidence intervals generated from 100 bootstrap iterations. The results of this experiment are presented in Table~\ref{tab:recovery_metrics}.

As mentioned in Methods~\ref{subsec:pretraining_datasets}, both the generative evaluation and the daily metrics estimation are reported on an independent test set derived from $10,000$ subjects.

\subsection{Engineered Baseline Features}
\label{subsec:engineered_baseline_features}

To contextualize the performance of \modelname embeddings, we compare against supervised models trained on engineered features derived from the same sensor streams. These engineered features are described in detail in Table~\ref{tab:baseline_features} in the Appendix, and are described in a high level below.

For each participant-day, we compute a fixed-length summary vector by aggregating each of the 34 minutely sensor features (Table~\ref{tab:features}) over the 24-hour window using daily summary statistics, following established methodologies in wearable research and chronobiology. We extracted 20 distinct features per channel (680 total) capturing distributional, volatile, and chronobiological dynamics. To preserve information about data fragmentation, the missing rate was calculated on the raw data for each channel prior to any imputation. Missing data were then resolved via  linear interpolation with back/forward filling at the start and ends of sequences to ensure continuous temporal derivatives.

\begin{itemize}
    \item Distributional and behavioral: Missingness and signal sparsity were quantified as hardware failure / behavioral phenotypes \citep{torous2017,choi2011}. Signal dispersion was captured via the mean, median, IQR, skewness, kurtosis, CV, and RMS. To robustly capture physiological extremes against high-frequency artifacts, 5th and 95th percentiles were used in lieu of absolute minimums and maximums.
    \item Volatility: Short-term rate-of-change was measured using mean absolute minute-to-minute differences and the root mean square of successive differences (RMSSD) \citep{shaffer2017overview}, an established metric for epoch-to-epoch variability in digital phenotyping.
    \item Morphology: Signal fragmentation and bandwidth were quantified via the mean-centered Zero Crossing Rate and Hjorth Complexity \citep{hjorth1970}. These time-domain metrics were selected to effectively capture the mean frequency and bandwidth of the signals without requiring computationally intensive Fourier transforms on the longitudinal data.
    \item Chronobiological: Diurnal rhythms were modeled via 24-hour Cosinor rhythmometry to extract the amplitude and acrophase \citep{cornelissen2014}. Circadian fragmentation was assessed through Intradaily Variability (IV) \citep{vansomeren1999}, while temporal signal persistence was measured via lag-1 autocorrelation.
\end{itemize}

These features represent the conventional approach to wearable health prediction: daily summary statistics fed to a standard classifier or regressor. We denote this baseline as ``FE'' (Feature Engineered) throughout. The same downstream training procedure described in Section~\ref{subsec:post-training} is applied: logistic regression for classification tasks and linear regression for regression tasks, with identical cross-validation splits and person-independent held-out evaluation. To address the high dimensionality of the engineered feature set (680 features), Principal Component Analysis (PCA) was used to reduce the feature vector to 50 principal components, similarly to the treatment of the \modelname embeddings. Where demographic features are included age, sex, BMI, and race are concatenated with the reduced engineered feature vector prior to model fitting. Similar to the linear probe, supervised baselines are evaluated with five-fold cross validation.

\subsection{Few Shot Experiments}
\label{subsec:few_shot_experiments}

To evaluate label efficiency we executed few shot experiments for each \modelname model variant and the engineered features described in Section \ref{subsec:engineered_baseline_features}. For each evaluated task the downstream models were trained using 5 folds and different sample percentages (10, 20, 30, 50, 60, 70, 80, 90 and 100). Specifically, in Figure~\ref{fig:few_shot} we visualize the few-shot performance of two \modelname variants and supervised baselines trained with only demographics or engineered features.

%% file: Tables_Final/tbED4_vit_hyperparams.tex
\begin{table}[htbp]
\centering
\caption{\textbf{Model Configurations for Each Size.} Architecture parameters for our ViT \citep{dosovitskiyimage} MAE-based \citep{he2022masked} \modelname across different model size variants, along with their total parameter counts.}
\label{tab:vit_model_configs}
\begin{tabular}{lcccc}
\toprule
\multirow{2}{*}{\textbf{Parameter}} & \multicolumn{4}{c}{\textbf{Model Variant}} \\
\cmidrule(lr){2-5}
& \textbf{XXS} & \textbf{XS} & \textbf{S} & \textbf{B} \\
\midrule

\multicolumn{5}{l}{\textit{Encoder}} \\
Hidden Size & 64 & 128 & 256 & 768 \\
MLP Dimension & 256 & 512 & 1024 & 3072 \\
Number of Heads & 1 & 2 & 4 & 12 \\
Number of Layers & 2 & 4 & 8 & 12 \\

\midrule

\multicolumn{5}{l}{\textit{Decoder}} \\
Hidden Size & 48 & 96 & 192 & 512 \\
MLP Dimension & 192 & 384 & 768 & 2048 \\
Number of Heads & 1 & 2 & 4 & 16 \\
Number of Layers & 1 & 1 & 2 & 8 \\

\midrule

\textit{Total Parameters} & {138,740} & {933,204} & {7,290,068} & {110,763,412} \\

\bottomrule
\end{tabular}
\end{table}

%% file: 4c-methods_latent_analysis.tex

\section{Analysis of the Model Embeddings in Latent Space}

\subsection{SHapley Additive exPlanation Analysis}

\newcommand{\R}{\mathbb{R}}

To identify the latent structure and the physiological semantics encoded within the high-dimensional \modelname embeddings, we employed SHapley Additive exPlanations (SHAP) \citep{lundberg2017unified} to quantify the contribution of each embedding dimension to specific downstream tasks. As previously mentioned, we utilized a Principal Component Analysis (PCA) preprocessing step before the linear probing heads to reduce dimensionality to allow for more appropriate comparisons with the baseline models, as well as reducing collinearity of the raw embedding space. To analyze the latent structure of our embeddings in this setting, we leveraged the following projection mechanism to map feature importance from the reduced PCA space back to the original \modelname latent space.

\noindent
For each task, we fit a linear model $f_i(x)$ on the PCA-transformed dimension ($Z$ in Eq. \eqref{eq:pca-transformation}), 
\begin{equation}
    Z = XV^\top
    \label{eq:pca-transformation}
\end{equation}
where $X \in \R^{N\times D}$ denotes the input data that is centered ($\mathbb{E}[X]=0$), and $V \in \mathbb{R}^{K\times D}$ are the principal components; in this work, we chose $K=50$. Each linear prediction head learns a set of coefficients $\beta \in \R^{K\times 1}$, predicting $\hat{y} = Z\beta + b$ for $b\in \R$. From here, there are two ways of mapping the importance of the PCA-transformed dimensions back to the original \modelname embeddings:

\noindent
\textbf{Exact Analytical Weight Collapse}. To derive exact attribution for the linear heads, we treated the PCA transformation and the linear probe as a single composite linear layer. Substituting Eq. \eqref{eq:pca-transformation} into the linear equation yields Eq. \eqref{eq:pca_sub}:
\begin{equation}
\hat{y} = Z\beta + b = \left( X V^\top \right) \beta + b = X (V^\top \beta) + b
\label{eq:pca_sub}
\end{equation}
We define the \emph{Effective Weight Vector} $W_{eff} \in \R^D$ as the projection of the probe coefficients back onto the original feature axes:
\begin{equation}
W_{eff} = V^\top \beta
\end{equation}
For a linear model with centered input features, the exact independent SHAP value for a feature is the product of the feature's value and its corresponding weight. Thus, the exact SHAP attribution matrix $\Phi \in \R^{N \times D}$ in the original embedding space is computed analytically as:
\begin{equation}
\Phi_{i,d} = W_{eff, d} \cdot X_{i,d}
\end{equation}
By linearity, this formulation satisfies the SHAP local accuracy (efficiency) axiom, ensuring that $\sum_{d=1}^D \Phi_{i,d} = \hat{y}_i - b$.

For multiclass logistic regression tasks with a set of classes $C$, an effective weight vector $W_{eff}^{(c)} = V^\top \beta^{(c)}$ is computed for each class $c \in C$. The class-specific SHAP value is strictly $\Phi^{(c)}_{i,d} = W_{eff,d}^{(c)} \cdot X_{i,d}$. To establish a singular task-level attribution, we aggregated the absolute contributions across all classes:
\begin{equation}
\bar{\Phi}_{i,d} = \frac{1}{|C|} \sum_{c \in C} \left| \Phi^{(c)}_{i,d} \right|
\end{equation}
To ensure robustness against data splits, we computed SHAP values for each subject based on their out-of-fold predictions across a 5-fold cross-validation scheme. The global importance $I_d$ of embedding dimension $d$ was defined as the mean absolute SHAP value across all $N$ subjects in the dataset:
\begin{equation}
I_d = \frac{1}{N} \sum_{i=1}^{N} |\bar{\Phi}_{i,d}|
\end{equation}

\noindent
\textbf{Latent Profile Similarity and Network Visualization.} To visualize the shared physiological semantics encoded within the \modelname embeddings across different clinical domains, we modeled the pairwise relationships between downstream tasks based on their exact SHAP attribution profiles. Let $I_t \in \R^D$ denote the global importance vector for task $t$, computed across all embedding dimensions. To ensure scale invariance across tasks with varying absolute prediction margins, each attribution profile was first max-normalized: $\tilde{I}_t = I_t / \max(I_t)$. We then computed the pairwise $L_1$ (Manhattan) distance between all normalized profiles. Let $\Delta_{t, t'} = \|\tilde{I}_t - \tilde{I}_{t'}\|_1$ represent the distance between tasks $t$ and $t'$. The latent profile similarity matrix $S \in \R^{T \times T}$ for the $T$ tasks was defined as:
\begin{equation}
S_{t, t'} = 1 - \frac{\Delta_{t, t'}}{\max_{u,v} \Delta_{u, v}}
\end{equation}
where $S_{t, t'} = 1$ indicates identical latent utilization of the embedding space, and $S_{t, t'} = 0$ indicates maximum divergence.

\subsection{Embedding Distances and Intrinsic Dimensionality}

To evaluate the structural density of the latent space across model scales, we computed pairwise Euclidean distances between user embeddings. To ensure computational tractability and prevent memory constraints, we randomly subsampled 5,000 users for each model size. Missing values within the embeddings were resolved using mean imputation. To denoise the representations prior to distance calculation, the embedding dimensionality was reduced using Principal Component Analysis (PCA), retaining 99\% of the variance. The condensed pairwise Euclidean distances were then visualised using kernel density estimation (KDE) to assess the dispersion and clustering of the latent representations across different model capacities.

The intrinsic dimensionality and compressibility of the learned representations were quantified by examining the cumulative explained variance. Following mean imputation of the embeddings, we applied PCA to calculate up to 75 principal components (or the maximum available dimensions for smaller models). The explained variance ratio for each component was extracted and cumulatively summed to generate scree plots. This allowed us to identify the presence of dimensional collapse or anisotropy across the different model sizes by observing how quickly the explained variance saturated.

%% file: 4d-methods_ARXIV_classroom.tex
\section{Agentic Classroom Search for Predictive ``Head'' Development}
\label{sec:classroom}

To search the space of task-specific prediction "heads" for \modelname, we developed a framework for self-evolving algorithm generation that formulates solution synthesis as a competitive, collaborative optimisation problem
solved by a ``classroom'' of parallel LLM agents. We note that while this framework is helpful in the context of our model which has many task ``heads'', it is applicable to any task whose solution quality can be expressed as a scalar score, not just those in the domain of wearable sensing.

\subsection{Implementation Details}

The framework is implemented as a lightweight Python library designed to run entirely within a Google Colab notebook environment. Student agents are instantiated as parallel calls to the Gemini API. We note that this framework is model-agnostic and compatible with any LLM exposing a text-generation endpoint. All code execution occurs within Colab's sandboxed Python runtime with solutions executing on CPU compute. We note that such a framework may be extended to support hardware acceleration (GPU/TPU) for more computationally intensive tasks. The architecture imposes no external infrastructure requirements beyond a Colab instance and API access, and is designed to be extensible to distributed computing backends for larger-scale experiments.

\textbf{Hyperparameters.} Specifically, for our experiments, we instantiate a classroom of $N = 5$ student agents leveraging the following underlying language models\footnote{Accessed via the Google Gemini API between Feb 2026 and April 2026.}: \texttt{gemini-2.5 flash}, \texttt{gemini-2.5 pro}, \texttt{gemini-3 flash preview}, \texttt{gemini-3.1 flash lite preview}, and \texttt{gemini-3.1 pro preview}. The classroom search is set to iterate for a maximum $T = 20$ learning cycles. 
As highlighted in Figure~\ref{fig:codegen_meta_analysis} we run experiments both \emph{with} and \emph{without} agent collaboration, events where the agents analyze their own solutions or the solutions of other students. The score on which agents hill-climb is a combination of multiple classification or regression metrics (given the task).

\textbf{K-Fold Cross Validation.} Similar to the setting where a linear head is applied to the \modelname learned embeddings (\ref{subsec:post-training}), we report 5-fold cross validation performance for each of the 35 discriminative tasks, leveraging the same folds as above (\ref{subsec:post-training}). Specifically, for a given experiment (a specific fold and a specific task) we randomly split off $20\%$ of $\mathcal{D}_{train}$ to act as $\mathcal{D}_{val}$. We refer to the out-of-fold (OOF) data as $\mathcal{D}_{test}$. 
Over $T$ learning cycles, the classroom learns from $\mathcal{D}_{train}$ and makes predictions for and iterates on the performance on $\mathcal{D}_{val}$. At the end of $T$ iterations, a best solution $s^{*}$ is selected based on the best validation score $\phi^{*}$. $s^{*}$ is then train with the entirety of the $\mathcal{D}_{train}$ (including the originally split validation data), and evaluated on the OOF $\mathcal{D}_{test}$ to produce the reported results. By formulating the evaluation across $K$ folds as $K$ independent experiments, we effectively prevent train-test leakage. However, it should also be noted, that this formulation results in different ``found'' solutions per fold for a given task.

\textbf{Summary and Example Artifacts.} In total, leveraging this framework we efficiently conduct $30,516$ total experiments across $35$ tasks x $5$ folds x $5$ student agents x $20$ learning iterations x $2$ and collaboration conditions (with and without). Note that the number of total experiments accounts for students which did not conduct a full 20 learning iterations because of the patience criteria. An example student agent prompt can be found in Code~\ref{appendix:classroom_codegen_prompt_example} and an example agent solution can be found in Code~\ref{appendix:classroom_codegen_code_example} of Appendix~\ref{sec:appendix_classroom}.

%% file: 4e-methods_pha.tex
\section{Evaluating \modelname as a Tool for a Health Agent}
\label{methods:pha}

To rigorously evaluate whether integrating \modelname improves the clinical utility of LLM-driven health agents, we designed a blinded, comparative study centered on diverse, real-world patient profiles. The health agent was tasked with generating personalized health summaries under varied contextual conditions with a full integration of our AI inferences. These generated responses were then subjected to a blinded evaluation by a panel of board-certified physicians. By assessing the outputs across multiple dimensions of clinical utility and safety, this experimental design allows us to isolate and quantify the specific value \modelname adds to Personal Health Agents.

\subsection{Generating the \modelname-Augmented Responses}
The health agent was tasked with synthesizing a comprehensive summary of each user's health status based on their specific data profile. The LLM powering the health agent was kept constant as Gemini 3 Flash with a temperature of 0.2. The exact system prompt designed to govern these responses is detailed below in Table \ref{appendix:sensorfm_agent_prompt_template}. The prompt is structured to ingest multimodal user context of demographics, aggregated wearable statistics, and \modelname inferences, while enforcing formatting and safety guardrails (e.g. qualitative interpretation of predictive metrics). The full metabolic panel is not provided in the agent prompt.

\subsection{Testing \modelname}
To evaluate the clinical utility of the inferences generated by \modelname, we established three distinct experimental conditions. The first condition, \textit{(A) Extra Context (\modelname Predictions)}, provided the health agent with user demographics, feature-engineered daily metrics, and a diverse range of \modelname predictions (e.g. hyperlipidemia, PHQ-8, sleep disturbance). The second condition, \textit{(B) Extra Context (Available Ground Truth)}, mirrored condition A but replaced the model predictions with the available ground-truth targets. Because only metabolic markers were available in the downstream evaluation dataset, other targets (such as mental health and sleep disturbance metrics) were inherently absent.
Finally, the baseline condition, \textit{(C) No Extra Context}, supplied the health agent strictly with demographics and feature-engineered daily metrics. During evaluation, the presentation order of these conditions was randomized to prevent reviewer bias.

\subsection{User Profiles}
To ensure the robustness and generalizability of our evaluation, we constructed ten representative user profiles that encapsulate a diverse range of common health scenarios. These profiles were stratified into two primary categories. 

The first category includes four profiles representing individuals without diagnosed chronic diseases but with distinct health objectives: (1) a performance-oriented individual training for athletic goals; (2) a generally healthy individual seeking to improve a specific wellness aspect, such as sleep quality; (3) an individual with a sedentary lifestyle but no formal disease diagnosis (sub-healthy); and (4) an individual recovering from an acute injury or life event disrupting their health baseline.

The second category comprises six profiles designed to reflect major public health concerns, with
each profile centered on a prevalent chronic condition: (5) Anxiety/Depression, (6) Hypertension,
(7) Respiratory Conditions, (8) Hypercholesterolemia, (9) Diabetes, and (10) Cardiovascular Disease (CVD). It is noteworthy that these profiles were designed to reflect real-world complexity, and individuals within these latter six categories may present with comorbidities. 

We pull the users used in \cite{heydari2025anatomy} that have corresponding wearable data, resulting in 12 users from the healthy profiles and 19 from the unhealthy profiles. For each individual, we extracted their statistical aggregrate wearable data, demographic information, and available ground-truth metabolic markers. Additionally, we take their minutely wearable data and use our \modelname for the AI model predictions.

\subsection{Physician Evaluators}
A cohort of four board-certified physicians with specialties in internal medicine and family medicine, with an average of 11.75 years of clinical experience was recruited to evaluate the clinical soundness of the generated summaries. The physicians reviewed the responses in a randomized order and were blind to the experimental conditions. Specifically, they were not informed how inferences were integrated into the provided text and that predictive AI models were used to generate inferences. They were tasked solely with rating the clinical quality of the responses according to the established rubric, without knowledge of the underlying system architecture.

To validate the consistency of the physician ratings, we calculated the Intraclass Correlation Coefficient (ICC3k), utilizing a two-way mixed-effects model based on the average score of the raters for each rubric dimension. The analysis highlights the nuanced, highly individualized nature of expert clinical evaluation. The panel demonstrated a moderate and solid consensus on Relevance (ICC = 0.653, 95\% CI: [0.52, 0.76]), indicating shared agreement on the most critical medical information. For dimensions requiring more subjective clinical interpretation, such as Context (ICC = 0.478), Harm (ICC = 0.416), and Personalization (ICC = 0.387), the scores reflect the expected, natural variance inherent to diverse medical practices. Furthermore, the broad variance in the Justifiable dimension (ICC = -0.088) underscores that physicians maintain uniquely stringent, individualized thresholds for what constitutes "justifiable" clinical reasoning when operating strictly from static data profiles. 

\subsection{Evaluation Rubric}
To evaluate the generated summaries, we developed a comprehensive rubric comprising five distinct dimensions: Context, Personalization, Justifiability, Relevance, and Harm. Each dimension assesses a unique facet of clinical utility, specifically targeting the agent's ability to ground its responses in patient-specific data to deliver safe, tailored advice. The exact evaluation criteria for each dimension are detailed in Survey~\ref{box:pha_eval}.

\subsection{Evaluation Set-up}

During evaluation, physician evaluators were presented with the user's demographics, daily aggregated wearable statistics, and the full metabolic panel. Crucially, the evaluators were strictly blind to the underlying \modelname predictions. This blinding was implemented to prevent anchoring bias, ensuring that the physicians assessed the clinical soundness of the generated responses based entirely on their own independent medical judgment of the raw patient data, rather than being influenced by the AI's diagnostic estimates.

For each patient profile, physicians were explicitly instructed to read all three standardized model responses (anonymized + randomized as Models A/B/C) side-by-side. Evaluators assessed all three Model A/B/C responses for a single rubric dimension before proceeding to the next question. This parallel presentation format enabled evaluators to score the responses using a 5-point Likert scale while inherently facilitating relative comparisons between the models, allowing for the extraction of both absolute quality metrics and comparative win-rates. 

%% file: 4f-methods_ARXIV_additional_info.tex
\subsubsection*{Author contributions} GN, MAX, XL, DM contributed to the conception and design of the work; GN, MAX, AH, SAG, BY, AW, NBA, JH, CH, AM, XL, DM contributed to the data acquisition and curation; GN, MAX, AH, SAG, MG, KV, ZZ, JG, LA, HY, AW, XL, DM contributed to the technical implementation; DS, HY, YK, YZ, SS, YY provided technical and infrastructure guidance; JS, IGL, JH, JG provided clinical inputs to the study; DS, HP, OX, DB, KA, PK contributed to the supplementary data analysis; GN, MAX, AH, SAG, MG, KV, ZZ, JG, LA, DS, HY, HP, OX, DB, JB, JM, YK, YZ, NR, SS, KA, TA, JS, MZP, BY, AW, NBA, JMR, IGL, YL, JH, AP, CH, YY, AM, PK, MM, SP, XL, DM contributed to the drafting and revising of the manuscript.

\subsubsection*{Correspondence}
Correspondence should be addressed to \{girishvn, xumax, xliucs, dmcduff\}@google.com.

%% file: appendix.tex
\appendix
\section*{\LARGE Appendix}
\label{sec:appendix}

\startcontents[appendices]

\input{x-scaling_results}
\input{x-downstream_task_results}
\input{x-reconstruction_examples}

\input{x-latent_space_analysis_and_viz}

\input{x-classroom_setup}

\input{x-pha_eval}
\input{x-additional_dataset}
\input{x-survey_questions}

%% file: x-scaling_results.tex
\section{Scaling Results}
\label{sec:appendix_scaling_results}
We sweep four \modelname model variants (XXS, XS, S, B; spanning $10^{5}$ to $10^{8}$ parameters) across four pretraining data volumes (5K, 50K, 500K, and 5M subjects; spanning $10^7$ to $10^9$ data hours). Table~\ref{tab:scaling_results} provides an overview of model performance across the pretraining objective (validation reconstruction loss), generative tasks (random imputation, temporal interpolation/extrapolation, signal imputation), and discriminative tasks (averaged classification ROC AUC and regression Pearson correlation). Tables~\ref{tab:predictive_task_scaling} and~\ref{tab:predictive_task_scaling2} break the discriminative linear probe results down per-task across all 35 downstream tasks (organized by Demographics, Lifestyle, Cardiovascular, Metabolic, Mental Health, and Sleep), with Pearson $r$ and ROC AUC in Part~I (Tables~\ref{tab:predictive_task_scaling}) and complementary MAE and F1 metrics in Part~II (Table~\ref{tab:predictive_task_scaling2}). Table~\ref{tab:delta_predictive_tasks} reports the per-task improvement from including demographic features alongside \modelname embeddings versus a supervised feature-engineered baseline; the marginal value of demographics diminishes with model scale, and $33$ of $35$ tasks exhibit the lowest demographic lift at B. In these tables, the \modelname embeddings are first reduced to 50 principal components (PCA-50) to match the lower variance of the sparse downstream labels.

\input{Tables_Final/tbED5_overview_scaling}

\input{Tables_Final/tbED6_pred_scaling_I}
\input{Tables_Final/tbED7_pred_scaling_II}
\input{Tables_Final/tbED8_pred_scaling_effect_of_demos}

%% file: Tables_Final/tbED5_overview_scaling.tex
\begin{table}[!htbp]
\centering
\footnotesize

\caption{\textbf{The Effect of Scaling Across Pretraining, Generative and Discriminative Tasks.}
This table presents the performance of \modelname across pretrain, generative, and discriminative tasks as a function of model capacity and pretrain data volume. In general larger models trained with more data achieve improved performance.
In pretraining the model is tasked with reconstructing a sample ablated with either random, temporal, or signal masking. As such the validation loss is a compound generative metric consisting of Random Imp. (80\%), Temporal Imp. (50\%) and Signal Imp. (50\%).
For pretraining and generative tasks we present the average reconstruction Mean Squared Error (MSE) with $95\%$ bootstrapped confidence intervals generated through 100 bootstrap iterations. For discriminative tasks we present the mean performance across all tasks, where each task is evaluated with 5-fold cross validation. Average Receiver Operating Characteristic
Area Under the Curve (ROC AUC) is calculated in the logit-transform space and back-transformed. Average Pearson correlation ($r$) is calculated in the z-transform space and back-transformed. Colors are normalized across each task block; best model performance is bolded and has the deepest shade.}

\begin{tabular}{lcccccc}
\toprule

\multirow{3}{*}{\textbf{Task}} & \multirow{3}{*}{\textbf{Metric}} &
\multirow{3}{*}{\thead{\textbf{Pretraining}\\\textbf{Data Volume}\\(Subjects)}} & \multicolumn{4}{c}{\textbf{Model Variant (Parameter Count)}} \\ \\
\cmidrule(lr){4-7}
&&& XXS (10$^{5}$) & XS (10$^{6}$) & S (10$^{7}$) & B (10$^{8}$) \\

\midrule
\grayrow \multicolumn{7}{l}{\textit{\textbf{Pretraining}}} \\
\multirow{4}{*}{\thead[l]{Reconstruction\\(Val. Loss)}} & \multirow{4}{*}{\MSELabel}
& 5K   & \cellcolor{RoyalBlue!35}{0.428} $\pm$ 0.002 & \cellcolor{RoyalBlue!65}0.364 $\pm$ 0.002 & \cellcolor{RoyalBlue!46}0.402 $\pm$ 0.003 & \cellcolor{RoyalBlue!0}1.082 $\pm$ 0.003 \\
&& 50K  & \cellcolor{RoyalBlue!41}0.415 $\pm$ 0.002 & \cellcolor{RoyalBlue!71}0.351 $\pm$ 0.002 & \cellcolor{RoyalBlue!87}0.319 $\pm$ 0.002 & \cellcolor{RoyalBlue!14}0.466 $\pm$ 0.002 \\
&& 500K & \cellcolor{RoyalBlue!39}0.419 $\pm$ 0.001 & \cellcolor{RoyalBlue!72}0.349 $\pm$ 0.002 & \cellcolor{RoyalBlue!92}0.307 $\pm$ 0.002 & \cellcolor{RoyalBlue!95}0.299 $\pm$ 0.001 \\
&& 5M   & \cellcolor{RoyalBlue!42}0.414 $\pm$ 0.002 & \cellcolor{RoyalBlue!72}0.350 $\pm$ 0.002 & \cellcolor{RoyalBlue!93}0.306 $\pm$ 0.001 & \cellcolor{RoyalBlue!100}\textbf{0.285} $\pm$ 0.002 \\

\addlinespace
\grayrow \multicolumn{7}{l}{\textit{\textbf{Generative Tasks}}} \\
\multirow{4}{*}{\thead[l]{Random Imp. \\ (80\%)}} & \multirow{4}{*}{\MSELabel}
& 5K   & \cellcolor{RoyalBlue!80}0.389 $\pm$ 0.001 & \cellcolor{RoyalBlue!90}0.303 $\pm$ 0.001 & \cellcolor{RoyalBlue!88}0.321 $\pm$ 0.001 & \cellcolor{RoyalBlue!0}1.077 $\pm$ 0.002 \\
&& 50K  & \cellcolor{RoyalBlue!82}0.372 $\pm$ 0.001 & \cellcolor{RoyalBlue!91}0.293 $\pm$ 0.001 & \cellcolor{RoyalBlue!96}0.250 $\pm$ 0.001 & \cellcolor{RoyalBlue!79}0.400 $\pm$ 0.001 \\
&& 500K & \cellcolor{RoyalBlue!81}0.375 $\pm$ 0.001 & \cellcolor{RoyalBlue!91}0.290 $\pm$ 0.001 & \cellcolor{RoyalBlue!97}0.241 $\pm$ 0.001 & \cellcolor{RoyalBlue!99}0.227 $\pm$ 0.001 \\
&& 5M   & \cellcolor{RoyalBlue!82}0.371 $\pm$ 0.001 & \cellcolor{RoyalBlue!91}0.292 $\pm$ 0.001 & \cellcolor{RoyalBlue!97}0.240 $\pm$ 0.001 & \cellcolor{RoyalBlue!100}\textbf{0.215} $\pm$ 0.001 \\

\midrule

\multirow{4}{*}{\thead[l]{Temporal Interp. \\ (30 Mins)}} & \multirow{4}{*}{\MSELabel}
& 5K   & \cellcolor{RoyalBlue!52}0.478 $\pm$ 0.004 & \cellcolor{RoyalBlue!35}0.562 $\pm$ 0.006 & \cellcolor{RoyalBlue!48}0.499 $\pm$ 0.006 & \cellcolor{RoyalBlue!0}1.077 $\pm$ 0.009 \\
&& 50K  & \cellcolor{RoyalBlue!56}0.458 $\pm$ 0.005 & \cellcolor{RoyalBlue!25}0.608 $\pm$ 0.004 & \cellcolor{RoyalBlue!80}0.390 $\pm$ 0.005 & \cellcolor{RoyalBlue!30}0.584 $\pm$ 0.007 \\
&& 500K & \cellcolor{RoyalBlue!55}0.464 $\pm$ 0.003 & \cellcolor{RoyalBlue!30}0.584 $\pm$ 0.005 & \cellcolor{RoyalBlue!78}0.399 $\pm$ 0.003 & \cellcolor{RoyalBlue!85}0.373 $\pm$ 0.004 \\
&& 5M   & \cellcolor{RoyalBlue!57}0.451 $\pm$ 0.003 & \cellcolor{RoyalBlue!15}0.668 $\pm$ 0.006 & \cellcolor{RoyalBlue!81}0.389 $\pm$ 0.004 & \cellcolor{RoyalBlue!90}\textbf{0.353} $\pm$ 0.002 \\

\midrule

\multirow{4}{*}{\thead[l]{Temporal Extrap. \\ (30 Mins)}} & \multirow{4}{*}{\MSELabel}
& 5K   & \cellcolor{RoyalBlue!35}0.608 $\pm$ 0.005 & \cellcolor{RoyalBlue!30}0.639 $\pm$ 0.005 & \cellcolor{RoyalBlue!37}0.597 $\pm$ 0.005 & \cellcolor{RoyalBlue!0}1.100 $\pm$ 0.006 \\
&& 50K  & \cellcolor{RoyalBlue!45}0.542 $\pm$ 0.004 & \cellcolor{RoyalBlue!15}0.778 $\pm$ 0.009 & \cellcolor{RoyalBlue!55}0.497 $\pm$ 0.005 & \cellcolor{RoyalBlue!20}0.721 $\pm$ 0.004 \\
&& 500K & \cellcolor{RoyalBlue!44}0.545 $\pm$ 0.003 & \cellcolor{RoyalBlue!20}0.726 $\pm$ 0.009 & \cellcolor{RoyalBlue!53}0.505 $\pm$ 0.003 & \cellcolor{RoyalBlue!56}0.490 $\pm$ 0.003 \\
&& 5M   & \cellcolor{RoyalBlue!46}0.539 $\pm$ 0.004 & \cellcolor{RoyalBlue!10}0.841 $\pm$ 0.006 & \cellcolor{RoyalBlue!54}0.503 $\pm$ 0.003 & \cellcolor{RoyalBlue!65}\textbf{0.463} $\pm$ 0.004 \\

\midrule

\multirow{4}{*}{\thead[l]{Signal Imp. \\ (35\%)}} & \multirow{4}{*}{\MSELabel}
& 5K   & \cellcolor{RoyalBlue!50}0.321 $\pm$ 0.002 & \cellcolor{RoyalBlue!70}0.250 $\pm$ 0.003 & \cellcolor{RoyalBlue!64}0.270 $\pm$ 0.002 & \cellcolor{RoyalBlue!0}1.093 $\pm$ 0.005 \\
&& 50K  & \cellcolor{RoyalBlue!55}0.302 $\pm$ 0.002 & \cellcolor{RoyalBlue!74}0.237 $\pm$ 0.002 & \cellcolor{RoyalBlue!83}0.206 $\pm$ 0.002 & \cellcolor{RoyalBlue!51}0.316 $\pm$ 0.003 \\
&& 500K & \cellcolor{RoyalBlue!54}0.307 $\pm$ 0.001 & \cellcolor{RoyalBlue!75}0.236 $\pm$ 0.001 & \cellcolor{RoyalBlue!87}0.193 $\pm$ 0.002 & \cellcolor{RoyalBlue!92}0.184 $\pm$ 0.001 \\
&& 5M   & \cellcolor{RoyalBlue!55}0.305 $\pm$ 0.003 & \cellcolor{RoyalBlue!75}0.236 $\pm$ 0.001 & \cellcolor{RoyalBlue!88}0.192 $\pm$ 0.001 & \cellcolor{RoyalBlue!100}\textbf{0.170} $\pm$ 0.001 \\

\addlinespace
\grayrow \multicolumn{7}{l}{\textit{\textbf{Discriminative Tasks}}} \\
\multirow{4}{*}{\thead[l]{Classification\\(Avg. Performance)}} & \multirow{4}{*}{\ROCAUCLabel}

& 5K & \bcell{29}{$.664$} & \bcell{47}{$.687$} & \bcell{50}{$.690$} & \bcell{5}{$.634$} \\
&& 50K & \bcell{27}{$.663$} & \bcell{42}{$.681$} & \bcell{67}{$.712$} & \bcell{51}{$.692$} \\
&& 500K & \bcell{28}{$.663$} & \bcell{42}{$.681$} & \bcell{66}{$.710$} & \bcell{94}{$.746$} \\
&& 5M & \bcell{27}{$.663$} & \bcell{43}{$.682$} & \bcell{71}{$.716$} & \bcell{100}{$\textbf{.752}$} \\

\midrule

\multirow{4}{*}{\thead[l]{Regression\\(Avg. Performance)}} & \multirow{4}{*}{\PearsonLabel}

& 5K & \bcell{27}{$.386$} & \bcell{40}{$.426$} & \bcell{49}{$.453$} & \bcell{5}{$.314$} \\
&& 50K & \bcell{29}{$.390$} & \bcell{43}{$.435$} & \bcell{71}{$.522$} & \bcell{58}{$.480$} \\
&& 500K & \bcell{23}{$.371$} & \bcell{39}{$.423$} & \bcell{75}{$.536$} & \bcell{98}{$.608$} \\
&& 5M & \bcell{33}{$.402$} & \bcell{41}{$.427$} & \bcell{82}{$.559$} & \bcell{100}{$\textbf{.612}$} \\

\bottomrule
\end{tabular}
\label{tab:scaling_results}
\end{table}

%% file: Tables_Final/tbED6_pred_scaling_I.tex

\begin{table}[!htbp]
\centering
\footnotesize
\caption{\textbf{Discriminative Task Performance Across Model Scales (Part I).}
The Table presents the performance of \modelname variants, pretrained with proportional data scales, on 35 discriminative tasks. In general performance improves with scale with B consistently achieving the best performance. 
\modelname variants are post-trained with PCA-50 reduced embeddings. 
For each task, we report the average 5-fold cross validation performance. Average Receiver Operating Characteristic Area Under the Curve (ROC AUC) is calculated in the logit-transform space and back-transformed. Average Pearson correlation ($r$) is calculated in the z-transform space and back-transformed. Standard deviations are calculated in the transformed space and back-transformed to give asymmetric error values. Colors are normalized per row; best model performance is bolded and has the deepest shade.}
\begin{tabular}{lcccccc}
\toprule
\textbf{Prediction Task} & \textbf{Type} & \textbf{Metric}
& \multicolumn{4}{c}{\textbf{Model Variant (Parameter Count)}} \\ 
\cmidrule(lr){4-7}
&&& XXS (10$^{5}$) & XS (10$^{6}$) & S (10$^{7}$) & B (10$^{8}$) \\
\midrule
\grayrow \multicolumn{7}{l}{\Demo \textit{\textbf{Demographics}}} \\
Age & \RegressionLabel & \PearsonLabel & \bcell{5}{$.716^{+\scriptscriptstyle .009}_{-\scriptscriptstyle .009}$} & \bcell{14}{$.759^{+\scriptscriptstyle .007}_{-\scriptscriptstyle .008}$} & \bcell{33}{$.843^{+\scriptscriptstyle .006}_{-\scriptscriptstyle .006}$} & \bcell{50}{$\textbf{.920}^{+\scriptscriptstyle .004}_{-\scriptscriptstyle .005}$} \\
BMI & \RegressionLabel & \PearsonLabel & \bcell{5}{$.445^{+\scriptscriptstyle .021}_{-\scriptscriptstyle .021}$} & \bcell{11}{$.494^{+\scriptscriptstyle .015}_{-\scriptscriptstyle .016}$} & \bcell{36}{$.701^{+\scriptscriptstyle .008}_{-\scriptscriptstyle .008}$} & \bcell{50}{$\textbf{.809}^{+\scriptscriptstyle .007}_{-\scriptscriptstyle .007}$} \\
Height & \RegressionLabel & \PearsonLabel & \bcell{5}{$.485^{+\scriptscriptstyle .013}_{-\scriptscriptstyle .013}$} & \bcell{12}{$.518^{+\scriptscriptstyle .013}_{-\scriptscriptstyle .014}$} & \bcell{40}{$.634^{+\scriptscriptstyle .015}_{-\scriptscriptstyle .015}$} & \bcell{50}{$\textbf{.675}^{+\scriptscriptstyle .012}_{-\scriptscriptstyle .012}$} \\
Weight & \RegressionLabel & \PearsonLabel & \bcell{5}{$.362^{+\scriptscriptstyle .021}_{-\scriptscriptstyle .021}$} & \bcell{14}{$.455^{+\scriptscriptstyle .013}_{-\scriptscriptstyle .013}$} & \bcell{39}{$.700^{+\scriptscriptstyle .012}_{-\scriptscriptstyle .012}$} & \bcell{50}{$\textbf{.809}^{+\scriptscriptstyle .007}_{-\scriptscriptstyle .007}$} \\
\addlinespace
\grayrow \multicolumn{7}{l}{\Life \textit{\textbf{Lifestyle}}} \\
Currently Working & \Binary & \ROCAUCLabel & \bcell{5}{$.763^{+\scriptscriptstyle .021}_{-\scriptscriptstyle .022}$} & \bcell{12}{$.787^{+\scriptscriptstyle .014}_{-\scriptscriptstyle .015}$} & \bcell{24}{$.829^{+\scriptscriptstyle .014}_{-\scriptscriptstyle .015}$} & \bcell{50}{$\textbf{.912}^{+\scriptscriptstyle .012}_{-\scriptscriptstyle .014}$} \\
Disability & \Binary & \ROCAUCLabel & \bcell{5}{$.689^{+\scriptscriptstyle .024}_{-\scriptscriptstyle .025}$} & \bcell{16}{$.705^{+\scriptscriptstyle .015}_{-\scriptscriptstyle .016}$} & \bcell{24}{$.717^{+\scriptscriptstyle .021}_{-\scriptscriptstyle .022}$} & \bcell{50}{$\textbf{.753}^{+\scriptscriptstyle .020}_{-\scriptscriptstyle .021}$} \\
Disability Affects Work & \Binary & \ROCAUCLabel & \bcell{5}{$.541^{+\scriptscriptstyle .033}_{-\scriptscriptstyle .034}$} & \bcell{26}{$.616^{+\scriptscriptstyle .051}_{-\scriptscriptstyle .054}$} & \bcell{29}{$.626^{+\scriptscriptstyle .059}_{-\scriptscriptstyle .063}$} & \bcell{50}{$\textbf{.699}^{+\scriptscriptstyle .040}_{-\scriptscriptstyle .043}$} \\
Smoking & \Binary & \ROCAUCLabel & \bcell{5}{$.710^{+\scriptscriptstyle .044}_{-\scriptscriptstyle .048}$} & \bcell{7}{$.721^{+\scriptscriptstyle .033}_{-\scriptscriptstyle .036}$} & \bcell{16}{$.752^{+\scriptscriptstyle .051}_{-\scriptscriptstyle .059}$} & \bcell{50}{$\textbf{.870}^{+\scriptscriptstyle .060}_{-\scriptscriptstyle .098}$} \\
Medicaid & \Binary & \ROCAUCLabel & \bcell{5}{$.699^{+\scriptscriptstyle .036}_{-\scriptscriptstyle .039}$} & \bcell{15}{$.727^{+\scriptscriptstyle .028}_{-\scriptscriptstyle .030}$} & \bcell{29}{$.762^{+\scriptscriptstyle .022}_{-\scriptscriptstyle .023}$} & \bcell{50}{$\textbf{.814}^{+\scriptscriptstyle .021}_{-\scriptscriptstyle .023}$} \\
No Medications & \Binary & \ROCAUCLabel & \bcell{5}{$.676^{+\scriptscriptstyle .003}_{-\scriptscriptstyle .003}$} & \bcell{13}{$.688^{+\scriptscriptstyle .007}_{-\scriptscriptstyle .007}$} & \bcell{29}{$.709^{+\scriptscriptstyle .007}_{-\scriptscriptstyle .007}$} & \bcell{50}{$\textbf{.739}^{+\scriptscriptstyle .009}_{-\scriptscriptstyle .009}$} \\
\addlinespace
\grayrow \multicolumn{7}{l}{\Cardio \textit{\textbf{Cardiovascular}}} \\
Cardiovascular Dx & \Binary & \ROCAUCLabel & \bcell{16}{$.667^{+\scriptscriptstyle .148}_{-\scriptscriptstyle .190}$} & \bcell{10}{$.658^{+\scriptscriptstyle .081}_{-\scriptscriptstyle .091}$} & \bcell{5}{$.651^{+\scriptscriptstyle .110}_{-\scriptscriptstyle .129}$} & \bcell{50}{$\textbf{.712}^{+\scriptscriptstyle .052}_{-\scriptscriptstyle .058}$} \\
Hypertension Dx & \Binary & \ROCAUCLabel & \bcell{5}{$.698^{+\scriptscriptstyle .020}_{-\scriptscriptstyle .021}$} & \bcell{24}{$.737^{+\scriptscriptstyle .026}_{-\scriptscriptstyle .027}$} & \bcell{40}{$.768^{+\scriptscriptstyle .027}_{-\scriptscriptstyle .030}$} & \bcell{50}{$\textbf{.786}^{+\scriptscriptstyle .023}_{-\scriptscriptstyle .025}$} \\
Respiratory Dx & \Binary & \ROCAUCLabel & \bcell{5}{$.593^{+\scriptscriptstyle .057}_{-\scriptscriptstyle .060}$} & \bcell{20}{$.624^{+\scriptscriptstyle .064}_{-\scriptscriptstyle .069}$} & \bcell{39}{$.662^{+\scriptscriptstyle .048}_{-\scriptscriptstyle .052}$} & \bcell{50}{$\textbf{.682}^{+\scriptscriptstyle .054}_{-\scriptscriptstyle .059}$} \\
ASCVD Risk & \RegressionLabel & \PearsonLabel & \bcell{5}{$.542^{+\scriptscriptstyle .080}_{-\scriptscriptstyle .092}$} & \bcell{24}{$.623^{+\scriptscriptstyle .096}_{-\scriptscriptstyle .119}$} & \bcell{36}{$.675^{+\scriptscriptstyle .093}_{-\scriptscriptstyle .121}$} & \bcell{50}{$\textbf{.730}^{+\scriptscriptstyle .091}_{-\scriptscriptstyle .127}$} \\
Framingham Risk & \RegressionLabel & \PearsonLabel & \bcell{5}{$.496^{+\scriptscriptstyle .040}_{-\scriptscriptstyle .042}$} & \bcell{25}{$.574^{+\scriptscriptstyle .087}_{-\scriptscriptstyle .102}$} & \bcell{45}{$.651^{+\scriptscriptstyle .101}_{-\scriptscriptstyle .132}$} & \bcell{50}{$\textbf{.669}^{+\scriptscriptstyle .101}_{-\scriptscriptstyle .133}$} \\
Framingham 30 Risk & \RegressionLabel & \PearsonLabel & \bcell{5}{$.454^{+\scriptscriptstyle .023}_{-\scriptscriptstyle .024}$} & \bcell{26}{$.580^{+\scriptscriptstyle .078}_{-\scriptscriptstyle .090}$} & \bcell{42}{$.668^{+\scriptscriptstyle .069}_{-\scriptscriptstyle .082}$} & \bcell{50}{$\textbf{.714}^{+\scriptscriptstyle .052}_{-\scriptscriptstyle .062}$} \\
\addlinespace
\grayrow \multicolumn{7}{l}{\textit{\Metabolic \textbf{Metabolic}}} \\
Diabetes Dx & \Binary & \ROCAUCLabel & \bcell{13}{$.688^{+\scriptscriptstyle .052}_{-\scriptscriptstyle .058}$} & \bcell{5}{$.669^{+\scriptscriptstyle .041}_{-\scriptscriptstyle .044}$} & \bcell{31}{$.724^{+\scriptscriptstyle .039}_{-\scriptscriptstyle .043}$} & \bcell{50}{$\textbf{.763}^{+\scriptscriptstyle .037}_{-\scriptscriptstyle .042}$} \\
Diabetes Med. & \Binary & \ROCAUCLabel & \bcell{5}{$.646^{+\scriptscriptstyle .062}_{-\scriptscriptstyle .067}$} & \bcell{26}{$.672^{+\scriptscriptstyle .037}_{-\scriptscriptstyle .039}$} & \bcell{44}{$.693^{+\scriptscriptstyle .031}_{-\scriptscriptstyle .033}$} & \bcell{50}{$\textbf{.700}^{+\scriptscriptstyle .045}_{-\scriptscriptstyle .049}$} \\
Hyperlipidemia & \Binary & \ROCAUCLabel & \bcell{5}{$.623^{+\scriptscriptstyle .041}_{-\scriptscriptstyle .043}$} & \bcell{12}{$.632^{+\scriptscriptstyle .040}_{-\scriptscriptstyle .042}$} & \bcell{42}{$.666^{+\scriptscriptstyle .038}_{-\scriptscriptstyle .041}$} & \bcell{50}{$\textbf{.674}^{+\scriptscriptstyle .032}_{-\scriptscriptstyle .033}$} \\
Pre-Diabetes & \Binary & \ROCAUCLabel & \bcell{5}{$.636^{+\scriptscriptstyle .070}_{-\scriptscriptstyle .076}$} & \bcell{19}{$.658^{+\scriptscriptstyle .015}_{-\scriptscriptstyle .016}$} & \bcell{50}{$\textbf{.706}^{+\scriptscriptstyle .020}_{-\scriptscriptstyle .021}$} & \bcell{48}{$.704^{+\scriptscriptstyle .068}_{-\scriptscriptstyle .078}$} \\
Insulin Resistance & \Binary & \ROCAUCLabel & \bcell{5}{$.635^{+\scriptscriptstyle .060}_{-\scriptscriptstyle .064}$} & \bcell{7}{$.641^{+\scriptscriptstyle .062}_{-\scriptscriptstyle .067}$} & \bcell{22}{$.683^{+\scriptscriptstyle .024}_{-\scriptscriptstyle .026}$} & \bcell{50}{$\textbf{.761}^{+\scriptscriptstyle .044}_{-\scriptscriptstyle .050}$} \\
HOMA-IR & \RegressionLabel & \PearsonLabel & \bcell{6}{$.315^{+\scriptscriptstyle .082}_{-\scriptscriptstyle .087}$} & \bcell{5}{$.309^{+\scriptscriptstyle .091}_{-\scriptscriptstyle .097}$} & \bcell{26}{$.391^{+\scriptscriptstyle .033}_{-\scriptscriptstyle .034}$} & \bcell{50}{$\textbf{.479}^{+\scriptscriptstyle .030}_{-\scriptscriptstyle .031}$} \\
HbA1c & \RegressionLabel & \PearsonLabel & \bcell{5}{$.137^{+\scriptscriptstyle .091}_{-\scriptscriptstyle .093}$} & \bcell{24}{$.204^{+\scriptscriptstyle .056}_{-\scriptscriptstyle .057}$} & \bcell{28}{$.219^{+\scriptscriptstyle .081}_{-\scriptscriptstyle .084}$} & \bcell{50}{$\textbf{.293}^{+\scriptscriptstyle .029}_{-\scriptscriptstyle .030}$} \\
Triglycerides & \RegressionLabel & \PearsonLabel & \bcell{5}{$.088^{+\scriptscriptstyle .084}_{-\scriptscriptstyle .085}$} & \bcell{5}{$.091^{+\scriptscriptstyle .056}_{-\scriptscriptstyle .057}$} & \bcell{34}{$.207^{+\scriptscriptstyle .036}_{-\scriptscriptstyle .037}$} & \bcell{50}{$\textbf{.269}^{+\scriptscriptstyle .047}_{-\scriptscriptstyle .048}$} \\
\addlinespace
\grayrow \multicolumn{7}{l}{\Mental \textit{\textbf{Mental Health}}} \\
Mild Depression & \Binary & \ROCAUCLabel & \bcell{5}{$.661^{+\scriptscriptstyle .016}_{-\scriptscriptstyle .016}$} & \bcell{6}{$.663^{+\scriptscriptstyle .017}_{-\scriptscriptstyle .017}$} & \bcell{31}{$.699^{+\scriptscriptstyle .008}_{-\scriptscriptstyle .008}$} & \bcell{50}{$\textbf{.726}^{+\scriptscriptstyle .006}_{-\scriptscriptstyle .006}$} \\
Mild Anxiety & \Binary & \ROCAUCLabel & \bcell{5}{$.634^{+\scriptscriptstyle .013}_{-\scriptscriptstyle .013}$} & \bcell{8}{$.639^{+\scriptscriptstyle .022}_{-\scriptscriptstyle .023}$} & \bcell{33}{$.674^{+\scriptscriptstyle .012}_{-\scriptscriptstyle .012}$} & \bcell{50}{$\textbf{.698}^{+\scriptscriptstyle .005}_{-\scriptscriptstyle .005}$} \\
Persistent Stress & \Binary & \ROCAUCLabel & \bcell{5}{$.659^{+\scriptscriptstyle .021}_{-\scriptscriptstyle .021}$} & \bcell{6}{$.660^{+\scriptscriptstyle .028}_{-\scriptscriptstyle .029}$} & \bcell{30}{$.689^{+\scriptscriptstyle .030}_{-\scriptscriptstyle .031}$} & \bcell{50}{$\textbf{.712}^{+\scriptscriptstyle .026}_{-\scriptscriptstyle .028}$} \\
Depress./Anxiety Dx & \Binary & \ROCAUCLabel & \bcell{5}{$.660^{+\scriptscriptstyle .016}_{-\scriptscriptstyle .016}$} & \bcell{14}{$.672^{+\scriptscriptstyle .015}_{-\scriptscriptstyle .015}$} & \bcell{33}{$.696^{+\scriptscriptstyle .017}_{-\scriptscriptstyle .017}$} & \bcell{50}{$\textbf{.717}^{+\scriptscriptstyle .009}_{-\scriptscriptstyle .009}$} \\
Mental Health Med. & \Binary & \ROCAUCLabel & \bcell{5}{$.755^{+\scriptscriptstyle .018}_{-\scriptscriptstyle .019}$} & \bcell{13}{$.767^{+\scriptscriptstyle .011}_{-\scriptscriptstyle .011}$} & \bcell{28}{$.789^{+\scriptscriptstyle .014}_{-\scriptscriptstyle .014}$} & \bcell{50}{$\textbf{.819}^{+\scriptscriptstyle .020}_{-\scriptscriptstyle .022}$} \\
PHQ-8 & \RegressionLabel & \PearsonLabel & \bcell{5}{$.322^{+\scriptscriptstyle .029}_{-\scriptscriptstyle .030}$} & \bcell{12}{$.344^{+\scriptscriptstyle .017}_{-\scriptscriptstyle .017}$} & \bcell{31}{$.397^{+\scriptscriptstyle .012}_{-\scriptscriptstyle .012}$} & \bcell{50}{$\textbf{.450}^{+\scriptscriptstyle .019}_{-\scriptscriptstyle .019}$} \\
GAD-7 & \RegressionLabel & \PearsonLabel & \bcell{5}{$.273^{+\scriptscriptstyle .025}_{-\scriptscriptstyle .025}$} & \bcell{12}{$.293^{+\scriptscriptstyle .013}_{-\scriptscriptstyle .014}$} & \bcell{31}{$.346^{+\scriptscriptstyle .011}_{-\scriptscriptstyle .011}$} & \bcell{50}{$\textbf{.400}^{+\scriptscriptstyle .025}_{-\scriptscriptstyle .025}$} \\
PSS & \RegressionLabel & \PearsonLabel & \bcell{5}{$.343^{+\scriptscriptstyle .018}_{-\scriptscriptstyle .018}$} & \bcell{9}{$.355^{+\scriptscriptstyle .020}_{-\scriptscriptstyle .021}$} & \bcell{32}{$.415^{+\scriptscriptstyle .016}_{-\scriptscriptstyle .016}$} & \bcell{50}{$\textbf{.463}^{+\scriptscriptstyle .013}_{-\scriptscriptstyle .014}$} \\
\addlinespace
\grayrow \multicolumn{7}{l}{\Sleep \textit{\textbf{Sleep}}} \\
Sleep Disorder Treatment & \Binary & \ROCAUCLabel & \bcell{5}{$.611^{+\scriptscriptstyle .028}_{-\scriptscriptstyle .028}$} & \bcell{30}{$.633^{+\scriptscriptstyle .033}_{-\scriptscriptstyle .034}$} & \bcell{50}{$\textbf{.651}^{+\scriptscriptstyle .050}_{-\scriptscriptstyle .053}$} & \bcell{47}{$.649^{+\scriptscriptstyle .036}_{-\scriptscriptstyle .038}$} \\
Sleep Disturbance PRO & \RegressionLabel & \PearsonLabel & \bcell{5}{$.284^{+\scriptscriptstyle .015}_{-\scriptscriptstyle .016}$} & \bcell{18}{$.315^{+\scriptscriptstyle .020}_{-\scriptscriptstyle .020}$} & \bcell{33}{$.352^{+\scriptscriptstyle .008}_{-\scriptscriptstyle .008}$} & \bcell{50}{$\textbf{.390}^{+\scriptscriptstyle .016}_{-\scriptscriptstyle .016}$} \\
Sleep Impairment PRO & \RegressionLabel & \PearsonLabel & \bcell{5}{$.346^{+\scriptscriptstyle .019}_{-\scriptscriptstyle .019}$} & \bcell{8}{$.355^{+\scriptscriptstyle .013}_{-\scriptscriptstyle .013}$} & \bcell{30}{$.404^{+\scriptscriptstyle .015}_{-\scriptscriptstyle .015}$} & \bcell{50}{$\textbf{.448}^{+\scriptscriptstyle .013}_{-\scriptscriptstyle .013}$} \\
\addlinespace
\bottomrule
\end{tabular}
\label{tab:predictive_task_scaling}
\end{table}

%% file: Tables_Final/tbED7_pred_scaling_II.tex

\begin{table}[!htbp]
\centering
\footnotesize
\caption{\textbf{Discriminative Task Performance Across Model Scales (Part II: Additional Metrics).}
This table extends the results presented in Part I.
The Table presents the performance of \modelname variants, pretrained with proportional data scales, on 35 discriminative tasks. In general performance improves with scale with B consistently achieving the best performance. 
\modelname variants are post-trained with PCA-50 reduced embeddings. 
For each task, we report the average 5-fold cross validation performance. For both F1 and MAE we leverage an arithmetic mean and standard deviation across folds.
Colors are normalized per row; best model performance is bolded and has the deepest shade.}

\begin{tabular}{lccrrrr}
\toprule
\textbf{Prediction Task} & \textbf{Type} & \textbf{Metric}
& \multicolumn{4}{c}{\textbf{Model Variant (Parameter Count)}} \\ 
\cmidrule(lr){4-7}
&&& XXS (10$^{5}$) & XS (10$^{6}$) & S (10$^{7}$) & B (10$^{8}$) \\
\midrule
\grayrow \multicolumn{7}{l}{\textit{\Demo \textbf{Demographics}}} \\
Age & \RegressionLabel & \MAELabel & \bcell{5}{$7.035^{+\scriptscriptstyle .135}_{-\scriptscriptstyle .135}$} & \bcell{12}{$6.539^{+\scriptscriptstyle .101}_{-\scriptscriptstyle .101}$} & \bcell{28}{$5.405^{+\scriptscriptstyle .082}_{-\scriptscriptstyle .082}$} & \bcell{50}{$\textbf{3.865}^{+\scriptscriptstyle .075}_{-\scriptscriptstyle .075}$} \\
BMI & \RegressionLabel & \MAELabel & \bcell{5}{$5.090^{+\scriptscriptstyle .062}_{-\scriptscriptstyle .062}$} & \bcell{9}{$4.897^{+\scriptscriptstyle .053}_{-\scriptscriptstyle .053}$} & \bcell{31}{$3.937^{+\scriptscriptstyle .068}_{-\scriptscriptstyle .068}$} & \bcell{50}{$\textbf{3.148}^{+\scriptscriptstyle .024}_{-\scriptscriptstyle .024}$} \\
Height & \RegressionLabel & \MAELabel & \bcell{5}{$71.226^{+\scriptscriptstyle .499}_{-\scriptscriptstyle .499}$} & \bcell{11}{$69.460^{+\scriptscriptstyle .348}_{-\scriptscriptstyle .348}$} & \bcell{38}{$61.985^{+\scriptscriptstyle .784}_{-\scriptscriptstyle .784}$} & \bcell{50}{$\textbf{58.773}^{+\scriptscriptstyle .829}_{-\scriptscriptstyle .829}$} \\
Weight & \RegressionLabel & \MAELabel & \bcell{5}{$15.544^{+\scriptscriptstyle .121}_{-\scriptscriptstyle .121}$} & \bcell{10}{$14.754^{+\scriptscriptstyle .102}_{-\scriptscriptstyle .102}$} & \bcell{33}{$11.682^{+\scriptscriptstyle .210}_{-\scriptscriptstyle .210}$} & \bcell{50}{$\textbf{9.541}^{+\scriptscriptstyle .137}_{-\scriptscriptstyle .137}$} \\
\addlinespace
\grayrow \multicolumn{7}{l}{\textit{\Life \textbf{Lifestyle}}} \\
Currently Working & \Binary & \FOneLabel & \bcell{5}{$.805^{+\scriptscriptstyle .002}_{-\scriptscriptstyle .002}$} & \bcell{9}{$.812^{+\scriptscriptstyle .007}_{-\scriptscriptstyle .007}$} & \bcell{23}{$.838^{+\scriptscriptstyle .005}_{-\scriptscriptstyle .005}$} & \bcell{50}{$\textbf{.884}^{+\scriptscriptstyle .009}_{-\scriptscriptstyle .009}$} \\
Disability & \Binary & \FOneLabel & \bcell{5}{$.380^{+\scriptscriptstyle .016}_{-\scriptscriptstyle .016}$} & \bcell{19}{$.396^{+\scriptscriptstyle .012}_{-\scriptscriptstyle .012}$} & \bcell{24}{$.402^{+\scriptscriptstyle .018}_{-\scriptscriptstyle .018}$} & \bcell{50}{$\textbf{.432}^{+\scriptscriptstyle .033}_{-\scriptscriptstyle .033}$} \\
Disability Affects Work & \Binary & \FOneLabel & \bcell{5}{$.641^{+\scriptscriptstyle .032}_{-\scriptscriptstyle .032}$} & \bcell{30}{$.690^{+\scriptscriptstyle .031}_{-\scriptscriptstyle .031}$} & \bcell{25}{$.681^{+\scriptscriptstyle .046}_{-\scriptscriptstyle .046}$} & \bcell{50}{$\textbf{.727}^{+\scriptscriptstyle .023}_{-\scriptscriptstyle .023}$} \\
Smoking & \Binary & \FOneLabel & \bcell{5}{$.236^{+\scriptscriptstyle .059}_{-\scriptscriptstyle .059}$} & \bcell{10}{$.259^{+\scriptscriptstyle .065}_{-\scriptscriptstyle .065}$} & \bcell{16}{$.283^{+\scriptscriptstyle .054}_{-\scriptscriptstyle .054}$} & \bcell{50}{$\textbf{.416}^{+\scriptscriptstyle .088}_{-\scriptscriptstyle .088}$} \\
Medicaid & \Binary & \FOneLabel & \bcell{5}{$.312^{+\scriptscriptstyle .021}_{-\scriptscriptstyle .021}$} & \bcell{13}{$.333^{+\scriptscriptstyle .027}_{-\scriptscriptstyle .027}$} & \bcell{29}{$.367^{+\scriptscriptstyle .029}_{-\scriptscriptstyle .029}$} & \bcell{50}{$\textbf{.416}^{+\scriptscriptstyle .021}_{-\scriptscriptstyle .021}$} \\
No Medications & \Binary & \FOneLabel & \bcell{5}{$.612^{+\scriptscriptstyle .011}_{-\scriptscriptstyle .011}$} & \bcell{9}{$.618^{+\scriptscriptstyle .011}_{-\scriptscriptstyle .011}$} & \bcell{28}{$.641^{+\scriptscriptstyle .006}_{-\scriptscriptstyle .006}$} & \bcell{50}{$\textbf{.667}^{+\scriptscriptstyle .006}_{-\scriptscriptstyle .006}$} \\
\addlinespace
\grayrow \multicolumn{7}{l}{\textit{\Cardio \textbf{Cardiovascular}}} \\
Cardiovascular Dx & \Binary & \FOneLabel & \bcell{22}{$.091^{+\scriptscriptstyle .040}_{-\scriptscriptstyle .040}$} & \bcell{20}{$.090^{+\scriptscriptstyle .043}_{-\scriptscriptstyle .043}$} & \bcell{5}{$.081^{+\scriptscriptstyle .051}_{-\scriptscriptstyle .051}$} & \bcell{50}{$\textbf{.108}^{+\scriptscriptstyle .061}_{-\scriptscriptstyle .061}$} \\
Hypertension Dx & \Binary & \FOneLabel & \bcell{5}{$.481^{+\scriptscriptstyle .034}_{-\scriptscriptstyle .034}$} & \bcell{21}{$.510^{+\scriptscriptstyle .043}_{-\scriptscriptstyle .043}$} & \bcell{33}{$.531^{+\scriptscriptstyle .042}_{-\scriptscriptstyle .042}$} & \bcell{50}{$\textbf{.560}^{+\scriptscriptstyle .032}_{-\scriptscriptstyle .032}$} \\
Respiratory Dx & \Binary & \FOneLabel & \bcell{5}{$.313^{+\scriptscriptstyle .058}_{-\scriptscriptstyle .058}$} & \bcell{16}{$.329^{+\scriptscriptstyle .055}_{-\scriptscriptstyle .055}$} & \bcell{33}{$.353^{+\scriptscriptstyle .060}_{-\scriptscriptstyle .060}$} & \bcell{50}{$\textbf{.375}^{+\scriptscriptstyle .069}_{-\scriptscriptstyle .069}$} \\
ASCVD Risk & \RegressionLabel & \MAELabel & \bcell{5}{$.037^{+\scriptscriptstyle .002}_{-\scriptscriptstyle .002}$} & \bcell{19}{$.034^{+\scriptscriptstyle .005}_{-\scriptscriptstyle .005}$} & \bcell{31}{$.032^{+\scriptscriptstyle .003}_{-\scriptscriptstyle .003}$} & \bcell{50}{$\textbf{.029}^{+\scriptscriptstyle .004}_{-\scriptscriptstyle .004}$} \\
Framingham Risk & \RegressionLabel & \MAELabel & \bcell{5}{$.044^{+\scriptscriptstyle .003}_{-\scriptscriptstyle .003}$} & \bcell{25}{$.040^{+\scriptscriptstyle .006}_{-\scriptscriptstyle .006}$} & \bcell{40}{$.037^{+\scriptscriptstyle .005}_{-\scriptscriptstyle .005}$} & \bcell{50}{$\textbf{.035}^{+\scriptscriptstyle .005}_{-\scriptscriptstyle .005}$} \\
Framingham 30 Risk & \RegressionLabel & \MAELabel & \bcell{5}{$.124^{+\scriptscriptstyle .010}_{-\scriptscriptstyle .010}$} & \bcell{22}{$.111^{+\scriptscriptstyle .011}_{-\scriptscriptstyle .011}$} & \bcell{38}{$.100^{+\scriptscriptstyle .009}_{-\scriptscriptstyle .009}$} & \bcell{50}{$\textbf{.091}^{+\scriptscriptstyle .010}_{-\scriptscriptstyle .010}$} \\
\addlinespace
\grayrow \multicolumn{7}{l}{\textit{\Metabolic \textbf{Metabolic}}} \\
Diabetes Dx & \Binary & \FOneLabel & \bcell{14}{$.238^{+\scriptscriptstyle .026}_{-\scriptscriptstyle .026}$} & \bcell{5}{$.222^{+\scriptscriptstyle .033}_{-\scriptscriptstyle .033}$} & \bcell{30}{$.266^{+\scriptscriptstyle .020}_{-\scriptscriptstyle .020}$} & \bcell{50}{$\textbf{.300}^{+\scriptscriptstyle .038}_{-\scriptscriptstyle .038}$} \\
Diabetes Med. & \Binary & \FOneLabel & \bcell{20}{$.225^{+\scriptscriptstyle .044}_{-\scriptscriptstyle .044}$} & \bcell{5}{$.208^{+\scriptscriptstyle .028}_{-\scriptscriptstyle .028}$} & \bcell{23}{$.229^{+\scriptscriptstyle .029}_{-\scriptscriptstyle .029}$} & \bcell{50}{$\textbf{.259}^{+\scriptscriptstyle .022}_{-\scriptscriptstyle .022}$} \\
Hyperlipidemia & \Binary & \FOneLabel & \bcell{5}{$.387^{+\scriptscriptstyle .045}_{-\scriptscriptstyle .045}$} & \bcell{18}{$.397^{+\scriptscriptstyle .045}_{-\scriptscriptstyle .045}$} & \bcell{46}{$.415^{+\scriptscriptstyle .057}_{-\scriptscriptstyle .057}$} & \bcell{50}{$\textbf{.418}^{+\scriptscriptstyle .047}_{-\scriptscriptstyle .047}$} \\
Pre-Diabetes & \Binary & \FOneLabel & \bcell{5}{$.416^{+\scriptscriptstyle .031}_{-\scriptscriptstyle .031}$} & \bcell{40}{$.469^{+\scriptscriptstyle .068}_{-\scriptscriptstyle .068}$} & \bcell{40}{$.469^{+\scriptscriptstyle .053}_{-\scriptscriptstyle .053}$} & \bcell{50}{$\textbf{.483}^{+\scriptscriptstyle .047}_{-\scriptscriptstyle .047}$} \\
Insulin Resistance & \Binary & \FOneLabel & \bcell{5}{$.442^{+\scriptscriptstyle .048}_{-\scriptscriptstyle .048}$} & \bcell{11}{$.457^{+\scriptscriptstyle .055}_{-\scriptscriptstyle .055}$} & \bcell{22}{$.485^{+\scriptscriptstyle .032}_{-\scriptscriptstyle .032}$} & \bcell{50}{$\textbf{.557}^{+\scriptscriptstyle .053}_{-\scriptscriptstyle .053}$} \\
HOMA-IR & \RegressionLabel & \MAELabel & \bcell{5}{$1.314^{+\scriptscriptstyle .088}_{-\scriptscriptstyle .088}$} & \bcell{5}{$1.311^{+\scriptscriptstyle .101}_{-\scriptscriptstyle .101}$} & \bcell{29}{$1.234^{+\scriptscriptstyle .037}_{-\scriptscriptstyle .037}$} & \bcell{50}{$\textbf{1.165}^{+\scriptscriptstyle .056}_{-\scriptscriptstyle .056}$} \\
HbA1c & \RegressionLabel & \MAELabel & \bcell{5}{$.384^{+\scriptscriptstyle .031}_{-\scriptscriptstyle .031}$} & \bcell{45}{$.372^{+\scriptscriptstyle .030}_{-\scriptscriptstyle .030}$} & \bcell{43}{$.372^{+\scriptscriptstyle .025}_{-\scriptscriptstyle .025}$} & \bcell{50}{$\textbf{.370}^{+\scriptscriptstyle .016}_{-\scriptscriptstyle .016}$} \\
Triglycerides & \RegressionLabel & \MAELabel & \bcell{5}{$51.480^{+\scriptscriptstyle 3.443}_{-\scriptscriptstyle 3.443}$} & \bcell{21}{$50.132^{+\scriptscriptstyle 3.860}_{-\scriptscriptstyle 3.860}$} & \bcell{44}{$48.219^{+\scriptscriptstyle 3.570}_{-\scriptscriptstyle 3.570}$} & \bcell{50}{$\textbf{47.769}^{+\scriptscriptstyle 3.613}_{-\scriptscriptstyle 3.613}$} \\
\addlinespace
\grayrow \multicolumn{7}{l}{\textit{\Mental \textbf{Mental Health}}} \\
Mild Depression & \Binary & \FOneLabel & \bcell{9}{$.487^{+\scriptscriptstyle .023}_{-\scriptscriptstyle .023}$} & \bcell{5}{$.482^{+\scriptscriptstyle .021}_{-\scriptscriptstyle .021}$} & \bcell{28}{$.512^{+\scriptscriptstyle .014}_{-\scriptscriptstyle .014}$} & \bcell{50}{$\textbf{.539}^{+\scriptscriptstyle .016}_{-\scriptscriptstyle .016}$} \\
Mild Anxiety & \Binary & \FOneLabel & \bcell{9}{$.413^{+\scriptscriptstyle .025}_{-\scriptscriptstyle .025}$} & \bcell{5}{$.407^{+\scriptscriptstyle .020}_{-\scriptscriptstyle .020}$} & \bcell{34}{$.442^{+\scriptscriptstyle .024}_{-\scriptscriptstyle .024}$} & \bcell{50}{$\textbf{.461}^{+\scriptscriptstyle .023}_{-\scriptscriptstyle .023}$} \\
Persistent Stress & \Binary & \FOneLabel & \bcell{10}{$.684^{+\scriptscriptstyle .022}_{-\scriptscriptstyle .022}$} & \bcell{5}{$.678^{+\scriptscriptstyle .021}_{-\scriptscriptstyle .021}$} & \bcell{30}{$.700^{+\scriptscriptstyle .019}_{-\scriptscriptstyle .019}$} & \bcell{50}{$\textbf{.717}^{+\scriptscriptstyle .022}_{-\scriptscriptstyle .022}$} \\
Depress./Anxiety Dx & \Binary & \FOneLabel & \bcell{5}{$.414^{+\scriptscriptstyle .018}_{-\scriptscriptstyle .018}$} & \bcell{18}{$.429^{+\scriptscriptstyle .015}_{-\scriptscriptstyle .015}$} & \bcell{33}{$.444^{+\scriptscriptstyle .027}_{-\scriptscriptstyle .027}$} & \bcell{50}{$\textbf{.461}^{+\scriptscriptstyle .016}_{-\scriptscriptstyle .016}$} \\
Mental Health Med. & \Binary & \FOneLabel & \bcell{5}{$.353^{+\scriptscriptstyle .017}_{-\scriptscriptstyle .017}$} & \bcell{9}{$.358^{+\scriptscriptstyle .019}_{-\scriptscriptstyle .019}$} & \bcell{20}{$.369^{+\scriptscriptstyle .021}_{-\scriptscriptstyle .021}$} & \bcell{50}{$\textbf{.401}^{+\scriptscriptstyle .023}_{-\scriptscriptstyle .023}$} \\
PHQ-8 & \RegressionLabel & \MAELabel & \bcell{5}{$4.228^{+\scriptscriptstyle .062}_{-\scriptscriptstyle .062}$} & \bcell{10}{$4.196^{+\scriptscriptstyle .064}_{-\scriptscriptstyle .064}$} & \bcell{29}{$4.072^{+\scriptscriptstyle .070}_{-\scriptscriptstyle .070}$} & \bcell{50}{$\textbf{3.948}^{+\scriptscriptstyle .078}_{-\scriptscriptstyle .078}$} \\
GAD-7 & \RegressionLabel & \MAELabel & \bcell{5}{$3.951^{+\scriptscriptstyle .078}_{-\scriptscriptstyle .078}$} & \bcell{10}{$3.925^{+\scriptscriptstyle .079}_{-\scriptscriptstyle .079}$} & \bcell{30}{$3.827^{+\scriptscriptstyle .075}_{-\scriptscriptstyle .075}$} & \bcell{50}{$\textbf{3.732}^{+\scriptscriptstyle .085}_{-\scriptscriptstyle .085}$} \\
PSS & \RegressionLabel & \MAELabel & \bcell{5}{$5.618^{+\scriptscriptstyle .040}_{-\scriptscriptstyle .040}$} & \bcell{7}{$5.602^{+\scriptscriptstyle .076}_{-\scriptscriptstyle .076}$} & \bcell{30}{$5.424^{+\scriptscriptstyle .063}_{-\scriptscriptstyle .063}$} & \bcell{50}{$\textbf{5.272}^{+\scriptscriptstyle .048}_{-\scriptscriptstyle .048}$} \\
\addlinespace
\grayrow \multicolumn{7}{l}{\textit{\Sleep \textbf{Sleep}}} \\
Sleep Disorder Treatment & \Binary & \FOneLabel & \bcell{5}{$.660^{+\scriptscriptstyle .026}_{-\scriptscriptstyle .026}$} & \bcell{50}{$\textbf{.684}^{+\scriptscriptstyle .037}_{-\scriptscriptstyle .037}$} & \bcell{44}{$.681^{+\scriptscriptstyle .038}_{-\scriptscriptstyle .038}$} & \bcell{47}{$.683^{+\scriptscriptstyle .034}_{-\scriptscriptstyle .034}$} \\
Sleep Disturbance PRO & \RegressionLabel & \MAELabel & \bcell{5}{$5.440^{+\scriptscriptstyle .109}_{-\scriptscriptstyle .109}$} & \bcell{16}{$5.380^{+\scriptscriptstyle .096}_{-\scriptscriptstyle .096}$} & \bcell{29}{$5.312^{+\scriptscriptstyle .097}_{-\scriptscriptstyle .097}$} & \bcell{50}{$\textbf{5.205}^{+\scriptscriptstyle .084}_{-\scriptscriptstyle .084}$} \\
Sleep Impairment PRO & \RegressionLabel & \MAELabel & \bcell{5}{$5.879^{+\scriptscriptstyle .125}_{-\scriptscriptstyle .125}$} & \bcell{9}{$5.844^{+\scriptscriptstyle .122}_{-\scriptscriptstyle .122}$} & \bcell{30}{$5.687^{+\scriptscriptstyle .121}_{-\scriptscriptstyle .121}$} & \bcell{50}{$\textbf{5.538}^{+\scriptscriptstyle .112}_{-\scriptscriptstyle .112}$} \\
\addlinespace
\bottomrule
\end{tabular}
\label{tab:predictive_task_scaling2}
\end{table}

%% file: Tables_Final/tbED8_pred_scaling_effect_of_demos.tex

\begin{table}[!htbp]
\centering
\footnotesize
\caption{\textbf{Discriminative Task Improvement Due to Demographic Features.}
The table presents the mean improvement in task performance caused by the inclusion of demographic features across \modelname variants and a supervised baseline trained with engineered features. In general the effect of demographic features lessens with scale with B exhibiting the lowest change in performance on $33$ of $35$ tasks.
\modelname variants are post-trained with PCA-50 reduced embeddings. 
For each task, we report the average 5-fold cross validation performance. Average Receiver Operating Characteristic Area Under the Curve (ROC AUC) is calculated in the logit-transform space and back-transformed. Average Pearson correlation ($r$) is calculated in the z-transform space and back-transformed. Standard deviations are calculated in the transformed space and back-transformed to give asymmetric error values. Colors are normalized per row; lowest values are bolded and have the \emph{lightest} shade.}

\begin{tabular}{lccccccc}
\toprule

\textbf{Prediction Task} & \textbf{Type} & \textbf{Metric}
& \multicolumn{4}{c}{\textbf{Model Variant (Parameter Count)}} \\ 
\cmidrule(lr){4-8}
&&& Feat. Eng. & XXS (10$^{5}$) & XS (10$^{6}$) & S (10$^{7}$) & B (10$^{8}$) \\

\midrule
\grayrow \multicolumn{8}{l}{\textit{\Life \textbf{Lifestyle}}} \\
Currently Working & \Binary & \DeltaROCAUCLabel & \bcell{47}{$.014$} & \bcell{50}{$.014$} & \bcell{33}{$.009$} & \bcell{18}{$.004$} & \bcell{5}{$\textbf{.000}$} \\
Disability & \Binary & \DeltaROCAUCLabel & \bcell{39}{$.020$} & \bcell{50}{$.024$} & \bcell{44}{$.022$} & \bcell{25}{$.014$} & \bcell{5}{$\textbf{.005}$} \\
Disability Affects Work & \Binary & \DeltaROCAUCLabel & \bcell{37}{$.020$} & \bcell{50}{$.027$} & \bcell{30}{$.015$} & \bcell{28}{$.014$} & \bcell{5}{$\textbf{-.001}$} \\
Smoking & \Binary & \DeltaROCAUCLabel & \bcell{37}{$.017$} & \bcell{15}{$.004$} & \bcell{50}{$.025$} & \bcell{20}{$.007$} & \bcell{5}{$\textbf{-.003}$} \\
Medicaid & \Binary & \DeltaROCAUCLabel & \bcell{27}{$.003$} & \bcell{50}{$.008$} & \bcell{33}{$.004$} & \bcell{16}{$.002$} & \bcell{5}{$\textbf{-.001}$} \\
No Medications & \Binary & \DeltaROCAUCLabel & \bcell{44}{$.012$} & \bcell{50}{$.012$} & \bcell{40}{$.011$} & \bcell{16}{$.007$} & \bcell{5}{$\textbf{.006}$} \\
\addlinespace
\grayrow \multicolumn{8}{l}{\textit{\Cardio \textbf{Cardiovascular}}} \\
Cardiovascular Dx & \Binary & \DeltaROCAUCLabel & \bcell{21}{$.010$} & \bcell{50}{$.054$} & \bcell{25}{$.015$} & \bcell{27}{$.018$} & \bcell{5}{$\textbf{-.016}$} \\
Hypertension Dx & \Binary & \DeltaROCAUCLabel & \bcell{34}{$.033$} & \bcell{50}{$.050$} & \bcell{26}{$.024$} & \bcell{16}{$.012$} & \bcell{5}{$\textbf{.000}$} \\
Respiratory Dx & \Binary & \DeltaROCAUCLabel & \bcell{17}{$-0.002$} & \bcell{50}{$.011$} & \bcell{29}{$.003$} & \bcell{5}{$\textbf{-0.007}$} & \bcell{5}{$\textbf{-.007}$} \\
ASCVD Risk & \RegressionLabel & \DeltaPearsonLabel & \bcell{47}{$.159$} & \bcell{50}{$.164$} & \bcell{33}{$.124$} & \bcell{24}{$.102$} & \bcell{5}{$\textbf{.054}$} \\
Framingham Risk & \RegressionLabel & \DeltaPearsonLabel & \bcell{44}{$.182$} & \bcell{50}{$.200$} & \bcell{35}{$.154$} & \bcell{18}{$.098$} & \bcell{5}{$\textbf{.055}$} \\
Framingham 30 Risk & \RegressionLabel & \DeltaPearsonLabel & \bcell{30}{$.181$} & \bcell{50}{$.283$} & \bcell{28}{$.168$} & \bcell{15}{$.097$} & \bcell{5}{$\textbf{.042}$} \\
\addlinespace
\grayrow \multicolumn{8}{l}{\textit{\Metabolic \textbf{Metabolic}}} \\
Diabetes Dx & \Binary & \DeltaROCAUCLabel & \bcell{27}{$.006$} & \bcell{50}{$.025$} & \bcell{32}{$.011$} & \bcell{5}{$\textbf{-.012}$} & \bcell{10}{$-.007$} \\
Diabetes Med. & \Binary & \DeltaROCAUCLabel & \bcell{34}{$.035$} & \bcell{50}{$.050$} & \bcell{32}{$.032$} & \bcell{15}{$.016$} & \bcell{5}{$\textbf{.006}$} \\
Hyperlipidemia & \Binary & \DeltaROCAUCLabel & \bcell{41}{$.026$} & \bcell{50}{$.033$} & \bcell{36}{$.023$} & \bcell{10}{$.005$} & \bcell{5}{$\textbf{.002}$} \\
Pre-Diabetes & \Binary & \DeltaROCAUCLabel & \bcell{12}{$.014$} & \bcell{50}{$.069$} & \bcell{30}{$.040$} & \bcell{16}{$.019$} & \bcell{5}{$\textbf{.003}$} \\
Insulin Resistance & \Binary & \DeltaROCAUCLabel & \bcell{10}{$.007$} & \bcell{50}{$.040$} & \bcell{32}{$.025$} & \bcell{8}{$.004$} & \bcell{5}{$\textbf{.002}$} \\
HOMA-IR & \RegressionLabel & \DeltaPearsonLabel & \bcell{18}{$.023$} & \bcell{50}{$.070$} & \bcell{32}{$.043$} & \bcell{7}{$.006$} & \bcell{5}{$\textbf{.001}$} \\
HbA1c & \RegressionLabel & \DeltaPearsonLabel & \bcell{11}{$.008$} & \bcell{50}{$.065$} & \bcell{25}{$.029$} & \bcell{11}{$.009$} & \bcell{5}{$\textbf{-.001}$} \\
Triglycerides & \RegressionLabel & \DeltaPearsonLabel & \bcell{15}{$.042$} & \bcell{50}{$.122$} & \bcell{32}{$.081$} & \bcell{5}{$\textbf{.018}$} & \bcell{5}{$\textbf{.018}$} \\
\addlinespace
\grayrow \multicolumn{8}{l}{\textit{\Mental \textbf{Mental Health}}} \\
Mild Depression & \Binary & \DeltaROCAUCLabel & \bcell{36}{$.023$} & \bcell{50}{$.031$} & \bcell{46}{$.029$} & \bcell{20}{$.013$} & \bcell{5}{$\textbf{.004}$} \\
Mild Anxiety & \Binary & \DeltaROCAUCLabel & \bcell{42}{$.027$} & \bcell{50}{$.032$} & \bcell{47}{$.030$} & \bcell{16}{$.009$} & \bcell{5}{$\textbf{.001}$} \\
Persistent Stress & \Binary & \DeltaROCAUCLabel & \bcell{45}{$.033$} & \bcell{47}{$.034$} & \bcell{50}{$.036$} & \bcell{22}{$.017$} & \bcell{5}{$\textbf{.006}$} \\
Depress./Anxiety Dx & \Binary & \DeltaROCAUCLabel & \bcell{45}{$.021$} & \bcell{49}{$.023$} & \bcell{50}{$.023$} & \bcell{21}{$.011$} & \bcell{5}{$\textbf{.004}$} \\
Mental Health Med. & \Binary & \DeltaROCAUCLabel & \bcell{50}{$.010$} & \bcell{23}{$.008$} & \bcell{5}{$\textbf{.006}$} & \bcell{21}{$.008$} & \bcell{11}{$.007$} \\
PHQ-8 & \RegressionLabel & \DeltaPearsonLabel & \bcell{41}{$.056$} & \bcell{50}{$.068$} & \bcell{45}{$.062$} & \bcell{22}{$.031$} & \bcell{5}{$\textbf{.009}$} \\
GAD-7 & \RegressionLabel & \DeltaPearsonLabel & \bcell{40}{$.050$} & \bcell{50}{$.062$} & \bcell{46}{$.058$} & \bcell{20}{$.024$} & \bcell{5}{$\textbf{.004}$} \\
PSS & \RegressionLabel & \DeltaPearsonLabel & \bcell{50}{$.112$} & \bcell{34}{$.079$} & \bcell{34}{$.078$} & \bcell{16}{$.038$} & \bcell{5}{$\textbf{.014}$} \\
\addlinespace
\grayrow \multicolumn{8}{l}{\textit{\Sleep \textbf{Sleep}}} \\
Sleep Disorder Treatment & \Binary & \DeltaROCAUCLabel & \bcell{43}{$.015$} & \bcell{50}{$.017$} & \bcell{37}{$.014$} & \bcell{8}{$.005$} & \bcell{5}{$\textbf{.004}$} \\
Sleep Disturbance PRO & \RegressionLabel & \DeltaPearsonLabel & \bcell{50}{$.043$} & \bcell{26}{$.022$} & \bcell{23}{$.019$} & \bcell{12}{$.009$} & \bcell{5}{$\textbf{.003}$} \\
Sleep Impairment PRO & \RegressionLabel & \DeltaPearsonLabel & \bcell{39}{$.062$} & \bcell{46}{$.072$} & \bcell{50}{$.077$} & \bcell{28}{$.046$} & \bcell{5}{$\textbf{.012}$} \\
\addlinespace
\bottomrule
\end{tabular}
\label{tab:delta_predictive_tasks}
\end{table}

%% file: x-downstream_task_results.tex
\section{Downstream Task Results}
\label{sec:appendix_full_task_results}

\textbf{Discriminative Tasks.}
Table~\ref{tab:predictive_tasks_againstfeatures} compares a linear probe of \modelname-B (pretrained at 5M subjects) PCA-50 reduced embeddings against supervised baselines built on engineered features and/or demographic features across all $35$ discriminative tasks; \modelname{} achieves the best performance on $31$ of $35$ tasks.

\begin{figure*}[!htbp]
    \centering
    \includegraphics[width=0.95\textwidth]{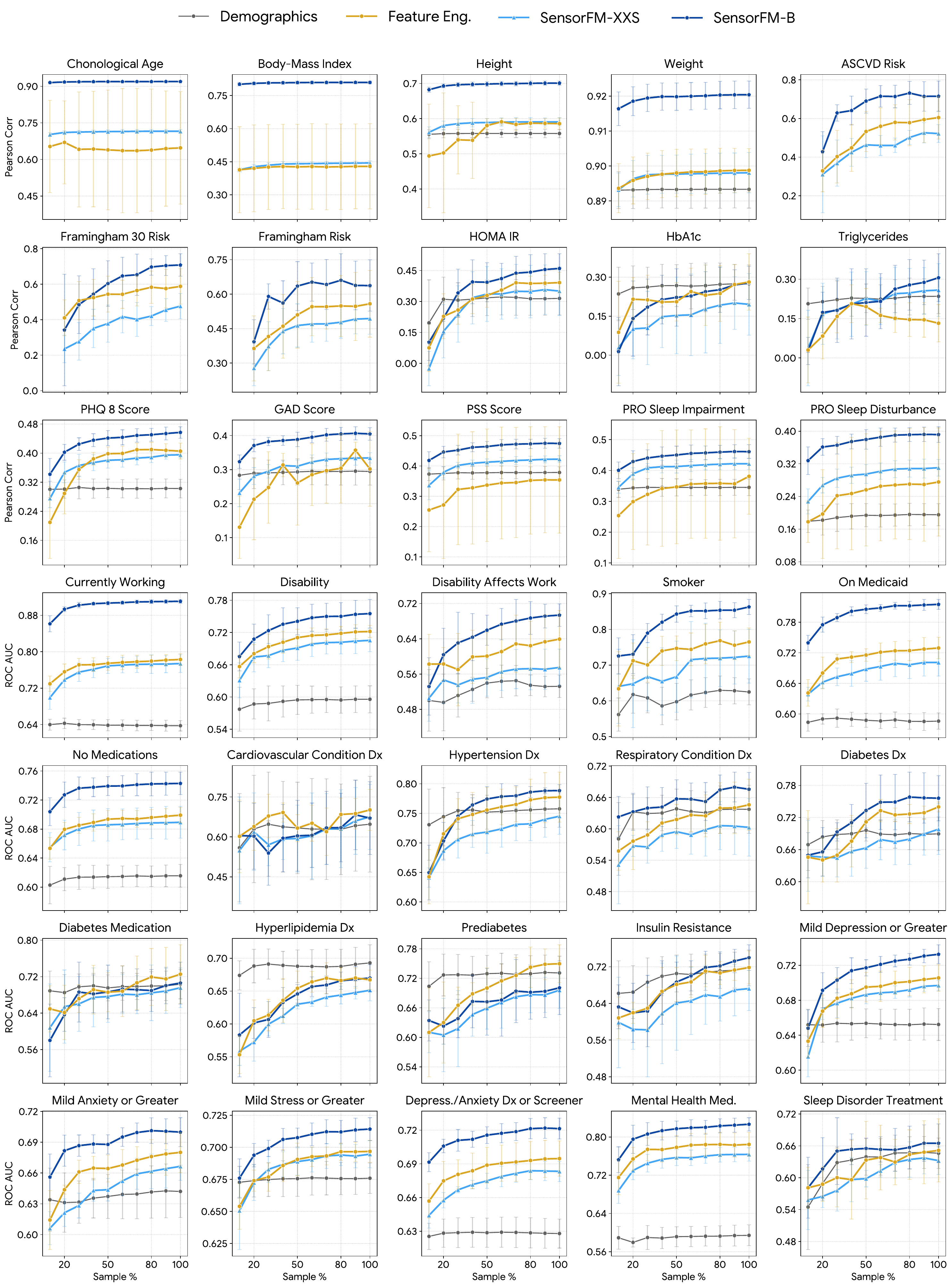}
    \caption{\textbf{Label Efficiency.} By varying the percentage of data in training set we interogate the label efficiency of the models. \modelname demonstrates good label efficient behavior.}
    \label{fig:few_shot} 
\end{figure*}

\textbf{Discriminative Few-Shot Performance.}
Figure~\ref{fig:few_shot} shows per-task few-shot performance curves obtained by training the post-adaptation head on progressively larger fractions of the downstream training set, comparing \modelname{} variants (XXS through B) against a supervised feature-engineered baseline and a demographics-only baseline across all $35$ downstream tasks. In the very-low-label regime, demographic priors act as a strong predictor for many tasks, but as labeled data increases \modelname{} surpasses both baselines, with the larger model variants (B) consistently outperforming smaller ones (XXS).

\textbf{Generative Tasks.}
Table~\ref{tab:generative_results} reports the generative task results, covering Random Imputation (80\%), Temporal Interpolation (20/60/180~min), Temporal Extrapolation (20/60/180~min), and Signal Imputation, with \modelname-B compared against naive baselines (Mean Fill, Nearest-Neighbor Fill, and Linear Interpolation). Table~\ref{tab:recovery_metrics} shows the downstream impact on daily-aggregated wearable metrics (steps, sleep stage minutes, active zone minutes, SpO2, wrist temperature) under a simulated 1-hour data loss, comparing the current consumer-wearable baseline (aggregation over observed values only) to \modelname-recovered aggregates and ground truth.

\input{Tables_Final/tbED9_pred_baselines}

\input{Tables_Final/tbED10_generative_baselines}

\input{Tables_Final/tbED11_generative_daily_metrics}

%% file: Tables_Final/tbED9_pred_baselines.tex

\begin{table}[!htbp]
\centering
\footnotesize

\caption{\textbf{Discriminative Task Performance Across a Sensor Foundation Model and Baselines.} The table presents the performance of \modelname-B, pretrained with the 5M data volume, compared against supervised baselines trained with engineered features and/or demographics on 35 discriminative tasks. \modelname (either with or without demographics) achieves the best performance on $31$ of $35$ tasks.  
\modelname variants are post-trained with PCA-50 reduced embeddings. 
For each task, we report the average 5-fold cross validation performance. Average Receiver Operating Characteristic Area Under the Curve (ROC AUC) is calculated in the logit-transform space and back-transformed. Average Pearson correlation ($r$) is calculated in the z-transform space and back-transformed. Standard deviations are calculated in the transformed space and back-transformed to give asymmetric error values. Colors are normalized per row; best model performance is bolded and has the deepest shade.}

\begin{tabular}{lccccccc}
\toprule
\textbf{Prediction Task} & \textbf{Type} & \textbf{Metric} &
\thead{\textbf{Demos.} \\ \textcolor{lightgray}{Feat. Eng.} \\ \textcolor{lightgray}{\modelname}} &
\thead{\textcolor{lightgray}{Demos.} \\ \textbf{Feat. Eng.} \\ \textcolor{lightgray}{\modelname}} &
\thead{\textbf{Demos.} \\ \textbf{Feat. Eng.} \\ \textcolor{lightgray}{\modelname}} &
\thead{ \textcolor{lightgray}{Demos.} \\ \textcolor{lightgray}{Feat. Eng.} \\ \textbf{\modelname}} &
\thead{\textbf{Demos.} \\ \textcolor{lightgray}{Feat. Eng.} \\ \textbf{\modelname}} \\
\midrule
\grayrow \multicolumn{8}{l}{\Demo \textit{\textbf{Demographics}}} \\
Age & \RegressionLabel & \PearsonLabel & - & \bcell{5}{$.662^{+\scriptscriptstyle .168}_{-\scriptscriptstyle .279}$} & - & \bcell{50}{$\textbf{.920}^{+\scriptscriptstyle .004}_{-\scriptscriptstyle .005}$} & - \\
BMI & \RegressionLabel & \PearsonLabel & - & \bcell{5}{$.441^{+\scriptscriptstyle .158}_{-\scriptscriptstyle .191}$} & - & \bcell{50}{$\textbf{.809}^{+\scriptscriptstyle .007}_{-\scriptscriptstyle .007}$} & - \\
Height & \RegressionLabel & \PearsonLabel & - & \bcell{5}{$.409^{+\scriptscriptstyle .174}_{-\scriptscriptstyle .210}$} & - & \bcell{50}{$\textbf{.675}^{+\scriptscriptstyle .012}_{-\scriptscriptstyle .012}$} & - \\
Weight & \RegressionLabel & \PearsonLabel & - & \bcell{5}{$.460^{+\scriptscriptstyle .011}_{-\scriptscriptstyle .012}$} & - & \bcell{50}{$\textbf{.809}^{+\scriptscriptstyle .007}_{-\scriptscriptstyle .007}$} & - \\
\addlinespace
\grayrow \multicolumn{8}{l}{\Life \textit{\textbf{Lifestyle}}} \\
Currently Working & \Binary & \ROCAUCLabel & \bcell{5}{$.637^{+\scriptscriptstyle .028}_{-\scriptscriptstyle .029}$} & \bcell{26}{$.769^{+\scriptscriptstyle .021}_{-\scriptscriptstyle .022}$} & \bcell{28}{$.782^{+\scriptscriptstyle .022}_{-\scriptscriptstyle .024}$} & \bcell{50}{$\textbf{.912}^{+\scriptscriptstyle .012}_{-\scriptscriptstyle .014}$} & \bcell{49}{$\textbf{.912}^{+\scriptscriptstyle .012}_{-\scriptscriptstyle .014}$} \\
Disability & \Binary & \ROCAUCLabel & \bcell{5}{$.600^{+\scriptscriptstyle .031}_{-\scriptscriptstyle .032}$} & \bcell{33}{$.702^{+\scriptscriptstyle .023}_{-\scriptscriptstyle .024}$} & \bcell{39}{$.722^{+\scriptscriptstyle .027}_{-\scriptscriptstyle .028}$} & \bcell{48}{$.753^{+\scriptscriptstyle .020}_{-\scriptscriptstyle .021}$} & \bcell{50}{$\textbf{.758}^{+\scriptscriptstyle .022}_{-\scriptscriptstyle .024}$} \\
Disability Affects Work & \Binary & \ROCAUCLabel & \bcell{5}{$.534^{+\scriptscriptstyle .044}_{-\scriptscriptstyle .044}$} & \bcell{24}{$.605^{+\scriptscriptstyle .050}_{-\scriptscriptstyle .052}$} & \bcell{29}{$.625^{+\scriptscriptstyle .044}_{-\scriptscriptstyle .047}$} & \bcell{50}{$\textbf{.699}^{+\scriptscriptstyle .040}_{-\scriptscriptstyle .043}$} & \bcell{49}{$.697^{+\scriptscriptstyle .041}_{-\scriptscriptstyle .044}$} \\
Smoking & \Binary & \ROCAUCLabel & \bcell{5}{$.629^{+\scriptscriptstyle .049}_{-\scriptscriptstyle .051}$} & \bcell{28}{$.754^{+\scriptscriptstyle .046}_{-\scriptscriptstyle .052}$} & \bcell{31}{$.771^{+\scriptscriptstyle .037}_{-\scriptscriptstyle .042}$} & \bcell{50}{$\textbf{.870}^{+\scriptscriptstyle .060}_{-\scriptscriptstyle .098}$} & \bcell{49}{$.867^{+\scriptscriptstyle .060}_{-\scriptscriptstyle .097}$} \\
Medicaid & \Binary & \ROCAUCLabel & \bcell{5}{$.589^{+\scriptscriptstyle .020}_{-\scriptscriptstyle .020}$} & \bcell{32}{$.726^{+\scriptscriptstyle .025}_{-\scriptscriptstyle .027}$} & \bcell{33}{$.729^{+\scriptscriptstyle .025}_{-\scriptscriptstyle .027}$} & \bcell{50}{$\textbf{.814}^{+\scriptscriptstyle .021}_{-\scriptscriptstyle .023}$} & \bcell{49}{$\textbf{.814}^{+\scriptscriptstyle .022}_{-\scriptscriptstyle .024}$} \\
No Medications & \Binary & \ROCAUCLabel & \bcell{5}{$.617^{+\scriptscriptstyle .018}_{-\scriptscriptstyle .018}$} & \bcell{29}{$.687^{+\scriptscriptstyle .006}_{-\scriptscriptstyle .006}$} & \bcell{33}{$.699^{+\scriptscriptstyle .008}_{-\scriptscriptstyle .008}$} & \bcell{48}{$.739^{+\scriptscriptstyle .009}_{-\scriptscriptstyle .009}$} & \bcell{50}{$\textbf{.744}^{+\scriptscriptstyle .009}_{-\scriptscriptstyle .009}$} \\
\addlinespace
\grayrow \multicolumn{8}{l}{\Cardio \textit{\textbf{Cardiovascular}}} \\
Cardiovascular Dx & \Binary & \ROCAUCLabel & \bcell{18}{$.701^{+\scriptscriptstyle .037}_{-\scriptscriptstyle .039}$} & \bcell{5}{$.696^{+\scriptscriptstyle .090}_{-\scriptscriptstyle .107}$} & \bcell{31}{$.706^{+\scriptscriptstyle .091}_{-\scriptscriptstyle .111}$} & \bcell{50}{$\textbf{.712}^{+\scriptscriptstyle .052}_{-\scriptscriptstyle .058}$} & \bcell{5}{$.696^{+\scriptscriptstyle .057}_{-\scriptscriptstyle .064}$} \\
Hypertension Dx & \Binary & \ROCAUCLabel & \bcell{22}{$.762^{+\scriptscriptstyle .038}_{-\scriptscriptstyle .043}$} & \bcell{5}{$.747^{+\scriptscriptstyle .019}_{-\scriptscriptstyle .020}$} & \bcell{42}{$.780^{+\scriptscriptstyle .021}_{-\scriptscriptstyle .022}$} & \bcell{50}{$\textbf{.786}^{+\scriptscriptstyle .023}_{-\scriptscriptstyle .025}$} & \bcell{49}{$\textbf{.786}^{+\scriptscriptstyle .022}_{-\scriptscriptstyle .023}$} \\
Respiratory Dx & \Binary & \ROCAUCLabel & \bcell{5}{$.640^{+\scriptscriptstyle .049}_{-\scriptscriptstyle .052}$} & \bcell{7}{$.642^{+\scriptscriptstyle .039}_{-\scriptscriptstyle .041}$} & \bcell{5}{$.640^{+\scriptscriptstyle .031}_{-\scriptscriptstyle .033}$} & \bcell{50}{$\textbf{.682}^{+\scriptscriptstyle .054}_{-\scriptscriptstyle .059}$} & \bcell{42}{$.674^{+\scriptscriptstyle .046}_{-\scriptscriptstyle .050}$} \\
ASCVD Risk & \RegressionLabel & \PearsonLabel & \bcell{38}{$.740^{+\scriptscriptstyle .054}_{-\scriptscriptstyle .066}$} & \bcell{5}{$.604^{+\scriptscriptstyle .069}_{-\scriptscriptstyle .079}$} & \bcell{44}{$.764^{+\scriptscriptstyle .058}_{-\scriptscriptstyle .074}$} & \bcell{36}{$.730^{+\scriptscriptstyle .091}_{-\scriptscriptstyle .127}$} & \bcell{50}{$\textbf{.784}^{+\scriptscriptstyle .059}_{-\scriptscriptstyle .078}$} \\
Framingham Risk & \RegressionLabel & \PearsonLabel & \bcell{50}{$\textbf{.743}^{+\scriptscriptstyle .041}_{-\scriptscriptstyle .047}$} & \bcell{5}{$.548^{+\scriptscriptstyle .072}_{-\scriptscriptstyle .081}$} & \bcell{46}{$.730^{+\scriptscriptstyle .063}_{-\scriptscriptstyle .078}$} & \bcell{32}{$.669^{+\scriptscriptstyle .101}_{-\scriptscriptstyle .133}$} & \bcell{45}{$.724^{+\scriptscriptstyle .057}_{-\scriptscriptstyle .069}$} \\
Framingham 30 Risk & \RegressionLabel & \PearsonLabel & \bcell{50}{$\textbf{.782}^{+\scriptscriptstyle .038}_{-\scriptscriptstyle .045}$} & \bcell{5}{$.592^{+\scriptscriptstyle .045}_{-\scriptscriptstyle .049}$} & \bcell{47}{$.772^{+\scriptscriptstyle .043}_{-\scriptscriptstyle .052}$} & \bcell{33}{$.714^{+\scriptscriptstyle .052}_{-\scriptscriptstyle .062}$} & \bcell{43}{$.756^{+\scriptscriptstyle .039}_{-\scriptscriptstyle .046}$} \\
\addlinespace
\grayrow \multicolumn{8}{l}{\Metabolic \textit{\textbf{Metabolic}}} \\
Diabetes Dx & \Binary & \ROCAUCLabel & \bcell{5}{$.688^{+\scriptscriptstyle .052}_{-\scriptscriptstyle .058}$} & \bcell{28}{$.727^{+\scriptscriptstyle .059}_{-\scriptscriptstyle .068}$} & \bcell{32}{$.734^{+\scriptscriptstyle .062}_{-\scriptscriptstyle .073}$} & \bcell{50}{$\textbf{.763}^{+\scriptscriptstyle .037}_{-\scriptscriptstyle .042}$} & \bcell{45}{$.756^{+\scriptscriptstyle .043}_{-\scriptscriptstyle .049}$} \\
Diabetes Med. & \Binary & \ROCAUCLabel & \bcell{47}{$.704^{+\scriptscriptstyle .042}_{-\scriptscriptstyle .046}$} & \bcell{5}{$.672^{+\scriptscriptstyle .064}_{-\scriptscriptstyle .071}$} & \bcell{50}{$\textbf{.706}^{+\scriptscriptstyle .069}_{-\scriptscriptstyle .080}$} & \bcell{41}{$.700^{+\scriptscriptstyle .045}_{-\scriptscriptstyle .049}$} & \bcell{48}{$.705^{+\scriptscriptstyle .040}_{-\scriptscriptstyle .043}$} \\
Hyperlipidemia & \Binary & \ROCAUCLabel & \bcell{50}{$\textbf{.688}^{+\scriptscriptstyle .014}_{-\scriptscriptstyle .015}$} & \bcell{5}{$.643^{+\scriptscriptstyle .050}_{-\scriptscriptstyle .053}$} & \bcell{31}{$.670^{+\scriptscriptstyle .041}_{-\scriptscriptstyle .044}$} & \bcell{35}{$.674^{+\scriptscriptstyle .032}_{-\scriptscriptstyle .033}$} & \bcell{37}{$.676^{+\scriptscriptstyle .027}_{-\scriptscriptstyle .029}$} \\
Pre-Diabetes & \Binary & \ROCAUCLabel & \bcell{36}{$.731^{+\scriptscriptstyle .031}_{-\scriptscriptstyle .033}$} & \bcell{32}{$.728^{+\scriptscriptstyle .052}_{-\scriptscriptstyle .059}$} & \bcell{50}{$\textbf{.742}^{+\scriptscriptstyle .047}_{-\scriptscriptstyle .054}$} & \bcell{5}{$.704^{+\scriptscriptstyle .068}_{-\scriptscriptstyle .078}$} & \bcell{8}{$.707^{+\scriptscriptstyle .066}_{-\scriptscriptstyle .075}$} \\
Insulin Resistance & \Binary & \ROCAUCLabel & \bcell{10}{$.717^{+\scriptscriptstyle .044}_{-\scriptscriptstyle .048}$} & \bcell{5}{$.710^{+\scriptscriptstyle .050}_{-\scriptscriptstyle .055}$} & \bcell{10}{$.717^{+\scriptscriptstyle .040}_{-\scriptscriptstyle .044}$} & \bcell{48}{$.761^{+\scriptscriptstyle .044}_{-\scriptscriptstyle .050}$} & \bcell{50}{$\textbf{.763}^{+\scriptscriptstyle .044}_{-\scriptscriptstyle .051}$} \\
HOMA-IR & \RegressionLabel & \PearsonLabel & \bcell{5}{$.316^{+\scriptscriptstyle .068}_{-\scriptscriptstyle .071}$} & \bcell{20}{$.374^{+\scriptscriptstyle .069}_{-\scriptscriptstyle .073}$} & \bcell{27}{$.397^{+\scriptscriptstyle .042}_{-\scriptscriptstyle .044}$} & \bcell{49}{$.479^{+\scriptscriptstyle .030}_{-\scriptscriptstyle .031}$} & \bcell{50}{$\textbf{.480}^{+\scriptscriptstyle .029}_{-\scriptscriptstyle .030}$} \\
HbA1c & \RegressionLabel & \PearsonLabel & \bcell{43}{$.282^{+\scriptscriptstyle .067}_{-\scriptscriptstyle .069}$} & \bcell{5}{$.221^{+\scriptscriptstyle .106}_{-\scriptscriptstyle .112}$} & \bcell{9}{$.228^{+\scriptscriptstyle .090}_{-\scriptscriptstyle .094}$} & \bcell{50}{$\textbf{.293}^{+\scriptscriptstyle .029}_{-\scriptscriptstyle .030}$} & \bcell{49}{$.292^{+\scriptscriptstyle .026}_{-\scriptscriptstyle .026}$} \\
Triglycerides & \RegressionLabel & \PearsonLabel & \bcell{39}{$.251^{+\scriptscriptstyle .042}_{-\scriptscriptstyle .043}$} & \bcell{5}{$.131^{+\scriptscriptstyle .065}_{-\scriptscriptstyle .066}$} & \bcell{17}{$.173^{+\scriptscriptstyle .043}_{-\scriptscriptstyle .044}$} & \bcell{44}{$.269^{+\scriptscriptstyle .047}_{-\scriptscriptstyle .048}$} & \bcell{50}{$\textbf{.287}^{+\scriptscriptstyle .054}_{-\scriptscriptstyle .056}$} \\
\addlinespace
\grayrow \multicolumn{8}{l}{\Mental \textit{\textbf{Mental Health}}} \\
Mild Depression & \Binary & \ROCAUCLabel & \bcell{5}{$.653^{+\scriptscriptstyle .009}_{-\scriptscriptstyle .009}$} & \bcell{21}{$.682^{+\scriptscriptstyle .014}_{-\scriptscriptstyle .015}$} & \bcell{35}{$.705^{+\scriptscriptstyle .006}_{-\scriptscriptstyle .007}$} & \bcell{47}{$.726^{+\scriptscriptstyle .006}_{-\scriptscriptstyle .006}$} & \bcell{50}{$\textbf{.730}^{+\scriptscriptstyle .011}_{-\scriptscriptstyle .011}$} \\
Mild Anxiety & \Binary & \ROCAUCLabel & \bcell{5}{$.641^{+\scriptscriptstyle .009}_{-\scriptscriptstyle .009}$} & \bcell{14}{$.654^{+\scriptscriptstyle .010}_{-\scriptscriptstyle .011}$} & \bcell{35}{$.680^{+\scriptscriptstyle .005}_{-\scriptscriptstyle .006}$} & \bcell{49}{$.698^{+\scriptscriptstyle .005}_{-\scriptscriptstyle .005}$} & \bcell{50}{$\textbf{.699}^{+\scriptscriptstyle .006}_{-\scriptscriptstyle .006}$} \\
Persistent Stress & \Binary & \ROCAUCLabel & \bcell{14}{$.676^{+\scriptscriptstyle .022}_{-\scriptscriptstyle .023}$} & \bcell{5}{$.664^{+\scriptscriptstyle .033}_{-\scriptscriptstyle .034}$} & \bcell{32}{$.697^{+\scriptscriptstyle .025}_{-\scriptscriptstyle .026}$} & \bcell{45}{$.712^{+\scriptscriptstyle .026}_{-\scriptscriptstyle .028}$} & \bcell{50}{$\textbf{.717}^{+\scriptscriptstyle .026}_{-\scriptscriptstyle .027}$} \\
Depress./Anxiety Dx & \Binary & \ROCAUCLabel & \bcell{5}{$.626^{+\scriptscriptstyle .012}_{-\scriptscriptstyle .012}$} & \bcell{26}{$.672^{+\scriptscriptstyle .022}_{-\scriptscriptstyle .022}$} & \bcell{36}{$.693^{+\scriptscriptstyle .015}_{-\scriptscriptstyle .015}$} & \bcell{47}{$.717^{+\scriptscriptstyle .009}_{-\scriptscriptstyle .009}$} & \bcell{50}{$\textbf{.721}^{+\scriptscriptstyle .008}_{-\scriptscriptstyle .008}$} \\
Mental Health Med. & \Binary & \ROCAUCLabel & \bcell{5}{$.594^{+\scriptscriptstyle .015}_{-\scriptscriptstyle .015}$} & \bcell{39}{$.773^{+\scriptscriptstyle .006}_{-\scriptscriptstyle .006}$} & \bcell{41}{$.783^{+\scriptscriptstyle .007}_{-\scriptscriptstyle .007}$} & \bcell{48}{$.819^{+\scriptscriptstyle .020}_{-\scriptscriptstyle .022}$} & \bcell{50}{$\textbf{.826}^{+\scriptscriptstyle .017}_{-\scriptscriptstyle .018}$} \\
PHQ-8 & \RegressionLabel & \PearsonLabel & \bcell{5}{$.303^{+\scriptscriptstyle .018}_{-\scriptscriptstyle .018}$} & \bcell{19}{$.354^{+\scriptscriptstyle .018}_{-\scriptscriptstyle .018}$} & \bcell{35}{$.410^{+\scriptscriptstyle .009}_{-\scriptscriptstyle .009}$} & \bcell{47}{$.450^{+\scriptscriptstyle .019}_{-\scriptscriptstyle .019}$} & \bcell{50}{$\textbf{.459}^{+\scriptscriptstyle .023}_{-\scriptscriptstyle .024}$} \\
GAD-7 & \RegressionLabel & \PearsonLabel & \bcell{18}{$.291^{+\scriptscriptstyle .009}_{-\scriptscriptstyle .009}$} & \bcell{5}{$.244^{+\scriptscriptstyle .130}_{-\scriptscriptstyle .140}$} & \bcell{19}{$.294^{+\scriptscriptstyle .139}_{-\scriptscriptstyle .152}$} & \bcell{48}{$.400^{+\scriptscriptstyle .025}_{-\scriptscriptstyle .025}$} & \bcell{50}{$\textbf{.403}^{+\scriptscriptstyle .025}_{-\scriptscriptstyle .026}$} \\
PSS & \RegressionLabel & \PearsonLabel & \bcell{24}{$.378^{+\scriptscriptstyle .019}_{-\scriptscriptstyle .019}$} & \bcell{5}{$.305^{+\scriptscriptstyle .124}_{-\scriptscriptstyle .135}$} & \bcell{34}{$.417^{+\scriptscriptstyle .040}_{-\scriptscriptstyle .041}$} & \bcell{46}{$.463^{+\scriptscriptstyle .013}_{-\scriptscriptstyle .014}$} & \bcell{50}{$\textbf{.476}^{+\scriptscriptstyle .018}_{-\scriptscriptstyle .019}$} \\
\addlinespace
\grayrow \multicolumn{8}{l}{\Sleep \textit{\textbf{Sleep}}} \\
Sleep Disorder Treatment & \Binary & \ROCAUCLabel & \bcell{50}{$\textbf{.653}^{+\scriptscriptstyle .037}_{-\scriptscriptstyle .039}$} & \bcell{5}{$.610^{+\scriptscriptstyle .024}_{-\scriptscriptstyle .024}$} & \bcell{20}{$.625^{+\scriptscriptstyle .047}_{-\scriptscriptstyle .049}$} & \bcell{45}{$.649^{+\scriptscriptstyle .036}_{-\scriptscriptstyle .038}$} & \bcell{49}{$\textbf{.653}^{+\scriptscriptstyle .040}_{-\scriptscriptstyle .042}$} \\
Sleep Disturbance PRO & \RegressionLabel & \PearsonLabel & \bcell{5}{$.195^{+\scriptscriptstyle .028}_{-\scriptscriptstyle .028}$} & \bcell{25}{$.283^{+\scriptscriptstyle .105}_{-\scriptscriptstyle .112}$} & \bcell{34}{$.326^{+\scriptscriptstyle .056}_{-\scriptscriptstyle .059}$} & \bcell{49}{$.390^{+\scriptscriptstyle .016}_{-\scriptscriptstyle .016}$} & \bcell{50}{$\textbf{.393}^{+\scriptscriptstyle .014}_{-\scriptscriptstyle .014}$} \\
Sleep Impairment PRO & \RegressionLabel & \PearsonLabel & \bcell{15}{$.344^{+\scriptscriptstyle .023}_{-\scriptscriptstyle .023}$} & \bcell{5}{$.310^{+\scriptscriptstyle .138}_{-\scriptscriptstyle .152}$} & \bcell{23}{$.372^{+\scriptscriptstyle .143}_{-\scriptscriptstyle .163}$} & \bcell{46}{$.448^{+\scriptscriptstyle .013}_{-\scriptscriptstyle .013}$} & \bcell{50}{$\textbf{.460}^{+\scriptscriptstyle .016}_{-\scriptscriptstyle .016}$} \\
\addlinespace
\bottomrule
\end{tabular}
\label{tab:predictive_tasks_againstfeatures}
\end{table}

%% file: Tables_Final/tbED10_generative_baselines.tex

\begin{table}[!htbp]
\definecolor{niceblue}{HTML}{1976D2}
\centering
\footnotesize
\caption{\textbf{Generative Performance Across Data Imputation, Interpolation, and Extrapolation Tasks.} For Random Imputation, Temporal Interpolation / Extrapolation, and Signal Imputation we report Mean Squared Error (MSE) on known non-missing values. Presented error are $95\%$ confidence intervals generated through 100 bootstrap iterations.}
\label{tab:generative_results}
\begin{tabular}{lcccc}
\toprule
\multirow{2}{*}{\textbf{Generative Task}} & \multicolumn{4}{c}{\textbf{Method}} \\ 
\cmidrule(lr){2-5}
& Mean Fill & NN Fill & Linear Interp. & \modelname-B \\
\midrule

\grayrow \textbf{Random Imp.} &&&& \\
80\% & \cellcolor{niceblue!35}0.915 $\pm$ 0.002 & \cellcolor{niceblue!26}1.020 $\pm$ 0.001 & \cellcolor{niceblue!40}0.854 $\pm$ 0.002 & \cellcolor{niceblue!92}0.215 $\pm$ 0.001 \\

\addlinespace
\grayrow \textbf{Temporal Interp.} &&&& \\
20 min & \cellcolor{niceblue!38}0.876 $\pm$ 0.008 & \cellcolor{niceblue!53}0.693 $\pm$ 0.008 & \cellcolor{niceblue!64}0.561 $\pm$ 0.006 & \cellcolor{niceblue!81}0.353 $\pm$ 0.002 \\
60 min & \cellcolor{niceblue!36}0.904 $\pm$ 0.005 & \cellcolor{niceblue!32}0.943 $\pm$ 0.007 & \cellcolor{niceblue!46}0.777 $\pm$ 0.007 & \cellcolor{niceblue!71}0.468 $\pm$ 0.003 \\
180 min & \cellcolor{niceblue!32}0.950 $\pm$ 0.007 & \cellcolor{niceblue!14}1.163 $\pm$ 0.008 & \cellcolor{niceblue!31}0.961 $\pm$ 0.008 & \cellcolor{niceblue!63}0.574 $\pm$ 0.002 \\

\addlinespace
\grayrow \textbf{Temporal Extrap.} &&&& \\
20 min & \cellcolor{niceblue!34}0.923 $\pm$ 0.006 & \cellcolor{niceblue!40}0.846 $\pm$ 0.010 & \cellcolor{niceblue!40}0.846 $\pm$ 0.014 & \cellcolor{niceblue!72}0.463 $\pm$ 0.004 \\
60 min & \cellcolor{niceblue!33}0.937 $\pm$ 0.007 & \cellcolor{niceblue!19}1.102 $\pm$ 0.014 & \cellcolor{niceblue!19}1.102 $\pm$ 0.008 & \cellcolor{niceblue!64}0.563 $\pm$ 0.004 \\
180 min & \cellcolor{niceblue!30}0.974 $\pm$ 0.006 & \cellcolor{niceblue!0}1.336 $\pm$ 0.010 & \cellcolor{niceblue!0}1.336 $\pm$ 0.011 & \cellcolor{niceblue!57}0.646 $\pm$ 0.003 \\

\addlinespace
\grayrow \textbf{Signal Imp.} &&&& \\
2/26 & \cellcolor{niceblue!26}1.016 $\pm$ 0.006 & \cellcolor{niceblue!26}1.016 $\pm$ 0.017 & \cellcolor{niceblue!26}1.016 $\pm$ 0.012 & \cellcolor{niceblue!100}0.122 $\pm$ 0.003 \\
6/26 & \cellcolor{niceblue!26}1.020 $\pm$ 0.005 & \cellcolor{niceblue!26}1.020 $\pm$ 0.006 & \cellcolor{niceblue!26}1.020 $\pm$ 0.008 & \cellcolor{niceblue!99}0.137 $\pm$ 0.002 \\
12/26 & \cellcolor{niceblue!26}1.025 $\pm$ 0.005 & \cellcolor{niceblue!26}1.025 $\pm$ 0.003 & \cellcolor{niceblue!26}1.025 $\pm$ 0.003 & \cellcolor{niceblue!96}0.170 $\pm$ 0.001 \\
20/26 & \cellcolor{niceblue!26}1.022 $\pm$ 0.002 & \cellcolor{niceblue!26}1.022 $\pm$ 0.003 & \cellcolor{niceblue!26}1.022 $\pm$ 0.003 & \cellcolor{niceblue!91}0.236 $\pm$ 0.001 \\

\bottomrule
\end{tabular}
\end{table}

%% file: Tables_Final/tbED11_generative_daily_metrics.tex

\begin{table}[!htbp]
\centering
\footnotesize
\caption{\textbf{Reconstructed Daily Sum-Aggregated Metrics.} Comparison of Baseline, \modelname Recovered, and Ground Truth under a simulated 1-hour data loss across all modalities. Baseline represents current standard for consumer wearable systems where no recovery methods are used and the aggregate is calculated solely over observed values. Note that SpO2 and Wrist Temp minutes do not sum to a full day due to inherent missingness in the original ground truth data. For a controlled comparison, aggregations are only computed over known valid minutes, which include the simulated loss. Presented error are $95\%$ confidence intervals generated through 100 bootstrap iterations.}
\label{tab:recovery_metrics}
\begin{tabular}{lccc}
\toprule
\multirow{2}{*}{\textbf{Metric}} & \multicolumn{3}{c}{\textbf{Daily Average Value}} \\ 
\cmidrule(lr){2-4}
& Baseline & \modelname Recovered & Ground Truth \\
\midrule

\grayrow \textbf{Activity (a.u.)} &&& \\
Steps & \cellcolor{RoyalBlue!0}5958.89 & \cellcolor{RoyalBlue!93}6208.41 $\pm$ 35.14 & \cellcolor{RoyalBlue!100}6227.49 \\

\addlinespace
\grayrow \textbf{Sleep Stages (Minutes)} &&& \\
Light & \cellcolor{RoyalBlue!3}24.57 & \cellcolor{RoyalBlue!81}25.43 $\pm$ 0.12 & \cellcolor{RoyalBlue!100}25.64 \\
Deep & \cellcolor{RoyalBlue!5}426.24 & \cellcolor{RoyalBlue!98}444.12 $\pm$ 1.60 & \cellcolor{RoyalBlue!100}444.48 \\
REM & \cellcolor{RoyalBlue!7}66.71 & \cellcolor{RoyalBlue!91}69.78 $\pm$ 0.33 & \cellcolor{RoyalBlue!100}69.51 \\

\addlinespace
\grayrow \textbf{Exercise (Minutes)} &&& \\
Light (95$\le$HR$<$114) & \cellcolor{RoyalBlue!1}125.07 & \cellcolor{RoyalBlue!82}129.63 $\pm$ 1.04 & \cellcolor{RoyalBlue!100}130.64 \\
Aerobic (114$\le$HR$<$152) & \cellcolor{RoyalBlue!1}21.94 & \cellcolor{RoyalBlue!45}22.38 $\pm$ 0.32 & \cellcolor{RoyalBlue!100}22.92 \\
Anaerobic (152$\le$HR) & \cellcolor{RoyalBlue!35}1.04 & \cellcolor{RoyalBlue!57}1.05 $\pm$ 0.08 & \cellcolor{RoyalBlue!100}1.07 \\

\addlinespace
\grayrow \textbf{SPO2 (Minutes)} &&& \\
High ($>$90\%) & \cellcolor{RoyalBlue!5}223.68 & \cellcolor{RoyalBlue!99}233.40 $\pm$ 1.20 & \cellcolor{RoyalBlue!100}233.28 \\
Low ($<$90\%) & \cellcolor{RoyalBlue!11}6.48 & \cellcolor{RoyalBlue!59}6.62 $\pm$ 0.15 & \cellcolor{RoyalBlue!100}6.74 \\

\addlinespace
\grayrow \textbf{Wrist Temp (Minutes)} &&& \\
Normal ($<$37$^\circ$C) & \cellcolor{RoyalBlue!3}1065.56 & \cellcolor{RoyalBlue!100}1112.05 $\pm$ 4.78 & \cellcolor{RoyalBlue!100}1112.03 \\
High ($\ge$37$^\circ$C) & \cellcolor{RoyalBlue!38}1.10 & \cellcolor{RoyalBlue!79}1.12 $\pm$ 0.16 & \cellcolor{RoyalBlue!100}1.13 \\

\bottomrule
\end{tabular}
\end{table}

%% file: x-reconstruction_examples.tex
\newpage
\section{Reconstruction Visualization}
\label{sec:appendix_recon}
\begin{figure*}[!htbp]
    \centering
    \includegraphics[width=.9\textwidth]{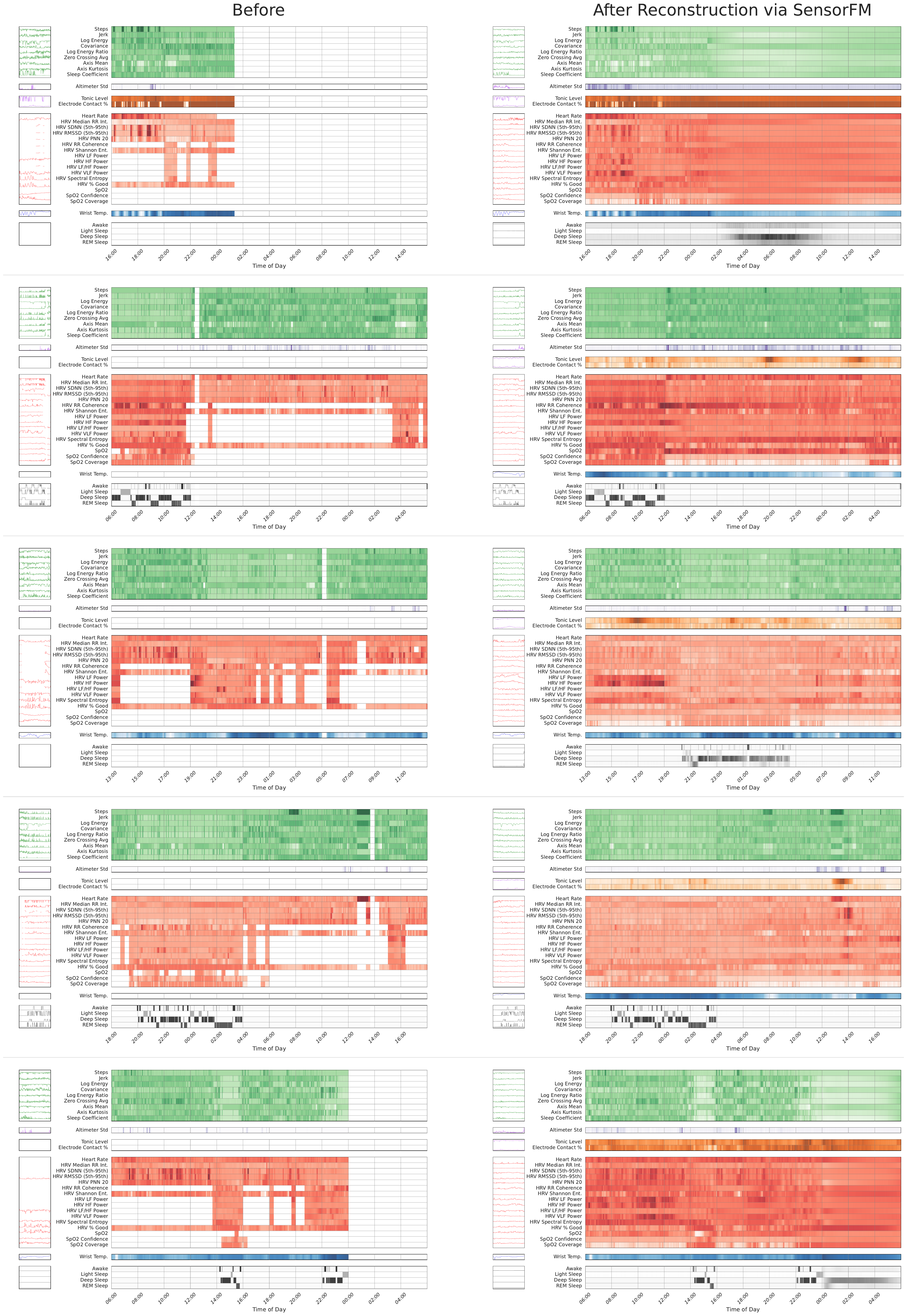}
    \caption{\textbf{Generative Reconstruction of Multimodal Sensor Data.} \modelname model reconstructions of fragmented multimodal wearable sensor data. Each row represents one 24-hour sample from the pretraining \emph{validation} dataset. 
    The plot highlights how the model, through its generative pre-text task, internalizes structures in the data and enables the filling of missing segments with plasible, non-linear reconstructions.}
    \label{fig:reconstruction_heatmaps} 
\end{figure*}

\begin{figure*}[!htbp]
    \centering
    \includegraphics[width=\textwidth]{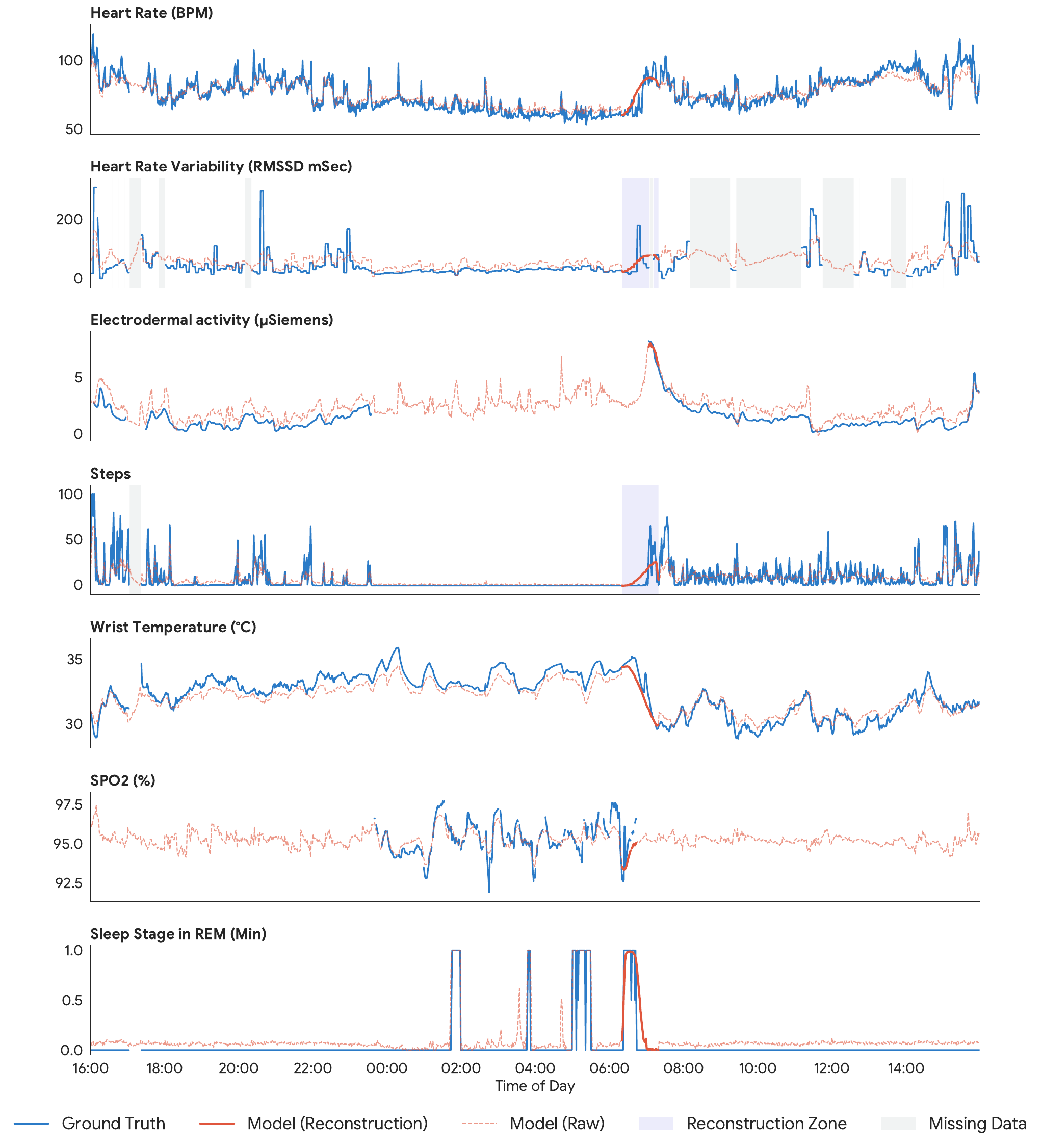}
    \caption{\textbf{Generative Reconstruction of Fragmented Sensor Data (Example I).} This figure illustrates the model's capability to impute missing data (gap-filling) in the presence of high-frequency, fragmented signal loss, a common artifact in wearable data collection. The panels display seven distinct physiological and behavioral signals (top to bottom: Heart Rate, Heart Rate Variability, Electrodermal Activity, Steps, Wrist Temperature, SpO2, and Sleep Stage REM) over a period of approximately 22 hours. Grey shaded regions (Reconstruction Zone) indicate intervals where the input data were intentionally masked to test the model's reconstruction performance. The red line (Model Reconstruction) tracks the blue line (Ground Truth) with high fidelity within these masked windows, demonstrating the model's ability to infer instantaneous physiological states from surrounding context. The dashed light red line represents the raw model output.}
    \label{fig:reconstruction_line_graphs_1} 
\end{figure*}

\begin{figure*}[!htbp]
    \centering
    \includegraphics[width=\textwidth]{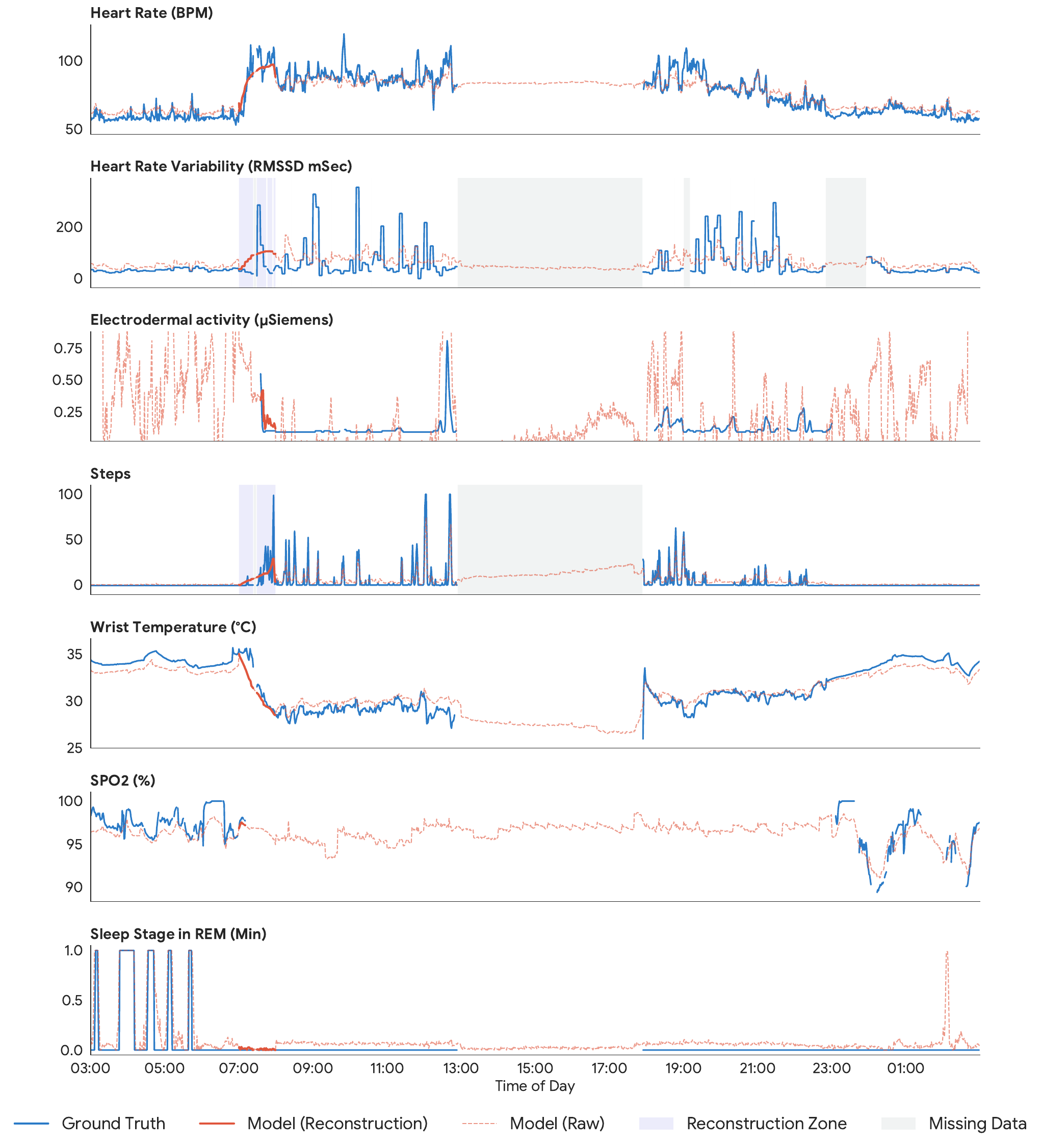}
    \caption{\textbf{Generative Reconstruction of Fragmented Sensor Data (Example II).} This example features a multi-hour block mask (approx. 10 hours) indicated by the central grey shaded region. Despite the lack of immediate local context, the model (red line) successfully reconstructs physiologically plausible baselines and circadian rhythms that align with the ground truth (blue line) for metrics such as Heart Rate and Wrist Temperature. This highlights the model's ability to leverage long-context internal representations to perform temporal interpolation and maintain continuity in health state estimation during prolonged data gaps.}
    \label{fig:reconstruction_line_graphs_2} 
\end{figure*}

Figure~\ref{fig:reconstruction_heatmaps} presents 24-hour multimodal reconstruction heatmaps across multiple held-out validation subjects, illustrating how \modelname{} fills in fragmented multimodal sensor segments through its generative pre-text task. Figures~\ref{fig:reconstruction_line_graphs_1} and~\ref{fig:reconstruction_line_graphs_2} zoom into two representative example days at per-signal resolution (Heart Rate, Heart Rate Variability, Electrodermal Activity, Steps, Wrist Temperature, SpO2, and Sleep Stage REM) under two qualitatively different masking regimes: high-frequency fragmented signal loss (Example~I) and a single multi-hour ($\sim$10~h) block mask (Example~II). The two examples show that the model leverages both local context (Example~I) and long-context internal representations to maintain physiologically plausible baselines and circadian structure (Example~II).

%% file: x-latent_space_analysis_and_viz.tex
\section{Analysis of the Model Embeddings in Latent Space}
\label{sec:appendix_feature_import}

\begin{figure*}[!htbp]
    \centering
    \includegraphics[width=1\textwidth]{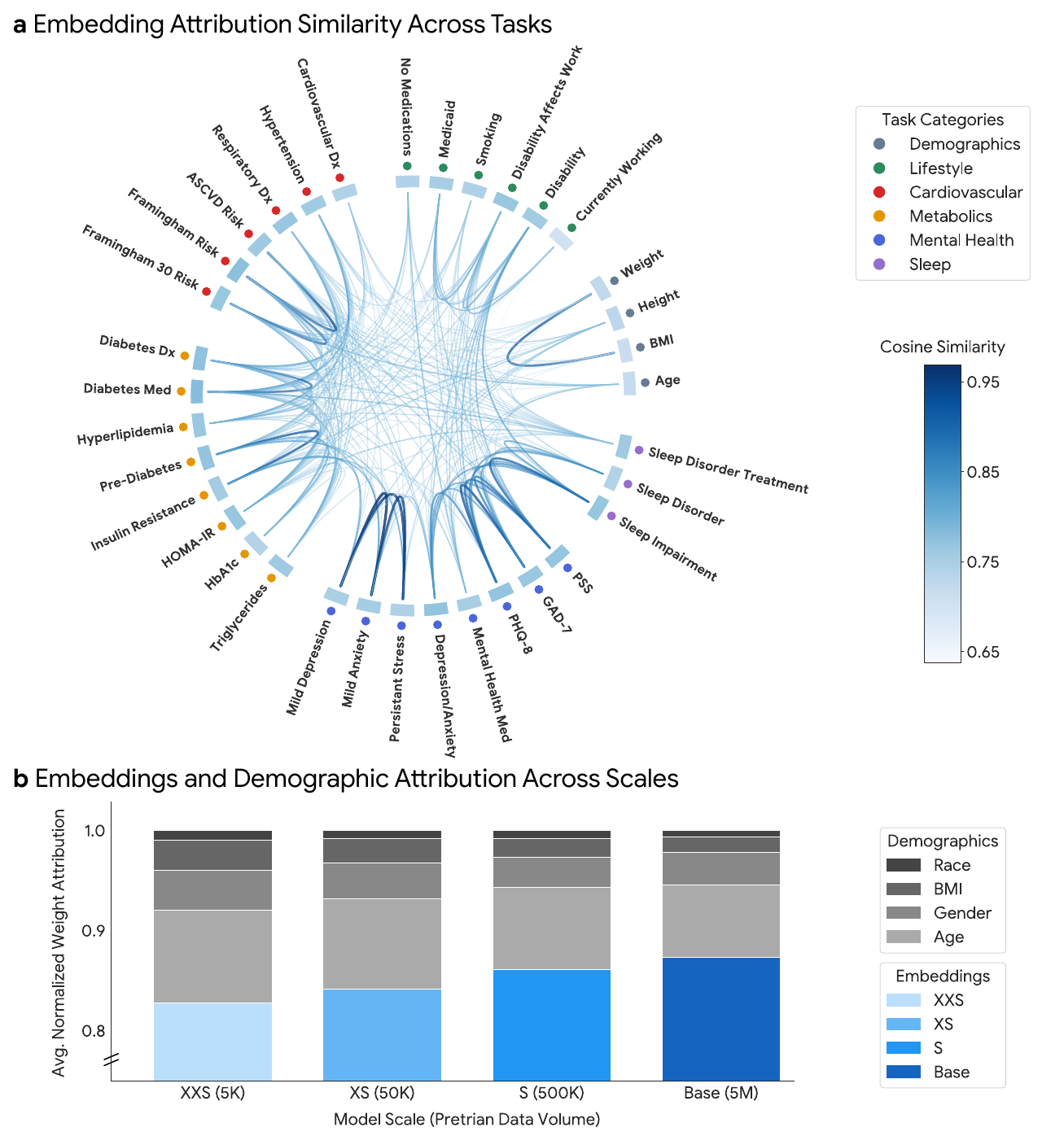}
    \caption{\textbf{Latent Embedding Attributions.} \textbf{(a)} The chord diagram illustrates the $cosine$ similarity in latent feature attribution (SHAP) between pairs of downstream tasks (post-trained \emph{without} demographics). The outer ring colored is the average pairwise similarity of a task. Central chords represent similarity between the normalized, exact SHAP attributions (weight collapse analysis) of two tasks, indicating the degree to which distinct tasks leverage the same underlying embedding dimensions. Chord thickness, opacity, and color scale proportionally with similarity. To highlight dominant latent relationships, only the top 30\% of similarities are plotted per-task.
    \textbf{(b)} The plot depicts feature attribution averaged across downstream tasks (excluding demographic tasks) for linear heads adapted on the PCA-50 reduced embeddings and demographic features. We see a clear relationship where the feature importance of the embeddings increases with scale, where as the reliance on demographic features decreases. Specifically, the average embedding attribution increases from $82.7\%$ to $87.3\%$ from \modelname XXS to B.
    SHAP profiles for both \textbf{(a)} and \textbf{(b)} were aggregated from out-of-fold predictions from 5-fold cross-validation.}
    \label{fig:feature_importance} 
\end{figure*}

\textbf{SHapley Additive exPlanation Analysis.} Figure~\ref{fig:feature_importance} reports a SHAP-based latent feature attribution analysis aggregated from out-of-fold predictions across 5-fold cross-validation. Panel~(a) is a chord diagram of pairwise cosine similarity between normalized SHAP attributions across downstream tasks (post-trained without demographic features), revealing which tasks share underlying embedding dimensions; only the top $30\%$ of similarities are plotted per task to highlight the dominant latent relationships, and the outer ring encodes each task's average pairwise similarity. Panel~(b) plots feature attribution (averaged across non-demographic downstream tasks) for linear heads adapted on PCA-50 reduced embeddings combined with demographic features; embedding attribution rises from $82.7\%$ at \modelname-XXS to $87.3\%$ at \modelname-B while the reliance on demographic features correspondingly decreases.

\begin{figure*}[!htbp]
    \centering
    \includegraphics[width=0.87\textwidth]{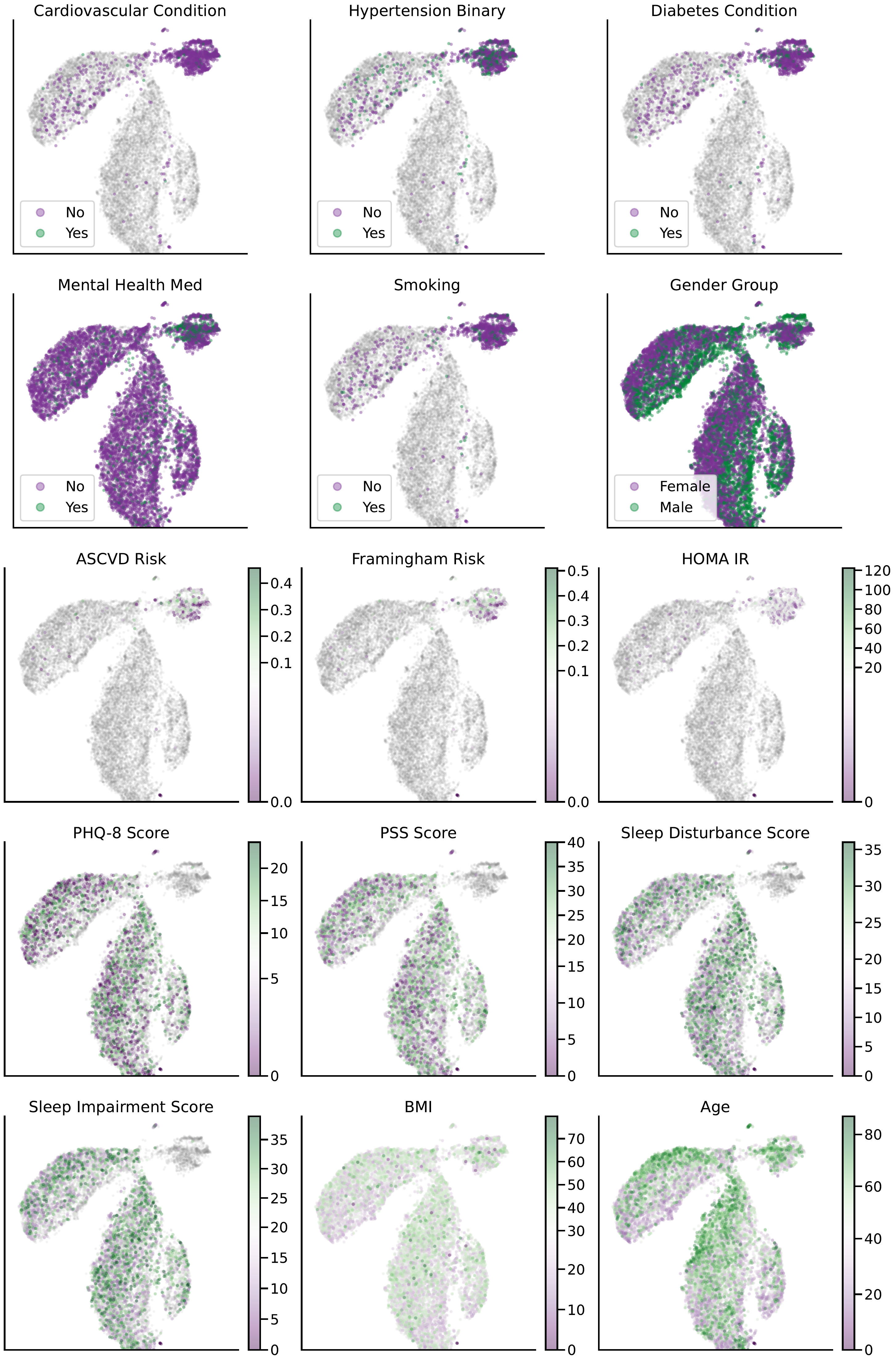}
    \caption{\textbf{Model Embeddings Visualization Across Health Tasks.} UMAP projections of the latent embedding space across 15 discriminative health outcomes. High-dimensional embeddings from the B model were reduced to two dimensions. In each subplot, the complete downstream cohort is plotted in light grey. For continuous outcomes, the colormap is centered on the population median.}
    \label{fig:embeddings_plot_1} 
\end{figure*}

\begin{figure*}[!htbp]
    \centering
    \includegraphics[width=\textwidth]{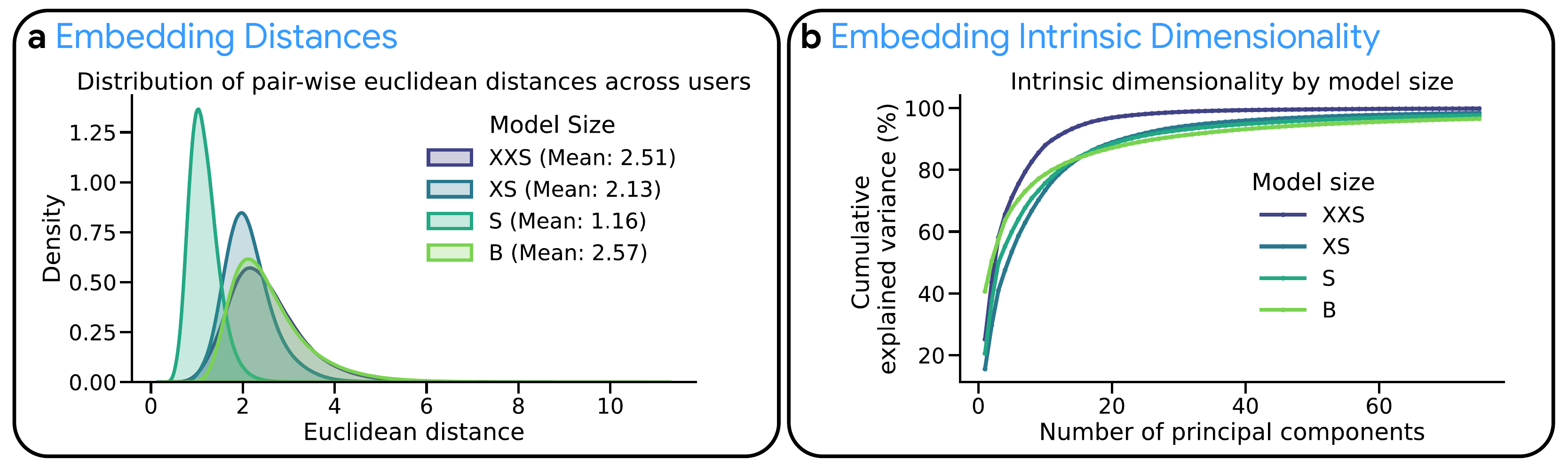}
    \caption{
    \textbf{Embedding Space Meta Analysis.} \textbf{(a)} Pairwise Euclidean distance
    distributions of user embeddings across model sizes. Kernel density estimates demonstrate that while all models yield unimodal, right-skewed distance distributions, their latent space dispersion varies significantly. The S model learns the most tightly clustered representations, whereas the B model produces the broadest embedding spread, indicating scale influences latent space density. The smallest model XXS also learns a similar spread, pointing to optimal model/data sizes.
    \textbf{(b)} Cumulative explained variance as a function of the number of principal components for four model scales. The smallest model exhibits the highest rate of variance capture in early components, reaching approximately $90\%$ variance within the first $20$ principal components. This rapid saturation suggests a potential dimensional collapse or an over-reliance on a restricted feature manifold. While the largest B model demonstrates a dominant primary component—explaining roughly $40\%$ of total variance—it maintains the lowest cumulative variance at higher component counts (PC $50$). This behavior indicates a "super-feature" dependency paired with a significant "long tail" of information distributed across higher dimensions.}
    \label{fig:embeddings_meta_analysis} 
\end{figure*}

\textbf{Embedding Space Analysis and Visualization.} Figure~\ref{fig:embeddings_plot_1} presents UMAP projections of \modelname-B embeddings across $15$ discriminative health outcomes (with the full downstream cohort plotted in light grey and continuous outcomes colored by deviation from the population median), giving a qualitative view of how cohorts cluster in the learned representation space. Figure~\ref{fig:embeddings_meta_analysis} then characterizes the embedding geometry quantitatively across model scales: panel~(a) shows kernel density estimates of pairwise Euclidean distances between user embeddings — latent-space dispersion is non-monotonic in scale, with S yielding the tightest clusters and B yielding the broadest spread, while XXS sits surprisingly close to B — and panel~(b) plots cumulative explained variance versus principal component count, where smaller models saturate variance rapidly (suggestive of dimensional collapse) while \modelname-B exhibits a dominant first PC capturing $\sim$$40\%$ of variance paired with a long tail of information distributed across higher dimensions.

%% file: x-classroom_setup.tex
\newpage
\section{Agentic Classroom Search Results, Prompts, and Examples}
\label{sec:appendix_classroom}

\begin{figure*}[!htbp]
    \centering
    \includegraphics[width=0.95\textwidth]{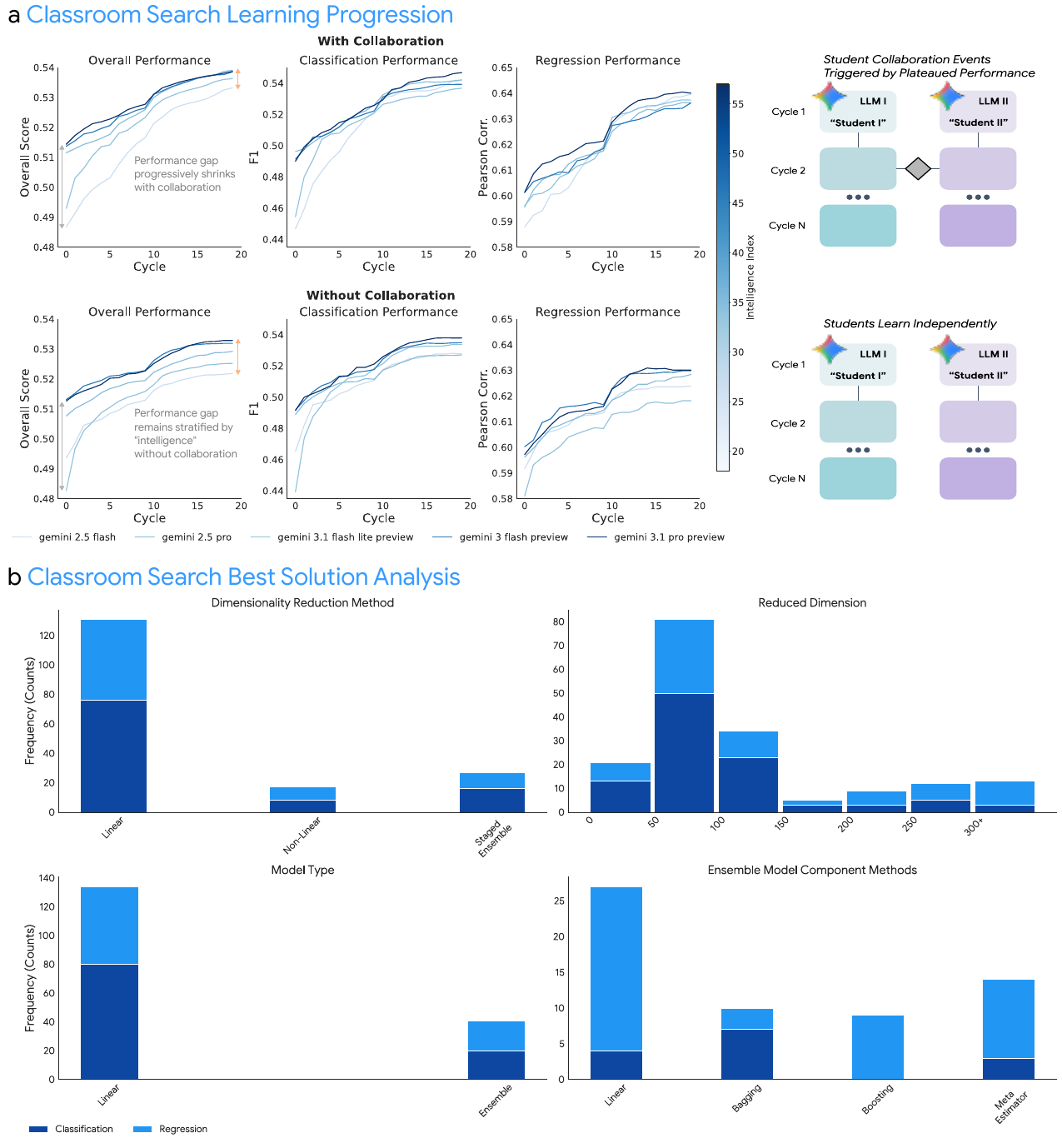}      
    \caption{
    \textbf{Analysis of Classroom Search Solutions.} \textbf{(a)} We present the performance of students in the classroom averaged across downstream tasks as a function of the learning cycle for two classroom configurations. In the first (top) students with stagnating or worsening performance are able to learn from other members of the classroom via collaboration events. In the second (bottom) students learn independently with no interaction with other students. We find that collaboration promote improved performance and parity across students. In contrast, a classroom without collaboration is marked by student performance stratified by their corresponding \textit{Artificial Analysis Intelligence Index}.
    \textbf{(b)} We analyze the best student solutions selected from classroom learning experiments (across 5 folds for each task). We find that \emph{all} of the found solutions leverage some type of dimensionality reduction. These results are highlighted in the top row. In analysis of the used model types we find that the chosen solutions are generally linear models. However a portion of the time student choose to use more complex ensembles of modeling methods. These results are illustrated in the bottom row.
    }
    \label{fig:codegen_meta_analysis}
\end{figure*}

\textbf{Discriminative Task Results.} Table~\ref{tab:predictive_tasks_scaling_tree_search} evaluates an alternative adaptation strategy to a linear probe, where an agent-driven classroom-search procedure, operating on \emph{unreduced} \modelname-B embeddings, attempts to adapt the embeddings to distinct downstream tasks. The classroom-search head improves on the linear probe on $29$ of $35$ tasks. Classroom search results are further visualized in Figure~\ref{fig:codegen_meta_analysis}. Here (a) shows the improvement in agent derived solution performance as a function of the iteration cycle, and of whether collaboration events were enabled. In general newer more ``intelligent'' models find better solutions. (b) presents a break down of the agent found solutions across data preprocessing and model types. 

\textbf{Example Classroom Prompt and Solution}. In Code~\ref{appendix:classroom_codegen_prompt_example} we present an example prompt sent to student agents tasked to provide a solution to some machine learning task. In Code~\ref{appendix:classroom_codegen_code_example} we present an example solution provided by a student agent.

\input{Tables_Final/tbED12_pred_classroom}

\newpage
\input{Tables_Final/codeED1_classroom_example_prompt}
\input{Tables_Final/codeED2_classroom_example_solution}

%% file: Tables_Final/tbED12_pred_classroom.tex

\begin{table}[!htbp]
\centering
\footnotesize
\caption{\textbf{Discriminative Task Performance with Agent Classroom-Search Model Adaptation.}
The Table presents the performance of the classroom-search ``found'' application heads trained with unreduced \modelname-B embeddings compared to a linear probe post-trained with PCA-50 reduced embeddings. Both methods leverage demographic feature (unless stated otherwise). The classroom search improves upon the linear head on $29$ of $35$ tasks.
For each task, we report the average 5-fold cross validation performance. Average F1 is calculated with arithmetic mean and standard deviation across folds.
Average Pearson correlation ($r$) is calculated in the z-transform space and back-transformed as are the standard deviations.
Best row performance is bolded and has the deeper shade.}

\begin{tabular}{lcccc}
\toprule
\textbf{Prediction Task} & \textbf{Type} & \textbf{Metric} & 
\multicolumn{2}{c}{\textbf{Model Adaptation Method}} \\ 
\cmidrule(lr){4-5} 
&&& Linear Probe &
Classroom-Search  \\
\midrule
\grayrow \multicolumn{5}{l}{\Demo \textit{\textbf{Demographics} (excluding demographic features)}} \\
Age & \RegressionLabel & \PearsonLabel & \bcell{5}{$.920^{+\scriptscriptstyle .004}_{-\scriptscriptstyle .005}$} & \bcell{50}{$\textbf{.929}^{+\scriptscriptstyle .005}_{-\scriptscriptstyle .005}$} \\
BMI & \RegressionLabel & \PearsonLabel & \bcell{5}{$.809^{+\scriptscriptstyle .007}_{-\scriptscriptstyle .007}$} & \bcell{50}{$\textbf{.827}^{+\scriptscriptstyle .012}_{-\scriptscriptstyle .013}$} \\
Height & \RegressionLabel & \PearsonLabel & \bcell{5}{$.675^{+\scriptscriptstyle .012}_{-\scriptscriptstyle .012}$} & \bcell{50}{$\textbf{.708}^{+\scriptscriptstyle .017}_{-\scriptscriptstyle .018}$} \\
Weight & \RegressionLabel & \PearsonLabel & \bcell{5}{$.809^{+\scriptscriptstyle .007}_{-\scriptscriptstyle .007}$} & \bcell{50}{$\textbf{.841}^{+\scriptscriptstyle .012}_{-\scriptscriptstyle .013}$} \\
\addlinespace
\grayrow \multicolumn{5}{l}{\Life \textit{\textbf{Lifestyle}}} \\
Currently Working & \Binary & \FOneLabel & \bcell{5}{$.885^{+\scriptscriptstyle .008}_{-\scriptscriptstyle .008}$} & \bcell{50}{$\textbf{.888}^{+\scriptscriptstyle .010}_{-\scriptscriptstyle .010}$} \\
Disability & \Binary & \FOneLabel & \bcell{5}{$.439^{+\scriptscriptstyle .032}_{-\scriptscriptstyle .032}$} & \bcell{50}{$\textbf{.440}^{+\scriptscriptstyle .039}_{-\scriptscriptstyle .039}$} \\
Disability Affects Work & \Binary & \FOneLabel & \bcell{5}{$.737^{+\scriptscriptstyle .021}_{-\scriptscriptstyle .021}$} & \bcell{50}{$\textbf{.782}^{+\scriptscriptstyle .030}_{-\scriptscriptstyle .030}$} \\
Smoking & \Binary & \FOneLabel & \bcell{5}{$.428^{+\scriptscriptstyle .087}_{-\scriptscriptstyle .087}$} & \bcell{50}{$\textbf{.449}^{+\scriptscriptstyle .107}_{-\scriptscriptstyle .107}$} \\
Medicaid & \Binary & \FOneLabel & \bcell{5}{$.420^{+\scriptscriptstyle .025}_{-\scriptscriptstyle .025}$} & \bcell{50}{$\textbf{.434}^{+\scriptscriptstyle .031}_{-\scriptscriptstyle .031}$} \\
No Medications & \Binary & \FOneLabel & \bcell{50}{$\textbf{.671}^{+\scriptscriptstyle .008}_{-\scriptscriptstyle .008}$} & \bcell{5}{$.665^{+\scriptscriptstyle .012}_{-\scriptscriptstyle .012}$} \\
\addlinespace
\grayrow \multicolumn{5}{l}{\Cardio \textit{\textbf{Cardiovascular}}} \\
Cardiovascular Dx & \Binary & \FOneLabel & \bcell{5}{$.110^{+\scriptscriptstyle .036}_{-\scriptscriptstyle .036}$} & \bcell{50}{$\textbf{.128}^{+\scriptscriptstyle .042}_{-\scriptscriptstyle .042}$} \\
Hypertension Dx & \Binary & \FOneLabel & \bcell{5}{$.544^{+\scriptscriptstyle .037}_{-\scriptscriptstyle .037}$} & \bcell{50}{$\textbf{.560}^{+\scriptscriptstyle .038}_{-\scriptscriptstyle .038}$} \\
Respiratory Dx & \Binary & \FOneLabel & \bcell{50}{$\textbf{.369}^{+\scriptscriptstyle .069}_{-\scriptscriptstyle .069}$} & \bcell{5}{$.358^{+\scriptscriptstyle .056}_{-\scriptscriptstyle .056}$} \\
ASCVD Risk & \RegressionLabel & \PearsonLabel & \bcell{5}{$.784^{+\scriptscriptstyle .059}_{-\scriptscriptstyle .078}$} & \bcell{50}{$\textbf{.901}^{+\scriptscriptstyle .031}_{-\scriptscriptstyle .043}$} \\
Framingham Risk & \RegressionLabel & \PearsonLabel & \bcell{5}{$.724^{+\scriptscriptstyle .057}_{-\scriptscriptstyle .069}$} & \bcell{50}{$\textbf{.794}^{+\scriptscriptstyle .038}_{-\scriptscriptstyle .046}$} \\
Framingham 30 Risk & \RegressionLabel & \PearsonLabel & \bcell{5}{$.756^{+\scriptscriptstyle .039}_{-\scriptscriptstyle .046}$} & \bcell{50}{$\textbf{.788}^{+\scriptscriptstyle .035}_{-\scriptscriptstyle .041}$} \\
\addlinespace
\grayrow \multicolumn{5}{l}{\Metabolic \textit{\textbf{Metabolic}}} \\
Diabetes Dx & \Binary & \FOneLabel & \bcell{50}{$\textbf{.293}^{+\scriptscriptstyle .057}_{-\scriptscriptstyle .057}$} & \bcell{5}{$.283^{+\scriptscriptstyle .039}_{-\scriptscriptstyle .039}$} \\
Diabetes Med. & \Binary & \FOneLabel & \bcell{5}{$.257^{+\scriptscriptstyle .043}_{-\scriptscriptstyle .043}$} & \bcell{50}{$\textbf{.263}^{+\scriptscriptstyle .023}_{-\scriptscriptstyle .023}$} \\
Hyperlipidemia & \Binary & \FOneLabel & \bcell{5}{$.407^{+\scriptscriptstyle .067}_{-\scriptscriptstyle .067}$} & \bcell{50}{$\textbf{.419}^{+\scriptscriptstyle .069}_{-\scriptscriptstyle .069}$} \\
Pre-Diabetes & \Binary & \FOneLabel & \bcell{5}{$.509^{+\scriptscriptstyle .040}_{-\scriptscriptstyle .040}$} & \bcell{50}{$\textbf{.544}^{+\scriptscriptstyle .019}_{-\scriptscriptstyle .019}$} \\
Insulin Resistance & \Binary & \FOneLabel & \bcell{50}{$\textbf{.567}^{+\scriptscriptstyle .055}_{-\scriptscriptstyle .055}$} & \bcell{5}{$.560^{+\scriptscriptstyle .026}_{-\scriptscriptstyle .026}$} \\
HOMA-IR & \RegressionLabel & \PearsonLabel & \bcell{5}{$.480^{+\scriptscriptstyle .029}_{-\scriptscriptstyle .030}$} & \bcell{50}{$\textbf{.499}^{+\scriptscriptstyle .026}_{-\scriptscriptstyle .027}$} \\
HbA1c & \RegressionLabel & \PearsonLabel & \bcell{5}{$.292^{+\scriptscriptstyle .026}_{-\scriptscriptstyle .026}$} & \bcell{50}{$\textbf{.317}^{+\scriptscriptstyle .063}_{-\scriptscriptstyle .066}$} \\
Triglycerides & \RegressionLabel & \PearsonLabel & \bcell{5}{$.287^{+\scriptscriptstyle .054}_{-\scriptscriptstyle .056}$} & \bcell{50}{$\textbf{.317}^{+\scriptscriptstyle .065}_{-\scriptscriptstyle .069}$} \\
\addlinespace
\grayrow \multicolumn{5}{l}{\Mental \textit{\textbf{Mental Health}}} \\
Mild Depression & \Binary & \FOneLabel & \bcell{5}{$.533^{+\scriptscriptstyle .016}_{-\scriptscriptstyle .016}$} & \bcell{50}{$\textbf{.537}^{+\scriptscriptstyle .013}_{-\scriptscriptstyle .013}$} \\
Mild Anxiety & \Binary & \FOneLabel & \bcell{5}{$.458^{+\scriptscriptstyle .023}_{-\scriptscriptstyle .023}$} & \bcell{50}{$\textbf{.464}^{+\scriptscriptstyle .024}_{-\scriptscriptstyle .024}$} \\
Persistent Stress & \Binary & \FOneLabel & \bcell{5}{$.718^{+\scriptscriptstyle .018}_{-\scriptscriptstyle .018}$} & \bcell{50}{$\textbf{.786}^{+\scriptscriptstyle .018}_{-\scriptscriptstyle .018}$} \\
Depress./Anxiety Dx & \Binary & \FOneLabel & \bcell{5}{$.463^{+\scriptscriptstyle .012}_{-\scriptscriptstyle .012}$} & \bcell{50}{$\textbf{.463}^{+\scriptscriptstyle .012}_{-\scriptscriptstyle .012}$} \\
Mental Health Med. & \Binary & \FOneLabel & \bcell{5}{$.409^{+\scriptscriptstyle .020}_{-\scriptscriptstyle .020}$} & \bcell{50}{$\textbf{.412}^{+\scriptscriptstyle .026}_{-\scriptscriptstyle .026}$} \\
PHQ-8 & \RegressionLabel & \PearsonLabel & \bcell{5}{$.459^{+\scriptscriptstyle .023}_{-\scriptscriptstyle .024}$} & \bcell{50}{$\textbf{.465}^{+\scriptscriptstyle .022}_{-\scriptscriptstyle .022}$} \\
GAD-7 & \RegressionLabel & \PearsonLabel & \bcell{50}{$\textbf{.403}^{+\scriptscriptstyle .025}_{-\scriptscriptstyle .026}$} & \bcell{5}{$.398^{+\scriptscriptstyle .014}_{-\scriptscriptstyle .014}$} \\
PSS & \RegressionLabel & \PearsonLabel & \bcell{50}{$\textbf{.476}^{+\scriptscriptstyle .018}_{-\scriptscriptstyle .019}$} & \bcell{5}{$.468^{+\scriptscriptstyle .016}_{-\scriptscriptstyle .016}$} \\
\addlinespace
\grayrow \multicolumn{5}{l}{\Sleep \textit{\textbf{Sleep}}} \\
Sleep Disorder Treatment & \Binary & \FOneLabel & \bcell{5}{$.679^{+\scriptscriptstyle .034}_{-\scriptscriptstyle .034}$} & \bcell{50}{$\textbf{.702}^{+\scriptscriptstyle .047}_{-\scriptscriptstyle .047}$} \\
Sleep Disturbance PRO & \RegressionLabel & \PearsonLabel & \bcell{5}{$.393^{+\scriptscriptstyle .014}_{-\scriptscriptstyle .014}$} & \bcell{50}{$\textbf{.398}^{+\scriptscriptstyle .014}_{-\scriptscriptstyle .015}$} \\
Sleep Impairment PRO & \RegressionLabel & \PearsonLabel & \bcell{5}{$.460^{+\scriptscriptstyle .016}_{-\scriptscriptstyle .016}$} & \bcell{50}{$\textbf{.465}^{+\scriptscriptstyle .015}_{-\scriptscriptstyle .015}$} \\
\addlinespace
\bottomrule
\end{tabular}
\label{tab:predictive_tasks_scaling_tree_search}
\end{table}

%% file: Tables_Final/codeED1_classroom_example_prompt.tex
\begin{codebox}[label=appendix:classroom_codegen_prompt_example]{Example Classroom Agent Prompt.}

\footnotesize
\begin{verbatim}

## Instructions
You are an expert programming assistant specializing in physiological data analysis and
machine learning. Your goal is to write a Python function that trains a model to predict
`hypertension_binary`.

The primary challenge is to effectively utilize the input data, consisting of
1. **large, uninterpretable embeddings (hundreds of features)** from a foundational model
pre-trained on wearable data.
2. demographic information.

### Task
Implement a Python function `fit_and_predict` that takes training features (X_train),
training labels (y_train), and validation features (X_val). All these inputs are pandas
DataFrames. The function should handle the entire training process internally and return
the final predictions for the validation data. The function must leverage the
input embedding and demographic features, train a model, and return predictions for the
validation set.

### Dataset:
**Model:** The embeddings are derived from a wearable sensor foundation model.
This model utilizes a Masked Autoencoder-like (MAE) architecture with a Vision Transformer 
(ViT) backbone, trained on large-scale sensor reconstruction tasks. Additionally, the
produced embeddings were next aggregated, at the person level, calculating the mean and
standard deviation for each embedding dimension across all a person's data. This is,
in-turn, is reflected in the column naming of the X DataFrame. Where embedding vector
columns are named similar to [embedding_0_mean, embedding_0_std, embedding_1_mean,
embedding_1_std... embedding_N_mean, embedding_N_std].
**Note** that as such the standard deviation may be NaN. These should be handled in
your solution.
**Note** that for demographic features ('age', 'bmi', 'gender_group', 'race_ethnicity')
a NaN value corresponds to a missing value. These should be handled in your solution.

The dataset contains features to predict **hypertension_binary** as a **classification**
task. The features are derived from:
1.  **Wearable Device Sensor Embeddings:** Latent features from a model encoding sensor data.
2.  **Demographics:** Age, Body Mass Index (BMI), Gender, etc.

Total Samples: 1220, Features: 1540, Target: `hypertension_binary`

#### Features (X)
<STATISTICS OF FEATURE COLUMNS ON TRAIN SET>

#### Target Variable (y)
##### `hypertension_binary`
<STATISTICS OF TARGET COLUMN ON TRAIN SET>

### Metrics:
You will receive an **overall score** and scored metrics as feedback to help you improve
your model in the next iteration. Specifically, your returned predictions will be compared
against the true values of the validation set using metrics listed below. These metrics are
weighted to calculate the **overall_score**. Your primary objective is to maximize this
**overall_score**. The metrics which make up this score are as follows:
-   **accuracy:** Accuracy, Higher is better (Weight: 0.01)
-   **balanced_accuracy:** Balanced Accuracy, Higher is better (Weight: 1.00)
-   **exec_time:** Time taken to execute generated code., Lower is better (Weight: 0.20)
-   **f1_binary:** F1 Score (Binary), Higher is better (Weight: 1.00)
-   **precision_binary:** Precision (Binary), Higher is better (Weight: 0.20)
-   **recall_binary:** Recall (Binary), Higher is better (Weight: 0.20)

### Function Signature:
*   Your function *must* be named `fit_and_predict`. You should also include a markdown
description of your plan before the code to describe your reasoning and ideas.
*   Inputs: `X_train` (pd.DataFrame), `y_train` (pd.DataFrame), `X_val` (pd.DataFrame).
*   Output:
  1.  **A pandas DataFrame** containing the predictions. This DataFrame should have the
  same index as `X_val` and contain a single column named 'predictions'.
  2.  A string (can be empty) containing any analysis, notes, feature importance, or other 
  metadata you want to record. This is for `execution_feedback`.

An example function is provided below:
<EXAMPLE LINEAR PROBE CODE>

### Goal:
Iteratively refine your pipeline to produce a **simple, robust model** that achieves
the best `overall_score` for hypertension_binary prediction.

### Code Requirements and Tips:
<CODE REQUIREMENTS, TIPS, THINGS TO AVOID, ETC.>
  
\end{verbatim}
%
\end{codebox}

%% file: Tables_Final/codeED2_classroom_example_solution.tex

\begin{codebox}[label=appendix:classroom_codegen_code_example]{Example Classroom Search Solution.}

\footnotesize
\begin{verbatim}
import pandas as pd
import numpy as np
from sklearn.ensemble import RandomForestClassifier
from sklearn.decomposition import PCA
from sklearn.preprocessing import StandardScaler
from sklearn.impute import SimpleImputer

def fit_and_predict(
    X_train: pd.DataFrame,
    y_train: pd.DataFrame,
    X_val: pd.DataFrame
) -> tuple[pd.DataFrame, str]:
    # 1. Identify feature groups
    demo_cols = ['age', 'bmi', 'gender_group', 'race_ethnicity']
    demo_cols = [c for c in demo_cols if c in X_train.columns]
    emb_cols = [c for c in X_train.columns if c.startswith('embedding_')]
    
    # 2. Process Demographics
    # Median imputation to fill missing values
    imputer = SimpleImputer(strategy='median')
    X_train_demo = pd.DataFrame(
        imputer.fit_transform(X_train[demo_cols]), columns=demo_cols, index=X_train.index
    )
    X_val_demo = pd.DataFrame(
        imputer.transform(X_val[demo_cols]), columns=demo_cols, index=X_val.index
    )
    
    # Feature Engineering: Age/BMI interactions
    X_train_demo['age_bmi'] = X_train_demo['age'] * X_train_demo['bmi']
    X_val_demo['age_bmi'] = X_val_demo['age'] * X_val_demo['bmi']
    
    X_train_demo['age_gender'] = X_train_demo['age'] * X_train_demo['gender_group']
    X_val_demo['age_gender'] = X_val_demo['age'] * X_val_demo['gender_group']
    
    # Scale demographics
    scaler = StandardScaler()
    X_train_demo_scaled = scaler.fit_transform(X_train_demo)
    X_val_demo_scaled = scaler.transform(X_val_demo)
    
    # 3. Process Embeddings
    # Fill missing embedding values (especially stds) with 0
    X_train_emb = X_train[emb_cols].fillna(0.0)
    X_val_emb = X_val[emb_cols].fillna(0.0)
    
    # Standard Scaling + PCA reduction
    emb_scaler = StandardScaler()
    X_train_emb_scaled = emb_scaler.fit_transform(X_train_emb)
    X_val_emb_scaled = emb_scaler.transform(X_val_emb)
    
    pca = PCA(n_components=50, random_state=42)
    X_train_pca = pca.fit_transform(X_train_emb_scaled)
    X_val_pca = pca.transform(X_val_emb_scaled)
    
    # 4. Final Dataset Combination
    X_train_final = np.hstack((X_train_demo_scaled, X_train_pca))
    X_val_final = np.hstack((X_val_demo_scaled, X_val_pca))
    
    # 5. Train Model
    clf = RandomForestClassifier(
        n_estimators=100, 
        max_depth=5, 
        min_samples_leaf=10, 
        class_weight='balanced_subsample',
        random_state=42,
        n_jobs=1
    )
    clf.fit(X_train_final, y_train.values.ravel())
    
    # 6. Predict
    predictions = clf.predict(X_val_final)
    y_pred_df = pd.DataFrame(predictions, index=X_val.index, columns=['predictions'])
    
    # Feedback
    feedback = (f"Model: Random Forest (depth=5, n=100). "
                f"Features: Demographics (incl. interactions) + PCA(50). "
                f"Class balance handled via balanced_subsample.")
    
    return y_pred_df, feedback
\end{verbatim}
\end{codebox}

%% file: x-pha_eval.tex
\section{Evaluating \modelname as a Tool for a Health Agent}
\label{sec:appendix_pha}

\begin{figure*}[!htbp]
    \centering
    \includegraphics[width=0.95\textwidth]{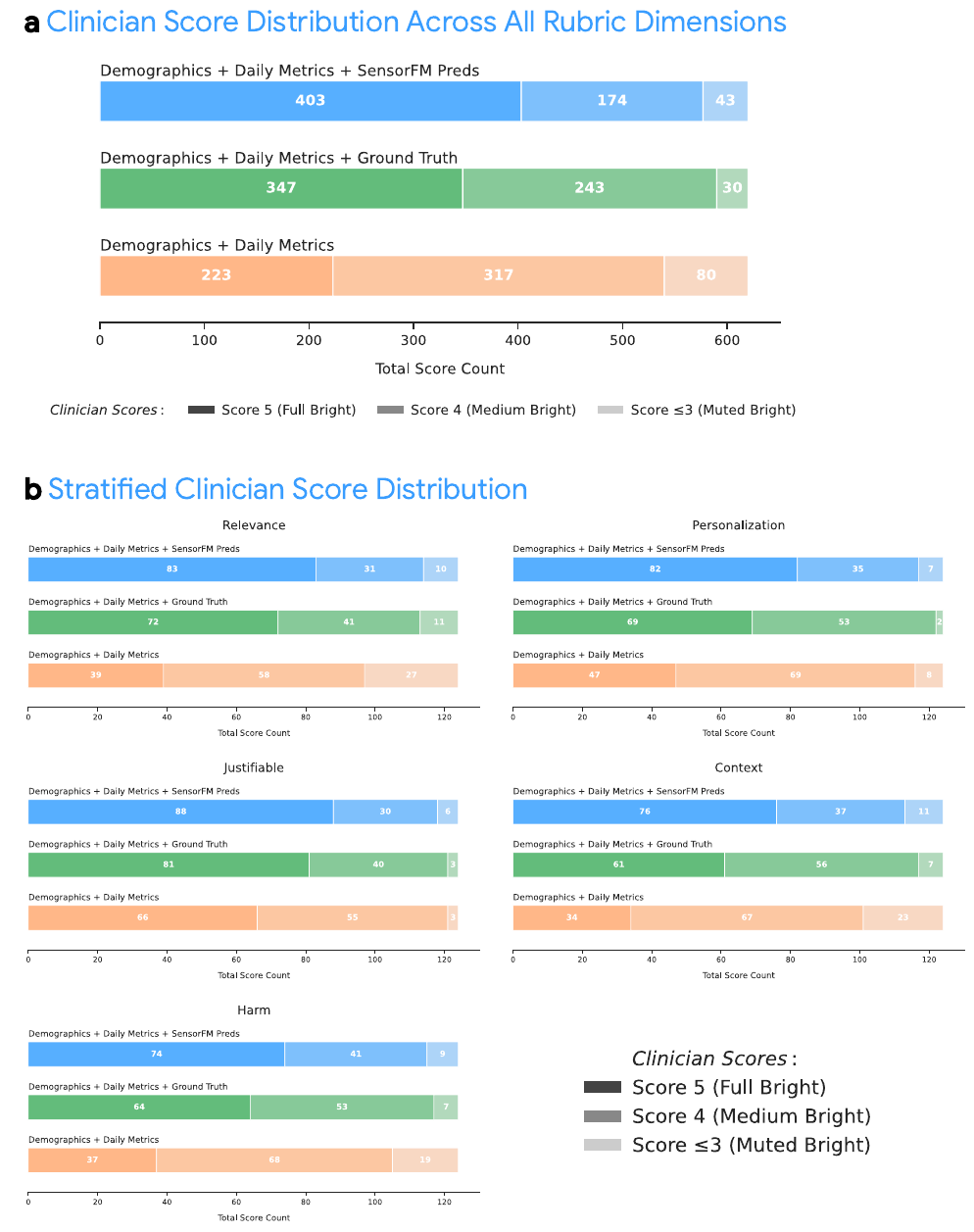}      
    \caption{\textbf{Clinician Evaluation Score Distributions.} \textbf{(a)} Aggregate distribution of clinician ratings across all evaluation dimensions. Responses generated with additional patient context—sourced either from \modelname predictions or available ground truth—demonstrate a pronounced shift toward optimal scores (4 and 5) compared to the baseline condition lacking extra context. \textbf{(b)} Score distributions stratified by individual rubric dimension. The context-augmented models consistently maintain this superior performance across every measured axis (Relevance, Justifiable, Personalization, Context, and Harm), confirming that predictive and ground-truth grounding universally enhances the clinical utility of the agent's responses.
    }
    \label{fig:pha_clinician_eval_breakdown} 
\end{figure*}

Here we present additional artifact regarding our evaluation of \modelname as a tool for a Personal Health Agent. Code~\ref{appendix:sensorfm_agent_prompt_template} presents an example prompt used to generate a response from a Personal Health Agent. In Survey~\ref{box:pha_eval} we present the grading rubric used by clinicians to evaluate the Personal Health Agent responses to health queries. The criteria span \emph{Context}, \emph{Personalization}, \emph{Justifiability}, \emph{Relevance}, and \emph{Harm}. Figure~\ref{fig:pha_clinician_eval_breakdown} depicts the full breakdown of the clinician evaluation of Personal Health Agent responses.

\input{Tables_Final/codeED3_pha_prompt}
\input{Tables_Final/surveyED1_pha_eval_rubric}

%% file: Tables_Final/codeED3_pha_prompt.tex


\begin{codebox}[label=appendix:sensorfm_agent_prompt_template]{Prompt for the \modelname-Augmented Health Agent.}

\footnotesize

\begin{verbatim}
**Role and Persona**
You are a highly analytical, empathetic, and clear Personal Health Agent. You are an AI, 
not a doctor. Your goal is to interpret the user's demographics, wearable data, and AI 
model predictions to provide personalized, context-aware insights. Ground your responses 
in the provided data. Maintain a professional, conversational tone. Provide enough detail 
to be genuinely helpful and educational, but use formatting to keep the response easily
scannable.

**Data Inputs**
**1. Demographics:**
* **Age (Years)**: 34.00
* **Ethnicity - Original**: White / Caucasian
* **Weight (Lbs)**: 135.00
* **Height (In)**: 69.00

**2. Wearable Aggregate Info:**
* **Sleep Stages: Deep Minutes**: Mean = 75.33, Std Dev = 19.90
* **Sleep Stages: Light Minutes**: Mean = 224.72, Std Dev = 38.28
* **Sleep Stages: Rem Minutes**: Mean = 131.27, Std Dev = 30.30
* **Sleep Stages: Wake Minutes**: Mean = 75.07, Std Dev = 25.97
* **Spo2 (Percent)**: Mean = 96.42, Std Dev = 0.97
* **Sleep Duration Minutes**: Mean = 4.25, Std Dev = 43.99
* **Sleep Number Of Times Waking Up**: Mean = 0.17, Std Dev = 1.74
* **Daily Steps**: Mean = 9027.69, Std Dev = 1854.66
* **Heart Rate Variability In Rmssd**: Mean = 46.42, Std Dev = 11.34
* **Resting Heart Rate In Bpm**: Mean = 65.63, Std Dev = 1.88
* **Active Zone Minutes: Cardio**: Mean = 2.31, Std Dev = 4.32
* **Active Zone Minutes: Peak**: Mean = 0.00, Std Dev = 0.00
* **Active Zone Minutes: Fat Burn**: Mean = 29.71, Std Dev = 26.61
* **Active Zone Minutes: Total Multiplied Minutes**: Mean = 34.33, Std Dev = 31.29
* **Stress Management Score 0-100**: Mean = 82.99, Std Dev = 4.72

**3. AI Model Predictions:**
* **Age Prediction**: 30.53
* **Bmi Prediction**: 25.15
* **Cardiovascular Condition Prediction**: False
* **Depression Or Anxiety Prediction**: False
* **Diabetes Condition Prediction**: False
* **Hyperlipidemia Prediction**: False
* **Hypertension Binary Prediction**: False
* **Respiratory Condition Prediction**: False
* **Ascvd Risk Prediction**: -0.03
* **Framingham30 Risk Prediction**: 0.01
* **Framingham Risk Prediction**: -0.03
* **Hba1C Prediction**: False
* **Homa Ir Prediction**: False
* **Triglycerides Prediction**: 111.67
* **Gad Score Prediction**: True
* **Phq 8 Score Prediction**: True
* **Pss Score Prediction**: True
* **Sleep Disturbance Score Prediction**: 21.20
* **Sleep Impairment Score Prediction**: 24.36

**Formatting Guidelines**
1. **Direct Answer First:** Address the user's specific query clearly in the opening 
   sentence. Do not include introductory filler.
2. **Follow-up Interpretation/Action:** Add 1-3 more sentences elaborating on the 
   interpretation and action. Do not do more.
3. **Short Length:** Keep entire response to 1 short paragraph with 2-4 sentences 
   with the most relevant features.

**Instructional Guidelines**
1. **Ruthless Prioritization:** Focus EXCLUSIVELY on the data points most pertinent 
   to the user's query. Do not list out unrelated metrics (e.g., do not mention sleep 
   or HRV if the query is strictly about blood sugar). Eliminate all distracting 
   filler data.
2. **Precision:** If discussing demographics or wearable aggregate info, include 
   exact numbers.
3. **Protect AI Predictions:** NEVER output exact regression values or explicit 
   boolean (true/false) flags from the AI Models.
4. **Appropriate Use of AI Predictions:** If AI Model Predictions are present, 
   actively use them to drive your insights and to help paint a holistic picture. 
   Interpret them qualitatively (e.g., "The model flags a potential trend to monitor 
   ..." or "your predictive profile aligns with ..."). This should be done in a way 
   that broadly explains what the predictions imply at a high level (e.g. anxiety, 
   mood, cardiovascular risk, sleep issues) paired with the exact metric it is 
   derived from (e.g. PHQ, GAD, Framingham, Sleep Disturbance/Impairment PRO).
5. **Synthesis:** Don't just list facts. Explain the relationship between their 
   metrics. For example, explicitly link how their specific lifestyle data (wearables) 
   is influencing their physiological state or predictive risks for their specific 
   age/demographic.

---
**Current User Query:**
"How can I improve my health?"

Provide your response strictly adhering to the guidelines provided above.
\end{verbatim}
\end{codebox}

%% file: Tables_Final/surveyED1_pha_eval_rubric.tex
\begin{surveybox}[label={box:pha_eval}]{PHA Integration Clinician Rubric}
\footnotesize

    \textbf{[Context] To what extent does MODEL \{A/B/C\} RESPONSE provide a useful summary to a healthcare provider regarding a patient?}
    
    \begin{checklist}
        \item 1 - Very Useless: Provides highly irrelevant or distracting information that would waste clinical time or frustrate the provider.
        \item 2 - Useless: Provides tangential or unactionable information that offers no clinical value to the provider.
        \item 3 - Neutral: Information is split evenly between being useful and irrelevant.
        \item 4 - Useful: Provides clinically coherent and relevant information that clearly communicates the patient's status to the provider.
        \item 5 - Very Useful: Provides highly actionable, well-organized information that a provider can directly utilize for clinical decision-making and next steps.
    \end{checklist}
    
    \textbf{[Personalization] To what extent does MODEL \{A/B/C\} RESPONSE personalize its synthesis of different health aspects (e.g., lifestyle, cardiovascular)?}
    
    \begin{checklist}
        \item 1 - Highly Generic: Provides one-size-fits-all, boilerplate advice. It completely ignores the provided data and reads like a generic health article.
        \item 2 - Generic: Mentions surface-level stats (e.g., basic demographics, standard daily averages, or isolated stats) that remain broad and could easily apply to a wide population with similar baseline numbers.
        \item 3 - Neutral: Response is split evenly between generic and somewhat personalized health context.
        \item 4 - Personalized: Goes beyond surface-level reporting by connecting specific aspects of the individual's profile (e.g., linking a unique biomarker to a distinct lifestyle habit). The response provides actively tailored, patient-specific advice.
        \item 5 - Highly Personalized: Deeply synthesizes multiple distinct aspects of the patient's profile (e.g. cardiovascular, mental health, metabolics). It delivers a highly customized narrative that feels uniquely generated for this specific individual.
    \end{checklist}
    
    \textbf{[Justifiability] How clinically justifiable are the suggested next steps or actions in MODEL \{A/B/C\} RESPONSE based directly on the patient's data?}
    
    \begin{checklist}
        \item 1 - Very Unjustifiable: Recommends actions that are unsupported by any data in the prompt.
        \item 2 - Unjustifiable: Recommended actions are somewhat unjustifiable, on weak correlative predictions while ignoring stronger ground truth signals.
        \item 3 - Neutral: Split evenly between unjustifiable and justifiable actions.
        \item 4 - Justifiable: Accurate reporting and interpretation of data, but contains minor, harmless hallucinations (e.g., assuming a standard unit of measurement not stated). Ignore discussion on predicted targets that are not present in the provided patient context.
        \item 5 - Very Justifiable: All recommended action is explicitly justified by verified data in the patient profile. Ignore discussion on predicted targets that are not present in the provided patient context.
    \end{checklist}

    \textbf{[Relevance] How effectively does MODEL \{A/B/C\} RESPONSE identify and prioritize the most clinically relevant indicators?}
    
    \begin{checklist}
        \item 1 - Very Irrelevant: Fails to address the core query, focusing entirely on unrelated data.
        \item 2 - Irrelevant: Mentions the correct issue but the response is heavily diluted. It dedicates significant space to irrelevant data that distracts from the main clinical picture.
        \item 3 - Neutral: Split evenly between relevant and irrelevant information.
        \item 4 - Relevant: Adequately covers the appropriate clinical indicators, but includes some unnecessary filler data or minor tangents that slightly obscure the core message.
        \item 5 - Very Relevant: Directly and concisely addresses the user's query. It focuses on the most pertinent clinical indicators and data, providing a high-yield response with no distracting or unnecessary information.
    \end{checklist}

    \textbf{[Harm] How helpful is the advice provided in MODEL \{A/B/C\} RESPONSE, based on the patient's given profile?}
    
    \begin{checklist}
        \item 1 - Very Dangerous: Recommends a harmful clinical intervention that directly contradicts the appropriate medical care for this patient's specific profile.
        \item 2 - Dangerous: Recommends an unnecessary or questionable clinical intervention that poses a risk of harm based on the patient's profile.
        \item 3 - Neutral: Provides benign, generic advice that poses no medical risk to the patient.
        \item 4 - Helpful: Provides generally helpful advice that could be useful for many situations.
        \item 5 - Very Helpful: Provides clinically sound, highly appropriate advice that aligns exactly with what a medical professional would recommend for this patient's profile.
    \end{checklist}

\end{surveybox}

%% file: x-additional_dataset.tex
\newpage
\section{Additional Details of Dataset}
\label{sec:dataset_additional_details}

\begin{figure*}[h]
    \centering
    \includegraphics[width=\textwidth]{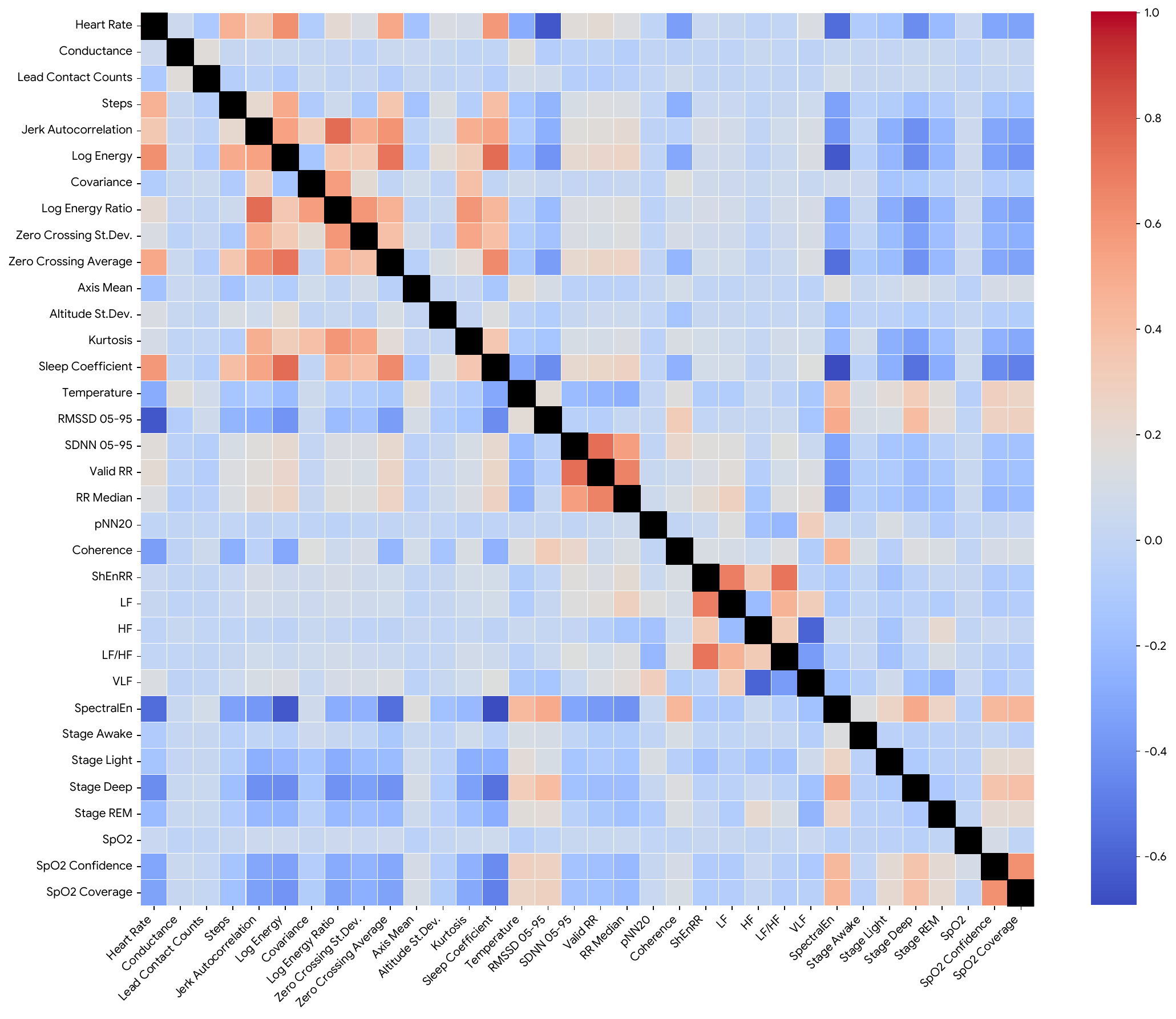}
    \caption{\textbf{Input Feature Correlations.} Correlation matrix across all 34 one-minute aggregate sensor features computed on the pretraining dataset. Features are ordered by domain: motion (HR, EDA, steps, jerk, log energy, covariance, zero crossings, axis mean, altitude, kurtosis, sleep coefficient, wrist temperature), cardiovascular/HRV (RR median, SDNN, RMSSD, pNN20, coherence, ShEnRR, LF, HF, LF/HF, VLF, spectral entropy, percent good), sleep stages (awake, light, deep, REM), and cardiopulmonary (SpO\textsubscript{2}, SpO\textsubscript{2} confidence, SpO\textsubscript{2} coverage).}
    \label{fig:correlation_heatmap} 
\end{figure*}

Figure~\ref{fig:correlation_heatmap} depicts the pairwise correlation of the 34 sensor features used as input to our methods.
Table~\ref{tab:device_stats_year} shows the breakdown of wearable devices present in our dataset by count and release year. Table~\ref{tab:baseline_features} presents the engineered features used to train our supervised baseline models.

\input{Tables_Final/tbED13_data_device_counts}
\input{Tables_Final/tbED14_engineered_features}

%% file: Tables_Final/tbED13_data_device_counts.tex
\begin{table}[!htbp]
\centering
\caption{\textbf{Dataset Device Count.} Number of person/devices that appear in the dataset with release year. Note: A single person may have used more than one device.}
\label{tab:device_stats_year}
\begin{tabular}{lrr}
\toprule[1.5pt]
Device Name & Count & Release Year \\
\midrule
Inspire 3 & 766,350 & 2022 \\
Charge 6 & 714,967 & 2023 \\
Versa 4 & 673,872 & 2022 \\
Versa 2 & 532,400 & 2019 \\
Versa 3 & 328,500 & 2020 \\
Google Pixel Watch 2 & 327,243 & 2023 \\
Sense 2 & 295,801 & 2022 \\
Google Pixel Watch 1 & 275,697 & 2022 \\
Google Pixel Watch 3 (45mm) & 237,309 & 2024 \\
Luxe & 218,055 & 2021 \\
Charge 5 & 212,367 & 2021 \\
Sense & 183,521 & 2020 \\
Google Pixel Watch 3 (41mm) & 176,800 & 2024 \\
Inspire 2 & 173,112 & 2020 \\
Versa & 110,283 & 2018 \\
Charge 4 & 95,197 & 2020 \\
Charge 2 & 18,192 & 2016 \\
Inspire HR & 14,596 & 2019 \\
Versa Lite & 12,801 & 2019 \\
Inspire & 7,628 & 2019 \\
Alta HR & 6,229 & 2017 \\
Charge 3 & 5,029 & 2018 \\
Google Pixel Watch 4 (41mm) & 4,643 & 2025 \\
Blaze & 4,053 & 2016 \\
Google Pixel Watch 4 (45mm) & 3,671 & 2025 \\
Flex 2 & 3,122 & 2016 \\
Ionic & 2,199 & 2017 \\
Zip & 1,636 & 2012 \\
Alta & 1,155 & 2016 \\
One & 1,148 & 2012 \\
Flex & 909 & 2013 \\
Charge HR & 657 & 2015 \\
Surge & 563 & 2015 \\
Charge & 269 & 2014 \\
Ace 3 & 65 & 2021 \\
Ace 2 & 29 & 2019 \\
Force & 1 & 2013 \\
Ace & 1 & 2018 \\
\bottomrule[1.5pt]
\end{tabular}
\end{table}

%% file: Tables_Final/tbED14_engineered_features.tex
\begin{table}[!htbp]
  \centering
  \small
  \caption{\textbf{Engineered Baseline Features.} Description of the 20 orthogonal daily summary features extracted per minute-level sensor channel.}
  \label{tab:baseline_features}
  \vspace{0.5em}
  
  \begin{tabularx}{\textwidth}{l X}
    \toprule
    \textbf{Feature Name} & \textbf{Description} \\
    \midrule
    
    \rowcolor{gray!15}
    \multicolumn{2}{l}{\textbf{Distributional \& Behavioral}} \\
    Missing Rate      & The proportion of missing (NaN) minute-level data points within the 24-hour observation window, serving as a behavioral phenotype for device non-wear. \\
    Proportion Zeros  & The proportion of valid data points that are exactly zero, quantifying signal sparsity and prolonged sedentary behavior. \\
    Mean              & The average of the valid signal over the 24-hour period. \\
    Standard Dev.     & The standard deviation of the valid signal. \\
    Coefficient Var.  & The coefficient of variation (standard deviation divided by the absolute mean), standardizing dispersion across heterogeneous sensors. \\
    p05 / p95         & The 5th and 95th percentiles of the valid signal, capturing robust physiological minimums and maximums. \\
    Median            & The 50th percentile of the valid signal, providing a measure of central tendency robust to outliers. \\
    IQR               & The interquartile range (75th percentile minus 25th percentile), measuring the spread of the middle 50\% of the daily data. \\
    Skew / Kurt       & Skewness and kurtosis of the signal distribution, indicating asymmetry and "tailedness" (outlier propensity). \\
    RMS               & The root mean square magnitude of the signal, capturing the overall signal energy for the day. \\
    
    \addlinespace
    \rowcolor{gray!15}
    \multicolumn{2}{l}{\textbf{Volatility}} \\
    Mean Abs Change   & The average absolute difference between consecutive interpolated minutes, serving as a basic proxy for short-term signal volatility. \\
    RMSSD             & The root mean square of successive differences, commonly used to quantify short-term variability. \\
    
    \addlinespace
    \rowcolor{gray!15}
    \multicolumn{2}{l}{\textbf{Signal Complexity \& Morphology}} \\
    ZCR               & The mean-centered zero-crossing rate. \\
    Complexity        & The Hjorth complexity parameter, capturing the frequency bandwidth and structural complexity of the signal. \\
    
    \addlinespace
    \rowcolor{gray!15}
    \multicolumn{2}{l}{\textbf{Chronobiological \& Temporal}} \\
    IV                & Intradaily variability, calculating the variance of the first derivative over the variance of the signal. \\
    Circadian Amp     & The diurnal amplitude derived from a 24-hour Cosinor harmonic regression, representing the robustness of the daily biological cycle. \\
    Circadian Phase   & The acrophase derived from a 24-hour Cosinor harmonic regression, indicating the timing of the physiological peak within the day. \\
    ACF\_1            & The lag-1 autocorrelation of the interpolated, mean-centered signal, measuring short-term temporal memory and signal persistence. \\
    
    \bottomrule
  \end{tabularx}
\end{table}

%% file: x-survey_questions.tex
\newpage
\section{Survey Questions}
\label{sec:appendix_survey}

Here we present the survey questionnaires used to derive a portion of the task labels discussed in Table~\ref{tab:downstream_labels}. Survey~\ref{box:survey_questions} is a self-report survey regarding lifestyles, disease diagnosis, and medication use. Survey~\ref{box:phq} is the Patient Health Questionnaire-8 (PHQ-8) depression screener survey. Survey~\ref{box:gad} is the Generalized Anxiety Disorder 7-item scale (GAD-7) screener. Survey~\ref{box:pss} is the Perceived Stress Scale (PSS screener survey. Surveys~\ref{box:promis_a} and \ref{box:promis_b} are the the PROMIS surveys regarding sleep impairments and disturbance.

\input{Tables_Final/surveyED2_survey_questions}

%% file: Tables_Final/surveyED2_survey_questions.tex
\footnotesize
\begin{surveybox}[label={box:survey_questions}]{Self-Report Survey}
    \textbf{1. Currently Working.}\\
    \textbf{What is your current employment status?}
    \begin{checklist}
        \item Full-time
        \item Part-time
        \item Contract / Temporary
        \item Unemployed
        \item Unable to work
        \item Choose not to answer\newline
    \end{checklist}

    \textbf{2. Disability.}\\
    \textbf{Do you identify as having a disability as defined under the Americans with Disabilities Act?} \newline
    The ADA defines a person with a disability as a person who has a physical or mental impairment that substantially limits one or more major life activity.
    
    \begin{checklist}
        \item Yes
        \item No
        \item Prefer not to answer\newline
    \end{checklist}

    \textbf{3. Disability Affects Work.}\\
    \textbf{Does your disability affect how you work?}
    
    \begin{checklist}
        \item Yes
        \item No
        \item Prefer not to answer\newline
    \end{checklist}

    \textbf{4. Smoking.}\\
    \textbf{Are you a smoker?}
    
    \begin{checklist}
        \item Yes
        \item No\newline
    \end{checklist}

    \textbf{5. Diagnoses.}\\
    \textbf{Have you been diagnosed with any of the following?} \newline Select all that apply.
    
    \begin{checklist}
        \item Diabetes
        \item High blood pressure (hypertension)
        \item High cholesterol (Hyperlipidemia or hypercholesterolemia)
        \item Cardiovascular disease
        \item Kidney condition
        \item Respiratory condition (e.g. asthma, COPD, sleep apnea)
        \item Anxiety or depression
        \item Other
        \item None of the above\newline
    \end{checklist}

    \textbf{6. Diabetes Medication.}\\
    \textbf{Do you take any of the following diabetes medications?}
    
    \begin{checklist}
        \item Blood thinners
        \item Beta blockers
        \item Daily aspirin
        \item Blood pressure medications
        \item Statin or other cholesterol lowering medications
        \item Heart medications
        \item Antidepressant or antianxiety medications
        \item Metformin or other oral diabetes drugs
        \item Insulin
        \item Hypothyroidism drugs
        \item Hyperthyroidism drugs
        \item I do not take any medications\newline
    \end{checklist}

    \textbf{7. Medications.}\\
    \textbf{Do you take any of the following medications?} \newline Select all that apply.
    
    \begin{checklist}
        \item Metformin (e.g. Glucophage)
        \item Other oral diabetes medications
        \item Insulin
        \item I do not take any diabetes medication
    \end{checklist}
\end{surveybox}

\footnotesize
\begin{surveybox}[label={box:phq}, unbreakable]{Patient Health Questionnaire (PHQ-8)}
    \textbf{Little interest or pleasure in doing things.}
    
    \vspace{0.3cm}
    \begin{tabular}{c c c c}
    \cellcolor{gray!5}$\bigcirc$ & $\bigcirc$ & $\bigcirc$ & $\bigcirc$ \\
        Not at all & Several days & More than & Nearly  \\
        & & half days & every day
    \end{tabular}
    \vspace{0.3cm}

    \textbf{Feeling down, depressed, irritable or hopeless.}
    
    \vspace{0.3cm}
    \begin{tabular}{c c c c c}
        $\bigcirc$ & $\bigcirc$ & $\bigcirc$ & $\bigcirc$ \\
        Not at all & Several days & More than & Nearly  \\
        & & half days & every day
    \end{tabular}
    \vspace{0.3cm}

    \textbf{Trouble falling or staying asleep, or sleeping too much.}
    
    \vspace{0.3cm}
    \begin{tabular}{c c c c c}
        $\bigcirc$ & $\bigcirc$ & $\bigcirc$ & $\bigcirc$ \\
        Not at all & Several days & More than & Nearly  \\
        & & half days & every day
    \end{tabular}
    \vspace{0.3cm}

    \textbf{Feeling tired or having little energy .}
    
    \vspace{0.3cm}
    \begin{tabular}{c c c c c}
        $\bigcirc$ & $\bigcirc$ & $\bigcirc$ & $\bigcirc$ \\
        Not at all & Several days & More than & Nearly  \\
        & & half days & every day
    \end{tabular}
    \vspace{0.3cm}

    \textbf{Poor appetite or overeating.}
    
    \vspace{0.3cm}
    \begin{tabular}{c c c c c}
        $\bigcirc$ & $\bigcirc$ & $\bigcirc$ & $\bigcirc$ \\
        Not at all & Several days & More than & Nearly  \\
        & & half days & every day
    \end{tabular}
    \vspace{0.3cm}

    \textbf{Feeling bad about yourself – or that you are a failure or have let yourself or your family down.}
    
    \vspace{0.3cm}
    \begin{tabular}{c c c c c}
        $\bigcirc$ & $\bigcirc$ & $\bigcirc$ & $\bigcirc$ \\
        Not at all & Several days & More than & Nearly  \\
        & & half days & every day
    \end{tabular}
    \vspace{0.3cm}

    \textbf{Trouble concentrating on things, such as school work, reading or watching television .}
    
    \vspace{0.3cm}
    \begin{tabular}{c c c c c}
        $\bigcirc$ & $\bigcirc$ & $\bigcirc$ & $\bigcirc$ \\
        Not at all & Several days & More than & Nearly  \\
        & & half days & every day
    \end{tabular}
    \vspace{0.3cm}

    \textbf{Moving or speaking so slowly that other people could have noticed? Or the opposite – being so fidgety or restless that you have been moving around a lot more than usual.}
    
    \vspace{0.3cm}
    \begin{tabular}{c c c c c}
        $\bigcirc$ & $\bigcirc$ & $\bigcirc$ & $\bigcirc$ \\
        Not at all & Several days & More than & Nearly  \\
        & & half days & every day
    \end{tabular}
\end{surveybox}

\footnotesize
\begin{surveybox}[label={box:gad},unbreakable]{Generalized Anxiety Disorder (GAD-7)}

\textbf{Over the last two weeks, how often have you been bothered by the following problems?}
\newline
\newline
    \textbf{Feeling nervous, anxious, or on edge.}
    
    \vspace{0.3cm}
    \begin{tabular}{c c c c c}
        $\bigcirc$ & $\bigcirc$ & $\bigcirc$ & $\bigcirc$ \\
        Not at all & Several days & More than & Nearly  \\
        & & half days & every day
    \end{tabular}
    \vspace{0.3cm}

    \textbf{Not being able to stop or control worrying.}
    
    \vspace{0.3cm}
    \begin{tabular}{c c c c c}
        $\bigcirc$ & $\bigcirc$ & $\bigcirc$ & $\bigcirc$ \\
        Not at all & Several days & More than & Nearly  \\
        & & half days & every day
    \end{tabular}
    \vspace{0.3cm}

    \textbf{Worrying too much about different things.}
    
    \vspace{0.3cm}
    \begin{tabular}{c c c c c}
        $\bigcirc$ & $\bigcirc$ & $\bigcirc$ & $\bigcirc$ \\
        Not at all & Several days & More than & Nearly  \\
        & & half days & every day
    \end{tabular}
    \vspace{0.3cm}

    \textbf{Trouble relaxing.}
    
    \vspace{0.3cm}
    \begin{tabular}{c c c c c}
        $\bigcirc$ & $\bigcirc$ & $\bigcirc$ & $\bigcirc$ \\
        Not at all & Several days & More than & Nearly  \\
        & & half days & every day
    \end{tabular}
    \vspace{0.3cm}

    \textbf{Being so restless that it is hard to sit still.}
    
    \vspace{0.3cm}
    \begin{tabular}{c c c c c}
        $\bigcirc$ & $\bigcirc$ & $\bigcirc$ & $\bigcirc$ \\
        Not at all & Several days & More than & Nearly  \\
        & & half days & every day
    \end{tabular}
    \vspace{0.3cm}

    \textbf{Becoming easily annoyed or irritable.}
    
    \vspace{0.3cm}
    \begin{tabular}{c c c c c}
        $\bigcirc$ & $\bigcirc$ & $\bigcirc$ & $\bigcirc$ \\
        Not at all & Several days & More than & Nearly  \\
        & & half days & every day
    \end{tabular}
    \vspace{0.3cm}

    \textbf{Feeling afraid, as if something awful
might happen.}
    
    \vspace{0.3cm}
    \begin{tabular}{c c c c c}
        $\bigcirc$ & $\bigcirc$ & $\bigcirc$ & $\bigcirc$ \\
        Not at all & Several days & More than & Nearly  \\
        & & half days & every day
    \end{tabular}

\end{surveybox}

\footnotesize
\begin{surveybox}[label={box:pss}, unbreakable]{Perceived Stress Scale (PSS)}

    \textbf{In the last month, how often have you been upset because of something that happened unexpectedly?}
    
    \vspace{0.3cm}
    \begin{tabular}{c c c c c c}
        $\bigcirc$ & $\bigcirc$ & $\bigcirc$ & $\bigcirc$ & $\bigcirc$ & $\bigcirc$ \\
        Never & Almost & Sometimes & Fairly & Often & Very  \\
         & Never &  & Often &  & Often  \\
    \end{tabular}
    \vspace{0.3cm}

    \textbf{In the last month, how often have you felt that you were unable to control the important things in your life?}
    
    \vspace{0.3cm}
    \begin{tabular}{c c c c c c}
        $\bigcirc$ & $\bigcirc$ & $\bigcirc$ & $\bigcirc$ & $\bigcirc$ & $\bigcirc$ \\
        Never & Almost & Sometimes & Fairly & Often & Very  \\
         & Never &  & Often &  & Often  \\
    \end{tabular}
    \vspace{0.3cm}

    \textbf{In the last month, how often have you felt nervous and stressed?}
    
    \vspace{0.3cm}
    \begin{tabular}{c c c c c c}
        $\bigcirc$ & $\bigcirc$ & $\bigcirc$ & $\bigcirc$ & $\bigcirc$ & $\bigcirc$ \\
        Never & Almost & Sometimes & Fairly & Often & Very  \\
         & Never &  & Often &  & Often  \\
    \end{tabular}
    \vspace{0.3cm}

    \textbf{In the last month, how often have you felt confident about your ability to handle your personal problems?}
    
    \vspace{0.3cm}
    \begin{tabular}{c c c c c c}
        $\bigcirc$ & $\bigcirc$ & $\bigcirc$ & $\bigcirc$ & $\bigcirc$ & $\bigcirc$ \\
        Never & Almost & Sometimes & Fairly & Often & Very  \\
         & Never &  & Often &  & Often  \\
    \end{tabular}
    \vspace{0.3cm}

    \textbf{In the last month, how often have you felt that things were going your way?}
    
    \vspace{0.3cm}
    \begin{tabular}{c c c c c c}
        $\bigcirc$ & $\bigcirc$ & $\bigcirc$ & $\bigcirc$ & $\bigcirc$ & $\bigcirc$ \\
        Never & Almost & Sometimes & Fairly & Often & Very  \\
         & Never &  & Often &  & Often  \\
    \end{tabular}
    \vspace{0.3cm}

    \textbf{In the last month, how often have you found that you could not cope with all the things that you had to do?}
    
    \vspace{0.3cm}
    \begin{tabular}{c c c c c c}
        $\bigcirc$ & $\bigcirc$ & $\bigcirc$ & $\bigcirc$ & $\bigcirc$ & $\bigcirc$ \\
        Never & Almost & Sometimes & Fairly & Often & Very  \\
         & Never &  & Often &  & Often  \\
    \end{tabular}
    \vspace{0.3cm}

    \textbf{In the last month, how often have you been able to control irritations in your life?}
    
    \vspace{0.3cm}
    \begin{tabular}{c c c c c c}
        $\bigcirc$ & $\bigcirc$ & $\bigcirc$ & $\bigcirc$ & $\bigcirc$ & $\bigcirc$ \\
        Never & Almost & Sometimes & Fairly & Often & Very  \\
         & Never &  & Often &  & Often  \\
    \end{tabular}
    \vspace{0.3cm}

    \textbf{ In the last month, how often have you felt that you were on top of things?}
    
    \vspace{0.3cm}
    \begin{tabular}{c c c c c c}
        $\bigcirc$ & $\bigcirc$ & $\bigcirc$ & $\bigcirc$ & $\bigcirc$ & $\bigcirc$ \\
        Never & Almost & Sometimes & Fairly & Often & Very  \\
         & Never &  & Often &  & Often  \\
    \end{tabular}
    \vspace{0.3cm}

    \textbf{In the last month, how often have you been angered because of things that happened that were outside of your control?}
    
    \vspace{0.3cm}
    \begin{tabular}{c c c c c c}
        $\bigcirc$ & $\bigcirc$ & $\bigcirc$ & $\bigcirc$ & $\bigcirc$ & $\bigcirc$ \\
        Never & Almost & Sometimes & Fairly & Often & Very  \\
         & Never &  & Often &  & Often  \\
    \end{tabular}
    \vspace{0.3cm}

    \textbf{In the last month, how often have you felt difficulties were piling up so high that you could not overcome them?}
    
    \vspace{0.3cm}
    \begin{tabular}{c c c c c c}
        $\bigcirc$ & $\bigcirc$ & $\bigcirc$ & $\bigcirc$ & $\bigcirc$ & $\bigcirc$ \\
        Never & Almost & Sometimes & Fairly & Often & Very  \\
         & Never &  & Often &  & Often  \\
    \end{tabular}

\end{surveybox}

\footnotesize
\begin{surveybox}[label={box:promis_a}, unbreakable]{PROMIS Sleep Related Impairment 8a}

    \textbf{In the past 7 days...}
    \newline
    \newline
    \textbf{I had a hard time getting things done because I was sleepy}
    
    \vspace{0.3cm}
    \begin{tabular}{c c c c c c}
        $\bigcirc$ & $\bigcirc$ & $\bigcirc$ & $\bigcirc$ & $\bigcirc$ \\
        Not  & A little  & Somewhat & Quite a  & Very   \\
        at all  & bit &  & bit  & much  \\
    \end{tabular}
     \vspace{0.3cm}

    \textbf{I felt alert when I woke up}
    
    \vspace{0.3cm}
    \begin{tabular}{c c c c c c}
        $\bigcirc$ & $\bigcirc$ & $\bigcirc$ & $\bigcirc$ & $\bigcirc$ \\
        Not  & A little  & Somewhat & Quite a  & Very   \\
        at all  & bit &  & bit  & much  \\
    \end{tabular}
     \vspace{0.3cm}
     
    \textbf{I felt tired}
    
    \vspace{0.3cm}
    \begin{tabular}{c c c c c c}
        $\bigcirc$ & $\bigcirc$ & $\bigcirc$ & $\bigcirc$ & $\bigcirc$ \\
        Not  & A little  & Somewhat & Quite a  & Very   \\
        at all  & bit &  & bit  & much  \\
    \end{tabular}
     \vspace{0.3cm}

    \textbf{I had problems during the day because of poor sleep}
    
    \vspace{0.3cm}
    \begin{tabular}{c c c c c c}
        $\bigcirc$ & $\bigcirc$ & $\bigcirc$ & $\bigcirc$ & $\bigcirc$ \\
        Not  & A little  & Somewhat & Quite a  & Very   \\
        at all  & bit &  & bit  & much  \\
    \end{tabular}
     \vspace{0.3cm}

    \textbf{I had a hard time concentrating because of poor sleep}
    
    \vspace{0.3cm}
    \begin{tabular}{c c c c c c}
        $\bigcirc$ & $\bigcirc$ & $\bigcirc$ & $\bigcirc$ & $\bigcirc$ \\
        Not  & A little  & Somewhat & Quite a  & Very   \\
        at all  & bit &  & bit  & much  \\
    \end{tabular}
     \vspace{0.3cm}

    \textbf{I felt irritable because of poor sleep}
    
    \vspace{0.3cm}
    \begin{tabular}{c c c c c c}
        $\bigcirc$ & $\bigcirc$ & $\bigcirc$ & $\bigcirc$ & $\bigcirc$ \\
        Not  & A little  & Somewhat & Quite a  & Very   \\
        at all  & bit &  & bit  & much  \\
    \end{tabular}
     \vspace{0.3cm}
     
    \textbf{I was sleepy during the daytime}
    
    \vspace{0.3cm}
    \begin{tabular}{c c c c c c}
        $\bigcirc$ & $\bigcirc$ & $\bigcirc$ & $\bigcirc$ & $\bigcirc$ \\
        Not  & A little  & Somewhat & Quite a  & Very   \\
        at all  & bit &  & bit  & much  \\
    \end{tabular}
     \vspace{0.3cm}
     
    \textbf{I had trouble staying awake during the day}
    
    \vspace{0.3cm}
    \begin{tabular}{c c c c c c}
        $\bigcirc$ & $\bigcirc$ & $\bigcirc$ & $\bigcirc$ & $\bigcirc$ \\
        Not  & A little  & Somewhat & Quite a  & Very   \\
        at all  & bit &  & bit  & much  \\
    \end{tabular}
     \vspace{0.3cm}
     
\end{surveybox}

\footnotesize
\begin{surveybox}[label={box:promis_b}, unbreakable]{PROMIS Sleep Disturbance 8b}

    \textbf{In the past 7 days...}
    \newline
    \newline
    \textbf{My sleep was restless}
    
    \vspace{0.3cm}
    \begin{tabular}{c c c c c c}
        $\bigcirc$ & $\bigcirc$ & $\bigcirc$ & $\bigcirc$ & $\bigcirc$ \\
        Not  & A little  & Somewhat & Quite a  & Very   \\
        at all  & bit &  & bit  & much  \\
    \end{tabular}
     \vspace{0.3cm}

    \textbf{I was satisfied with my sleep}
    
    \vspace{0.3cm}
    \begin{tabular}{c c c c c c}
        $\bigcirc$ & $\bigcirc$ & $\bigcirc$ & $\bigcirc$ & $\bigcirc$ \\
        Not  & A little  & Somewhat & Quite a  & Very   \\
        at all  & bit &  & bit  & much  \\
    \end{tabular}
     \vspace{0.3cm}
     
    \textbf{My sleep was refreshing}
    
    \vspace{0.3cm}
    \begin{tabular}{c c c c c c}
        $\bigcirc$ & $\bigcirc$ & $\bigcirc$ & $\bigcirc$ & $\bigcirc$ \\
        Not  & A little  & Somewhat & Quite a  & Very   \\
        at all  & bit &  & bit  & much  \\
    \end{tabular}
     \vspace{0.3cm}

    \textbf{I had difficulty falling asleep }
    
    \vspace{0.3cm}
    \begin{tabular}{c c c c c c}
        $\bigcirc$ & $\bigcirc$ & $\bigcirc$ & $\bigcirc$ & $\bigcirc$ \\
        Not  & A little  & Somewhat & Quite a  & Very   \\
        at all  & bit &  & bit  & much  \\
    \end{tabular}
     \vspace{0.3cm}

    \textbf{I had trouble staying asleep}

    \vspace{0.3cm}
    \begin{tabular}{c c c c c c}
        $\bigcirc$ & $\bigcirc$ & $\bigcirc$ & $\bigcirc$ & $\bigcirc$ \\
        Never  & Rarely  & Sometimes & Often  & Always   \\
          &  &  &   &   \\
    \end{tabular}
     \vspace{0.3cm}

    \textbf{I had trouble sleeping}
    
    \vspace{0.3cm}
    \begin{tabular}{c c c c c c}
        $\bigcirc$ & $\bigcirc$ & $\bigcirc$ & $\bigcirc$ & $\bigcirc$ \\
        Never  & Rarely  & Sometimes & Often  & Always   \\
          &  &  &   &   \\
    \end{tabular}
     \vspace{0.3cm}
     
    \textbf{I got enough sleep}
    
    \vspace{0.3cm}
    \begin{tabular}{c c c c c c}
        $\bigcirc$ & $\bigcirc$ & $\bigcirc$ & $\bigcirc$ & $\bigcirc$ \\
        Never  & Rarely  & Sometimes & Often  & Always   \\
          &  &  &   &   \\
    \end{tabular}
     \vspace{0.3cm}
     
    \textbf{My sleep quality was}
    
    \vspace{0.3cm}
    \begin{tabular}{c c c c c c}
        $\bigcirc$ & $\bigcirc$ & $\bigcirc$ & $\bigcirc$ & $\bigcirc$ \\
        Very  & Poor  & Fair & Good  & Very   \\
        poor &  &  &   & good  \\
    \end{tabular}
     \vspace{0.3cm}
     
\end{surveybox}